%% file: main.tex
\documentclass[runningheads]{llncs}

\usepackage{eccv}

\usepackage{eccvabbrv}

\usepackage{graphicx}
\usepackage{amsmath}
\usepackage{amssymb}
\usepackage{booktabs}
\usepackage{algorithm}
\usepackage{algpseudocode}
\usepackage{bbm}
\usepackage{multirow}
\usepackage[dvipsnames]{xcolor}
\usepackage{colortbl}
\usepackage{enumitem}
\usepackage{pifont}
\usepackage{tikz}
\usepackage{caption}
\usepackage{wrapfig}
\usepackage{dblfloatfix}
\usepackage[accsupp]{axessibility}

\newcommand{\checkyes}{\ding{51}}
\newcommand{\checkno}{\ding{55}}

\usepackage{hyperref}

\usepackage{orcidlink}

\begin{document}

\title{Adaptive Bounding Box Uncertainties via Two-Step Conformal Prediction}

\author{
Alexander Timans\inst{1}\thanks{Correspondence to <\texttt{a.r.timans@uva.nl}>}
\and
Christoph-Nikolas Straehle\inst{2}
\and
Kaspar Sakmann\inst{2}
\and \\
Eric Nalisnick\inst{1}
}

\authorrunning{A. Timans~et al.}

\institute{
UvA-Bosch Delta Lab, University of Amsterdam
\and
Bosch Center for AI, Robert Bosch GmbH
}

\maketitle

\input{text/abstract}    
\input{text/intro}

\input{text/background}
\input{text/related}

\input{text/methods_box}
\input{text/methods_label}

\input{text/exp}
\input{text/conclusion}

\section*{Acknowledgements} 
We thank members of the Bosch-UvA Delta Lab and anonymous reviewers for helpful discussions and feedback. This project was generously supported by the Bosch Center for Artificial Intelligence.

\bibliographystyle{splncs04}
\bibliography{ref}

\clearpage
\setcounter{page}{1}
\appendix
\input{text/X_appendix}

\end{document}

%% file: text/abstract.tex
\begin{abstract}
Quantifying a model’s predictive uncertainty is essential for safety-critical applications such as autonomous driving. We consider quantifying such uncertainty for multi-object detection. In particular, we leverage conformal prediction to obtain uncertainty intervals with guaranteed coverage for object bounding boxes. One challenge in doing so is that bounding box predictions are conditioned on the object's class label. Thus, we develop a novel two-step conformal approach that propagates uncertainty in predicted class labels into the uncertainty intervals of bounding boxes. This broadens the validity of our conformal coverage guarantees to include incorrectly classified objects, thus offering more actionable safety assurances. Moreover, we investigate novel ensemble and quantile regression formulations to ensure the bounding box intervals are adaptive to object size, leading to a more balanced coverage. Validating our two-step approach on real-world datasets for 2D bounding box localization, we find that desired coverage levels are satisfied with practically tight predictive uncertainty intervals. 

\keywords{Object Detection \and Conformal Prediction \and Uncertainty}
\end{abstract}

%% file: text/intro.tex
\section{Introduction}
\label{sec:intro}

Safety-critical applications in domains such as autonomous transportation \cite{mcallister2017concrete, watkins2023autonomflying} and mobile robotics \cite{lutjens2019safe} benefit greatly from accurate estimates of the model's predictive uncertainty. Yet one obstacle to principled uncertainty quantification (UQ) for computer vision is the pervasive use of deep neural networks, which are often unamenable to traditional techniques for UQ. The framework of \emph{Conformal Prediction} (CP) \cite{v.vovk2005, g.shafer2008, angelopoulos2023gentle} enables a form of distribution-free UQ that is agnostic to the predictive model's structure, rendering it well-suited for such `black-box' models.

\begin{figure}[t]
    \centering
    \includegraphics[width=\linewidth]{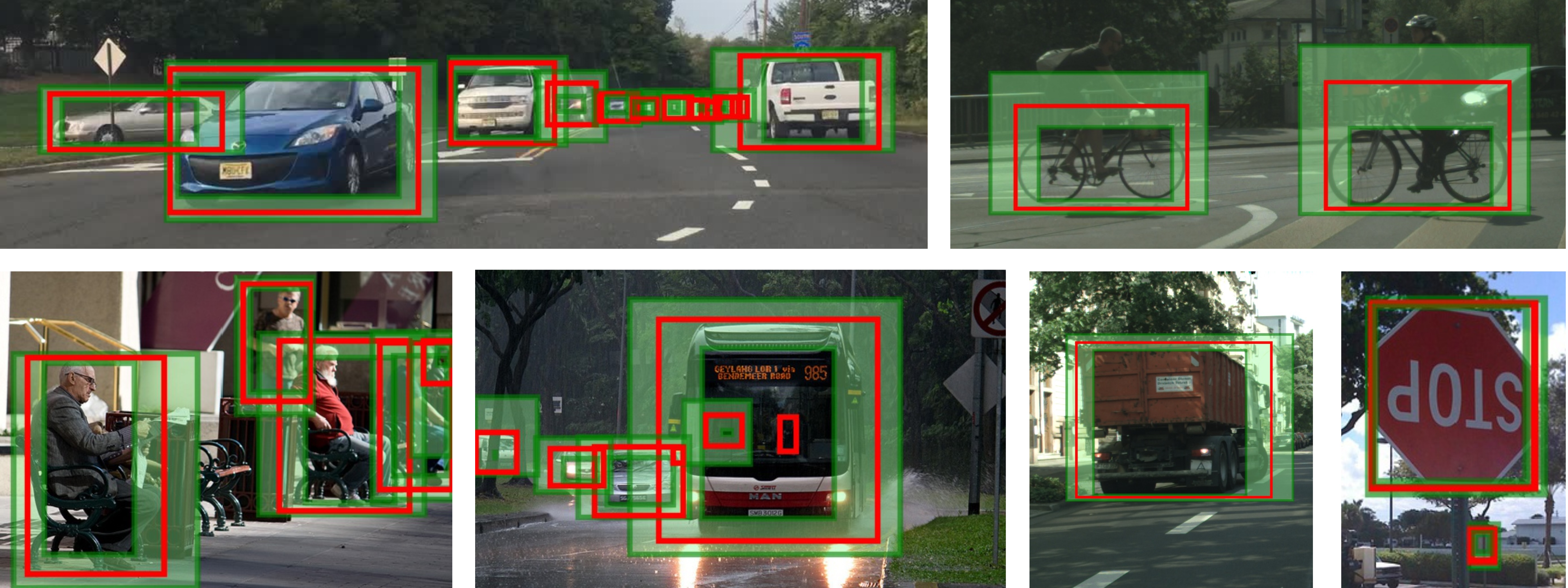}
    \caption{Examples of our method for multiple classes on test images. True bounding boxes are in red, two-sided prediction interval regions are shaded in green. Produced uncertainty estimates come with a probabilistic coverage guarantee of the true boxes.}
    \label{fig:pi-samples}
\end{figure}

\begin{figure*}[t]
    \centering
    \includegraphics[width=\linewidth]{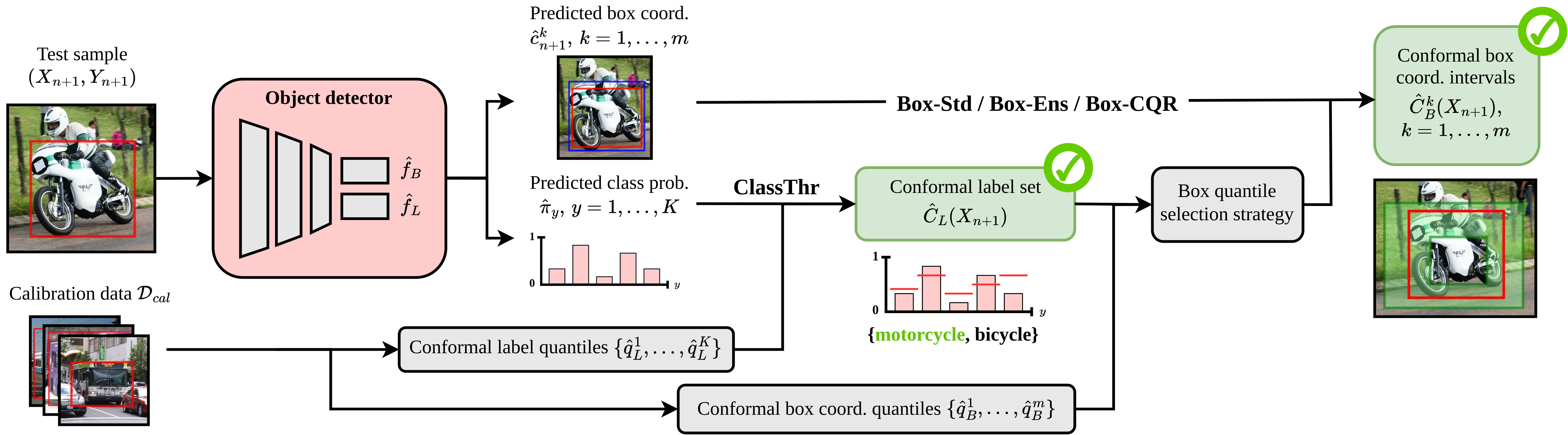}
    \caption{A diagram of our proposed \emph{two-step} conformal approach. We compute conformal quantiles for both class labels and box coordinates on calibration data following the CP framework. These are used on the predictions of a `black-box' object detector for a new test sample to (1) form a conformal label set with guarantee (\checkyes) which informs our box quantile choice, and (2) form a conformal prediction interval for the bounding box with guarantee (\checkyes), providing a reliable predictive uncertainty estimate.}
    \label{fig:method-diagram}
\end{figure*}

In this work, we propose a CP framework designed to quantify predictive uncertainties in multi-object detection tasks with multiple classes (see \autoref{fig:method-diagram}). CP allows us to produce computationally cheap, \emph{post-hoc} distribution-free prediction intervals, which come equipped with a coverage guarantee for the true bounding boxes of previously unseen objects (of known classes). Specifically, we provide users with the following safety assurance: \emph{``The conformal prediction interval covers the object's true bounding box with probability $(1-\alpha)$ for any known object class''}, where $\alpha$ is an acceptable margin of error. Such a guarantee can, \eg, in the context of autonomous driving, help certify collision avoidance by steering clear of the outer interval bounds, or in the case of robot picking, enforce cautious handling by demarcating a reliable grasping zone via the inner bounds. We provide visual examples of our obtained intervals in \autoref{fig:pi-samples} and \autoref{app:more-vis}.

Employing strategies based on ensembling and quantile regression, we ensure that the obtained intervals are adaptive to object size: they may grow or shrink in individual dimensions to account for object variability and prediction difficulty. A challenge to the desired assurance is that constructed intervals rely on the model's predicted class labels, which may be erroneous. We thus introduce an additional conformal step over the class labels, shielding against misclassification and ensuring that downstream coverage is satisfied. That is, our \emph{two-step} conformal pipeline remains theoretically and empirically valid regardless of the underlying object detector's predictive performance for either class labels or box coordinates -- the incurred costs are solely reflected in the obtained prediction interval sizes. In the experiments, we apply our methodology to multiple classes on several real-world 2D object detection datasets. We obtain bounding box prediction intervals that adhere to the desired guarantee, and are both adaptive and practically useful for downstream decision-making.

To summarize, our core contribution is an \emph{end-to-end} framework for safe bounding box uncertainties which is \emph{post-hoc}, efficient, and generalizable. In that process, we introduce several original concepts such as (i) ensemble and quantile CP adaptations for object detection, (ii) leveraging strong class-conditional guarantees for multi-class settings, and (iii) proposing a sequential two-step approach that propagates classification uncertainties forward.

%% file: text/background.tex
\section{Background}
\label{sec:background}

We begin by providing background on conformal prediction and the desired coverage guarantees, and then relate our object detection setting to it.

\subsection{Conformal prediction}
\label{subsec:background_conformal}

We consider the most common setting of split CP \cite{h.papadopoulos2007}, where we perform a single split to obtain hold-out calibration data $\mathcal{D}_{cal}=\{(X_i,Y_i)\}_{i=1}^{n} \sim P_{XY}$, as opposed to alternative partitioning schemes \cite{v.vovk2015, rf.barber2021}. If the general conformal procedure outlined in Algorithm \ref{algo:split-cp} (deferred to \autoref{app:split-cp}) is followed, a coverage guarantee for an unseen test sample $(X_{n+1},Y_{n+1}) \sim P_{XY}$ is provided in terms of a prediction set $\hat{C}(X_{n+1})$, where a finite-sample, distribution-free guarantee is given over the event of $\hat{C}(X_{n+1})$ containing $Y_{n+1}$. 

That is, assuming the samples $\mathcal{D}_{cal} \cup \{(X_{n+1}, Y_{n+1})\}$ are exchangeable -- a relaxation of the \emph{i.i.d.} assumption -- we obtain a probabilistic guarantee that
\begin{equation} 
    \mathbb{P}(Y_{n+1} \in \hat{C}(X_{n+1})) \ge 1-\alpha
\label{eq:conf-guarant}
\end{equation}
for some tolerated miscoverage rate $\alpha \in (0,1)$ \cite{g.shafer2008}. The provided guarantee is \emph{marginally} valid, since it holds on average across any sample $(X_{n+1},Y_{n+1})$ and set $\mathcal{D}_{cal}$ drawn from some fixed distribution $P_{XY}$ over $\mathcal{X} \times \mathcal{Y}$. This is in contrast to the ideal scenario of \emph{conditionally} valid coverage per input $X_{n+1}$, which has been shown to be impossible to achieve in a distribution-free manner \cite{v.vovk2012, r.foygelbarber2020}. However, recent work on in-between notions of conditionality such as \emph{group-} \cite{y.romano2020a, c.jung2022} and \emph{feature-conditional} \cite{m.sesia2021} strive towards more granular guarantees. 

In particular, \emph{class-conditional} validity can be achieved by applying CP separately to samples from each class \cite{m.cauchois2021, shi2013applications, v.vovk2005, m.sadinle2019a}, yielding the following guarantee:
\begin{equation}
    \mathbb{P}(Y_{n+1} \in \hat{C}(X_{n+1}) | Y_{n+1} = y) \ge 1-\alpha \quad \forall y \in \mathcal{Y},
\label{eq:class-cond-validity}
\end{equation}
where $\mathcal{Y} = \{1, \dots, K\}$ are distinct class labels. The class-conditional guarantee in \autoref{eq:class-cond-validity} is stronger and implies \autoref{eq:conf-guarant}, in that we aim to control the miscoverage level for samples within \emph{each class}. It also permits setting individual miscoverage levels $\{\alpha_y\}_{y \in \mathcal{Y}}$ per class if desired, and is robust to imbalances in class proportions \cite{podkopaev2021distribution, p.toccaceli2019a}. Such a class-conditional guarantee is precisely what we aim to provide. 

\textbf{Classification and regression.} Applying CP to a classification task yields conformal label \emph{prediction sets} $\hat{C}_{L}(X_{n+1}) \subseteq \{1, \dots, K\}$ as finite subsets of the $K$ class labels, at a target miscoverage level $\alpha_L$. For regression, the sets $\hat{C}_{B}(X_{n+1}) \subseteq \mathbb{R}$ take the form of \emph{prediction intervals} (PIs) on the target domain, at a target miscoverage level $\alpha_B$. Naturally, we have both $(\alpha_L, \alpha_B) \in (0,1)^2$.

\subsection{Object detection}
\label{subsec:background_od}

We next formalize our multi-object detection setting. Consider an input image $X \in \mathbb{R}^{H \times W \times D}$, where $H$, $W$ and $D$ correspond to image height, width and depth. For each image in $\mathcal{D}_{cal}$ we also receive a set of tuples $(c^1, c^2, c^3, c^4, l)$, where $(c^1, c^2, c^3, c^4) \in \mathbb{R}^4$ are the coordinates indicating an object's bounding box location in the image, and $l \in \{1, \dots, K\}$ represents the object's class label. 

Each tuple parameterizes an object, with a total of $O(X)$ true objects located in the image. For image $X$ we thus have objects $\{(c^1, c^2, c^3, c^4, l)_j\}_{j=1}^{O(X)}$. Note that the model predicts $\hat{O}(X)$ objects, and it is possible that $O(X) \ne \hat{O}(X)$. We model every object as an individual sample for our CP procedures, \ie, the same input image $X$ can produce multiple calibration samples of shape $(X, (c^1, c^2, c^3, c^4, l)_j)$, where $j=1, \dots, O(X)$ denote the contained objects. 

\textbf{Object detection model.} For our object detector $\hat{f}$, we define two separate output heads. The probabilistic classification head is defined as the map $\hat{f}_L: X \mapsto (\hat{\pi}_1,\dots,\hat{\pi}_K)$, where $\hat{\pi}_y$ is the model's estimate of the true class probability $\pi_{y}$ of some object in image $X$ belonging to class $y$. The object's class label is then $l = \arg\max_{y \in \{1,\dots,K\}}\,\hat{\pi}_y$. The bounding box regression head, denoted as $\hat{f}_B: X \mapsto (\hat{c}^1, \hat{c}^2, \hat{c}^3, \hat{c}^4)$, maps to an object's real-valued bounding box coordinates.

\subsection{Conformal prediction for object detection}
\label{subsec:background_conf_od}

Given our multi-object detection setting, we consider a class-conditional CP approach to be particularly meaningful. It is sensible to only leverage information on detected objects of the same class, \eg, class `car', to construct PIs for new objects of that class. In contrast, a general marginal approach will unintuitively also employ information from unrelated classes, such as `person' or `bicycle'.

We apply CP to the bounding boxes on a per-coordinate basis, previously denoted $(c^1, c^2, c^3, c^4)$. However, from now on let us consider the generalization to an arbitrary amount of coordinates $c^k, \, k=1, \dots, m$ \footnote{This easily permits extending our approach to higher-dimensional object parameterizations such as 3D bounding boxes.}. If we consider the class label $l$ within each group of objects belonging to a common class as fixed (since the same label is shared), the response of an individual sample $(X_i, Y_i)$ can be interpreted as a realization of the $m$ coordinates only, \ie, we define $Y_i := (c_i^1, \dots, c_i^m) \in \mathbb{R}^m$. The desired guarantee in \autoref{eq:class-cond-validity} is re-interpreted as 
\begin{equation}
    \mathbb{P}\left(\bigcap_{k=1}^{m}\left(c_{n+1}^k \in \hat{C}_{B}^{k}(X_{n+1})\right) \, | \, l_{n+1} = y\right) \ge 1-\alpha_B \quad \forall y \in \mathcal{Y},
    \label{eq:class-cond-validity-od}
\end{equation}
where components are indexed accordingly per specific coordinate dimension. For example, $\hat{C}_{B}^k(X_{n+1})$ is the $k$-th coordinate's prediction interval of an object of class $y$ (its class label $l_{n+1}$ matches $y$) located in image $X_{n+1}$. Applying CP per coordinate gives rise to multiple testing issues, which we address in \autoref{subsec:methods_mht}. 

\textbf{Practical limitations.} Naively applying class-conditional CP to the box coordinates necessitates a correct class label prediction in order to satisfy validity. That is, for \autoref{eq:class-cond-validity-od} to hold, a valid PI construction requires $\hat{l}_{n+1} = l_{n+1}$ for any considered class $y \in \mathcal{Y}$. We alleviate this practically limiting dependence on the model's classification ability using a conformal set-based classifier in \autoref{sec:methods_label} (see also \autoref{fig:method-diagram}). However, we of course still rely on the model's general detection abilities: the provided guarantees only hold for true objects that are actually detected (true positives) and do not account for undetected objects (false negatives), as also noted by Andéol \etal \cite{f.degrancey2022, l.andeol2023a}. Finally, the assumption on data exchangeability underlying CP requires $P_{XY}$ to remain fixed, albeit recent works have explored CP under settings of mild or known distribution shifts \cite{angelopoulos2023gentle, fontana2023cpreview}. 

%% file: text/related.tex
\section{Related work}
\label{sec:related}

Many existing approaches for uncertainty in bounding box regression leverage standard UQ techniques such as Bayesian inference \cite{choi2019odbayes, kraus2019odbayes, wirges2019odbayes}, loss attenuation \cite{he2019attenuation, le2018attenuation, lee2022attenuation}, or practical approximations like Monte Carlo Dropout \cite{pengodmcdropout, harakehodmcdropout, millerodmcdropout, yelleniodmcdropout} and Deep Ensembles \cite{fengodensemble, wangodensemble, millerodensemble}. These can require substantial modifications to the model architecture or training procedure, and do not provide a guarantee or statement of assurance about provided estimation quality. See Feng \etal~\cite{feng2021reviewpod} for a recent survey. A complementary line of work investigates the calibration of object detectors \cite{phanodcalibration, kuppersodcalibration, neumannodcalibration}, which can benefit our approach by improving the underlying `black-box' probabilistic model.

Conformal approaches have recently gained traction for computer vision and related tasks, with applications such as image classification \cite{y.romano2020aps, a.angelopoulos2022raps}, geometric pose estimation \cite{yang2023cpkeypoint}, or tracking and trajectory planning \cite{su2024cptracking, lindemann2023cpmpc, luo2022cpmpc, muthalicpmpc}. Yet, the domain remains comparatively unexplored given current surveys \cite{angelopoulos2023gentle, fontana2023cpreview}. Specific attempts at principled UQ for bounding boxes include using the Probably Approximatly Correct (PAC) framework to produce guarantees by composition of PAC sets at multiple modelling stages \cite{s.li2022pac, park2020pac}, or leveraging p-values and risk estimates obtained from concentration inequalities for related risk control \cite{an.angelopoulos2022ltt, bates2021distribution, an.angelopoulos2022aimgtoimg}. Such works differ from our two-step conformal approach in several ways, such as (i) considering different vision tasks and using different data modalities, (ii) integrating CP into complex modelling pipelines that cease to be \emph{post-hoc} and model-agnostic, or (iii) employing methods not based on conformal prediction. 

\noindent \textbf{Closest prior work.} Conformal PIs for bounding boxes have been previously considered by Andéol \etal~\cite{f.degrancey2022, l.andeol2023a}. However, a crucial limitation in their approach is that bounding box uncertainty is considered and evaluated for a \emph{single class} and for \emph{correctly classified} objects only. Thus only the simplest form of our guarantee in \autoref{eq:class-cond-validity-od} is provided, since the class label is known \emph{a priori} and therefore $\hat{l}_{n+1} = l_{n+1}$ trivially holds. This means that prevalent uncertainty in the class label predictions (which we address in \autoref{sec:methods_label}) is entirely ignored, making their approach unsuitable for settings with multiple interacting classes, such as autonomous driving. We also introduce several methodological improvements, such as (i) novel ensemble and quantile scoring functions for the bounding box setting, (ii) more informative two-sided intervals, and (iii) a multiple testing correction that exploits correlation structure between box coordinates, as opposed to more naive Bonferroni \cite{f.degrancey2022} or $\max$-corrections \cite{l.andeol2023a}.

%% file: text/methods_box.tex
\section{Conformal methods for box coordinates}
\label{sec:methods_box}

A key modelling decision in CP is the choice of scoring function $s:\mathcal{X}\times\mathcal{Y}\rightarrow\mathbb{R}$ to compute the required nonconformity scores (see \autoref{app:split-cp}). We consider three choices of scoring function and PI construction for each box coordinate $k \in \{1,\dots,m\}$, which we outline next. Additional implementation details can be found in \autoref{app:implement}.

\textbf{Standard conformal (Box-Std).} We firstly consider the standard approach of employing regression residuals $s(\hat{f}_{B}(X),\,Y) = |\hat{c}^k - c^k|$ as scores \cite{g.shafer2008}. The resulting PIs are constructed as $\hat{C}^{k}_{B}(X_{n+1}) = [\hat{c}^{k}_{n+1} - \hat{q}^k_B,\,\, \hat{c}^{k}_{n+1} + \hat{q}^k_B]$, where $\hat{q}^k_B$ denotes the computed conformal quantile for the $k$-th coordinate. While straightforward, this construction only permits for non-adaptive, fixed-width intervals. Andéol \etal~\cite{f.degrancey2022} use this approach to construct their one-sided intervals.

\textbf{Conformal ensemble (Box-Ens).} In order to produce more adaptive intervals, we next consider normalized residual scores \cite{j.lei2018} of the form $s(\hat{f}_{B}(X), \,Y) = |\hat{c}^k - c^k|/\hat{\sigma}(X)$, where $\hat{\sigma}$ is some form of heuristic uncertainty estimate (\ie, without guarantees) obtained from the underlying model. The resulting conformal PIs are constructed as $\hat{C}^{k}_{B}(X_{n+1}) = [\hat{c}^{k}_{n+1} - \hat{\sigma}(X_{n+1})\,\hat{q}^k_B,\,\, \hat{c}^{k}_{n+1} + \hat{\sigma}(X_{n+1})\,\hat{q}^k_B]$. By incorporating model uncertainty, the intervals can be re-scaled individually per coordinate to adapt their magnitude at test time. We can interpret this as an empirical conditioning on the particular test sample. We employ an ensemble of object detectors and quantify $\hat{\sigma}$ as the standard deviation of the ensemble's box coordinate predictions \cite{b.lakshminarayanan2017}. A joint coordinate prediction $\hat{c}^k$ is obtained from the ensemble via confidence-weighted box fusion \cite{r.solovyev2021}.

\textbf{Conformal quantile regression (Box-CQR).} As an alternative adaptive method, we extend the approach of Conformal Quantile Regression (CQR) \cite{y.romano2019} to our setting. Additional regression heads $\hat{Q}_B$ are trained with a \emph{quantile loss} alongside $\hat{f}_B$ to produce lower and upper quantile predictions $\hat{Q}_{\alpha_B/2}$ and $\hat{Q}_{1-\alpha_B/2}$ for the bounding box coordinates. Under regularity conditions, these predictors will asymptotically converge to the true conditional quantiles \cite{r.koenker1978, chaudhuri1991global}, motivating their viability. Following CQR, we define the scores as $s(\hat{Q}_B(X),\,Y) = \max\{\hat{Q}_{\alpha_B/2}(X) - c^k,\,c^k - \hat{Q}_{1-\alpha_B/2}(X)\}$, and construct the conformal PIs as $\hat{C}^{k}_{B}(X_{n+1}) = [\hat{Q}_{\alpha_B/2}(X_{n+1}) - \hat{q}^k_B,\,\, \hat{Q}_{1-\alpha_B/2}(X_{n+1}) + \hat{q}^k_B]$. The obtained interval ensures adaptivity through the use of its quantile predictions, which will differ in their distance relative to the mean coordinate prediction $\hat{c}^{k}_{n+1}$ per sample.

\subsection{Multiple testing correction}
\label{subsec:methods_mht}

Applying CP to each of the box coordinates $k=1,\dots,m$ separately gives rise to multiple testing issues, since the conformal procedure can be interpreted from a hypothesis testing view as running $m$ permutation tests on nonconformity in parallel \cite{v.vovk2005, shi2013applications}. This results in a guaranteed coverage of at most $(1 - m\cdot\alpha_B)$, as opposed to the desired rate of $(1 - \alpha_B)$ (see \autoref{app:mult-testing-problem} for further details). 

The naive Bonferroni correction \cite{vovk2020combiningpval} offers a possible remedy, since choosing $\alpha_{B}' = \alpha_{B}/m$ will satisfy target coverage. However, it is known to be overly conservative under positive dependency of the individual hypothesis \cite{vovk2022admissible}, a reasonable assumption given that the coordinates parametrize an object's bounding box jointly. In fact, Bates \etal~\cite{s.bates2022} assert that a set of conformal p-values exhibits positive dependency structure \emph{a priori} since they are jointly \emph{positive regression dependent on a subset} \cite{y.benjamini2001}. We leverage an alternate procedure by Timans \etal~\cite{timans2023maxrank}, which exploits correlation structure among box coordinates for a less conservative correction. Their $\max$-rank procedure adapts the Westfall \& Young~\cite{westfall1993resampling} permutation correction to make it suitable for the setting of conformal prediction (see \autoref{app:max-rank}). While the use of a $\max$-correction has been previously considered for CP \cite{l.andeol2023a, m.cauchois2021}, $\max$-rank operates in the more robust scale-invariant rank space. In addition, it requires less compute than previously proposed copula-based testing corrections \cite{sun2023copula, messoudi2021copula}.

%% file: text/methods_label.tex
\section{Class label prediction sets}
\label{sec:methods_label}

In practice, the object detector may incorrectly predict an object's class label given our multi-class setting. This complicates a direct application of class-conditional CP to an object's bounding box at test time, since we need to correctly select the conformal quantiles $\hat{q}^{k}_B,\,k \in \{1,\dots,m\}$ to construct bounding box intervals that satisfy \autoref{eq:class-cond-validity-od}. This limits intially provided safety assurances to correctly classified objects only, \ie, those where $\hat{l}_{n+1} = l_{n+1}$ successfully match.

To alleviate this restrictive dependence on the model's classification ability, our modelling pipeline in \autoref{fig:method-diagram} introduces an additional conformal step which preceeds the bounding box construction. Specifically, we consider applying CP to the model's classifier head $\hat{f}_L$ to first generate class label prediction sets $\hat{C}_L(X_{n+1})$ with a guarantee on label containment. These are subsequently used to select our box coordinate quantiles, ensuring the validity of our provided guarantees in \autoref{eq:class-cond-validity-od} is broadened to even include \emph{incorrectly} classified objects. 

We achieve this using another class-conditional CP approach on the class labels with a strict label coverage guarantee of 99\% (\ie, $\alpha_L = 0.01$). Thus, we approximate the condition $\hat{l}_{n+1} = l_{n+1}$ by effectively ensuring $l_{n+1} \in \hat{C}_L(X_{n+1})$. The resulting \emph{two-step} sequential approach maintains validity regardless of the object detector's classification or bounding box regression performance. The incurred costs are reflected in the obtained prediction set and interval sizes only. Our experiments in \autoref{subsec:exp_results_label} demonstrate that even under these strong safety assurances, our approach provides actionably tight PIs. We follow with a description of the employed conformal label set method and related baselines. 

\subsection{Conformal class thresholding (ClassThr)}
\label{subsec:methods_label_class}

We propose using a class-conditional variant of the prediction set classifier introduced by Sadinle \etal~\cite{m.sadinle2019a}, based on a similar conformal procedure as for the bounding boxes. Given our probabilistic classifier head $\hat{f}_L$, we define the scoring function $s(\hat{f}_L(X),\,y) = 1 - \hat{\pi}_{y}(X)$ for every class $y \in \mathcal{Y}$ to compute per-class conformal quantiles $\hat{q}_{L}^y$. The class label prediction set for a new object to classify is then given by $\hat{C}_{L}(X_{n+1}) = \{y \in \mathcal{Y}: \hat{\pi}_{y}(X_{n+1}) \ge 1 - \hat{q}_{L}^y\}$. The detailed procedure is given in Algorithm \ref{algo:label-cp}, deferred to \autoref{app:classthr-cp}. Importantly, class-conditional validity is ensured by comparing each class probability against its class-specific threshold on set inclusion. The class-conditional guarantee stated in \autoref{eq:class-cond-validity} can now be provided similarly for the object detector's classification task as 
\begin{equation}
    \mathbb{P}(l_{n+1} \in \hat{C}_{L}(X_{n+1}) \, | \, l_{n+1} = y) \ge 1-\alpha_{L} \quad \forall y \in \mathcal{Y}.
    \label{eq:class-cond-validity-label}
\end{equation}

\noindent \textbf{Impact on bounding box coverage guarantee.} A class-conditional guarantee is enforced for the label prediction sets (\autoref{eq:class-cond-validity-label}), since only imposing the weaker marginal guarantee could invalidate the subsequent class-conditional box guarantee (\autoref{eq:class-cond-validity-od}). If for instance a class is systematically undercovered, we would fail to retrieve the correct box quantiles for some of its associated objects, propagating the undercoverage down-stream. Our approach instead enforces guarantees of equivalent strength. Observe that we perform two distinct conformal procedures in sequence, rendering the coverage guarantees conditionally independent. Thus the down-stream coverage effect for their two-step application is that
\begin{equation}
\begin{split}
    &\mathbb{P}\left(l_{n+1} \in \hat{C}_{L}(X_{n+1}) \, \wedge \, \bigcap_{k=1}^{m}\left(c_{n+1}^k \in \hat{C}_{B}^{k}(X_{n+1})\right) \, | \, l_{n+1} = y\right) \\ &= \mathbb{P}\left(l_{n+1} \in \hat{C}_{L}(X_{n+1}) \, | \, l_{n+1} = y\right) \, \cdot \, \mathbb{P}\left(\bigcap_{k=1}^{m}\left(c_{n+1}^k \in \hat{C}_{B}^{k}(X_{n+1})\right) \, | \, l_{n+1} = y\right) \\ & \hspace{18em} \ge (1-\alpha_L)(1-\alpha_B) \quad \forall y \, \in \mathcal{Y}.
    \label{eq:sequential-appl-guarantee}
\end{split}
\end{equation}
That is, a preceeding label coverage guarantee of $(1-\alpha_L)$ will nominally only assure subsequent box coverage of $(1-\alpha_L)(1-\alpha_B)$. In our experiments we approximate $(1-\alpha_L)(1-\alpha_B) \approx (1-\alpha_B)$ by setting $\alpha_L=0.01$, thus alleviating the down-stream coverage reduction\footnote{As we observe in \autoref{subsec:exp_results_label}, enforcing $\alpha_L = 0$ leads to empirically inefficient box intervals.}. \autoref{eq:sequential-appl-guarantee} highlights that coverage trade-offs between objectives are possible depending on application-specific requirements. For example, a nominal box coverage of 90\% can be achieved by choosing either $\alpha_L=0.05, \alpha_B=0.05$ or $\alpha_L=0, \alpha_B=0.1$.

\subsection{Bounding box quantile selection}
\label{subsec:methods_label_downstream}

The obtained label prediction sets are subsequently used to select a valid box coordinate quantile for any interval construction following \autoref{sec:methods_box}. A natural quantile selection strategy is $\hat{q}_{B}^k = \max \{\hat{q}_{B}^{k,y}\}_{y \in \hat{C}_{L}(X_{n+1})} \, \forall k \in \{1,\dots,m\}$, where $\hat{q}_{B}^{k,y}$ is the quantile of the $k$-th coordinate for class $y$. Using a $\max$-operator on the label set is a valid but conservative approach, since all labels in the set are regarded as equally likely for every sample. Obtained box intervals thus tend to overcover, which could perhaps be alleviated with a different strategy, resulting in narrower PIs. Yet, we find even this straightforward selection to yield reasonably tight results. A hypothesis testing motivation for its use can be found in \autoref{app:box-quant-selection}.

\input{text/table_guarantees}

\subsection{Label set baselines}
\label{subsec:methods_label_baselines}

We compare obtained label predictions sets via conformal class thresholding (ClassThr) to several reasonable alternatives, whose nominal guarantees and expected empirical set sizes are outlined in \autoref{tab:nominal_guarantees_check}, and which we detail next.

\textbf{Top singleton set (Top).}  We return label prediction sets that only consist of the highest probability class for every sample, \ie, $\hat{C}_{L}(X_{n+1}) = \{y^*: \hat{\pi}_{y^*}(X_{n+1}) = \max_{y \in \{1,\dots,K\}} \,\hat{\pi}_{y}(X_{n+1})\}$. This approach returns singleton sets, is void of nominal guarantees, and its empirical coverage relies fully on the classifier's accuracy. The distinction to our initial condition $\hat{l}_{n+1} = l_{n+1}$, which we refer to as \textbf{Oracle}, is subtle: instead of ensuring correct quantile selection, we permit the use of potentially wrong quantiles to construct the box intervals.

\textbf{Density level set (Naive).} Assuming a perfectly calibrated classifier such that $\hat{\pi}_{y}(X) = \pi_{y}(X) \,\,\forall y \in \mathcal{Y}$, the optimal prediction set is provided by density level sets. That is, we collect all labels sorted by descending $\hat{\pi}_{y}$ until we reach probability mass $(1-\alpha_L)$. Under this assumption, prediction sets will also approach conditional coverage for any $X \in \mathcal{X}$ \cite{m.sadinle2019a, y.romano2020aps}. While unattainable in practical settings where the classifier tends to be miscalibrated (\ie, $\hat{\pi}_{y}(X) \neq \pi_{y}(X)$), it can be considered a theoretically motivated extension of the Top baseline.

\textbf{Full domain set (Full).} We consider taking the full set of possible class labels per sample, thus $\hat{C}_{L}(X_{n+1}) = \mathcal{Y}$ and $|\hat{C}_{L}(X_{n+1})| = K$. In combination with our quantile selection strategy this approach guarantees label coverage conditionally per sample, \ie, it ensures $\alpha_L = 0$. However, this comes at the cost of overly inflated label sets whose size is expected to propagate to the box intervals.

We do not consider other popular conformal approaches for classification such as APS \cite{y.romano2020aps} or RAPS \cite{a.angelopoulos2022raps} since they aim to empirically improve conditional coverage under the requirements of a \emph{marginal} guarantee -- these advantages do not extend to a class-conditional setting as ours.

%% file: text/table_guarantees.tex
\begin{table}[t]
    \centering
    \setlength{\tabcolsep}{6pt}
    \begin{tabular}{lclcl}
         & \multicolumn{2}{c}{\textbf{Label set}} & \multicolumn{2}{c}{\textbf{Box interval}} \\
         \cmidrule(lr){2-3}\cmidrule(lr){4-5}
         \multicolumn{1}{c}{\multirow{-1}{*}{\textbf{Method}}} & Guarantee & Size & Guarantee & Size \\
         \toprule
         Top & \checkno$^{*}$ & Single & \checkno$^{\dagger}$ & Small \\
         Naive & \checkno$^{*}$ & Small & \checkno$^{\dagger}$ & Small \\
         ClassThr & \checkyes & Medium & \checkyes & Medium \\
         Full & \checkyes & Large & \checkyes & Large \\
         \bottomrule
         \\
    \end{tabular}
    \caption{Provided \emph{nominal} coverage guarantees and expected \emph{empirical} prediction set/interval sizes for the considered label prediction set methods on the basis of both correctly and incorrectly classified objects. $^*$Top and Naive provide a label guarantee `for free' if $(1-\alpha_L)$ is below the classifier's accuracy level \emph{for each class}. $^\dagger$Top and Naive provide a box guarantee for correctly classified objects only. Naive may also satisfy both guarantees under practically unattainable perfect model calibration.}
    \label{tab:nominal_guarantees_check}
\end{table}

%% file: text/exp.tex
\section{Experiments}
\label{sec:exp}

For our experiments we primarily rely on pre-trained object detectors from \texttt{detectron2} \cite{wu2019detectron2}, based on a Faster R-CNN architecture and trained on COCO \cite{ty.lin2015coco}. We consider three datasets: COCO validation, Cityscapes \cite{Cordts2016Cityscapes} and BDD100k \cite{f.yu2020bdd}, which contain 2D bounding box annotations and are split into appropriate calibration and test sets. We run our two-step conformal procedure for a variety of classes, but focus reported results on a coherent set of object classes which exists across datasets: \emph{person, bicycle, motorcycle, car, bus} and \emph{truck} (see \autoref{app:dataset}).

Since the images can contain multiple objects, we require a pairing mechanism between true and predicted bounding boxes. Following prior work \cite{f.degrancey2022, l.andeol2023a} we perform Hungarian matching \cite{kuhn1955hungarian} based on an intersection-over-union (IoU) threshold of 0.5. Throughout, we set $\alpha_L=0.01, \alpha_B=0.1$ for a target box coverage of $(1-\alpha_L)(1-\alpha_B) \approx 90\%$, and employ $\max$-rank \cite{timans2023maxrank} for multiple testing correction. Results are averaged across multiple trials of data splitting. Additional results, including varying combinations of $(\alpha_L, \alpha_B)$, are in \autoref{app:results} \footnote{Our code is publicly available at \url{https://github.com/alextimans/conformal-od}.}.

\input{text/fig_violins}

\subsection{Metrics}
\label{subsec:exp_metrics}

Our approaches are validated by assessing the key desiderata of CP via relevant metrics described below, which jointly capture the desired notions of `reliable' uncertainty \cite{angelopoulos2023gentle}. We denote the test set of size $n_t$ as $\mathcal{D}_{test} = \{(X_j,Y_j)\}_{j=n+1}^{n+n_t}$.

\textbf{Validity.} We assess if nominal coverage guarantees are satisfied by verifying \emph{empirical coverage}, which we define in generality as 
\begin{equation}
    \label{eq:metric-cov}
    Cov = \frac{1}{n_t} \sum_{j=n+1}^{n+n_t}\mathbbm{1}[Y_{j} \in \hat{C}(X_{j})],
\end{equation}
where $\mathbbm{1}[\cdot]$ is the indicator function, of form $\mathbbm{1}[l_{j} \in \hat{C}_{L}(X_{j})]$ for label prediction sets and $\mathbbm{1}[\bigcap_{k=1}^{m}(c_{j}^k \in \hat{C}^{k}_B(X_{j}))]$ for box intervals. Note that $Cov$ is a random quantity parametrized by an empirical coverage distribution, and will deviate from nominal coverage based on factors such as calibration set size $|\mathcal{D}_{cal}|$ \cite{v.vovk2012}.

\noindent \textbf{Adaptivity.} To examine if target coverage is satisfied by an imbalance of over- and undercoverage across objects, similarly to \cite{a.angelopoulos2022raps, y.romano2020aps} we verify empirical coverage by stratification, namely across object sizes. We follow the COCO challenge \footnote{See \url{https://cocodataset.org/\#detection-eval}.} and stratify across three sizes by bounding box surface area: small ($Cov_S$, area $\le 32^2$), medium ($Cov_M$, area $\in (32^2, 96^2]$) and large ($Cov_L$, area $> 96^2$).

\textbf{Efficiency.} Obtained conformal prediction sets and intervals are desired to be as small as possible while still maintaining target coverage (\ie, remaining valid). We define the \emph{mean set size} for label prediction sets and \emph{mean prediction interval width ($MPIW$)} for bounding box prediction intervals as
\begin{equation}
    \label{eq:metric-size}
    \frac{1}{n_t} \sum_{j=n+1}^{n+n_t}|\hat{C}_L(X_{j})| \quad\text{and}\quad
    \frac{1}{n_{t} m} \sum_{j=n+1}^{n+n_t} \sum_{k=1}^{m}|\hat{C}_{B}^k(X_{j})|.
\end{equation}
\noindent That is, \emph{mean set size} denotes the average number of labels in the obtained sets, while $MPIW$ expresses the average interval width in terms of image pixels.

\textbf{Predictive performance.} We also follow standard practice and validate model performance using object detection-specific metrics from the COCO challenge, in particular average precision across multiple IoU thresholds (see \autoref{app:results}).

\input{text/table_baselines}

\subsection{Comparison of bounding box methods}
\label{subsec:exp_results}

Empirical coverage levels stratified by class labels as well as object sizes for the three proposed bounding box methods are displayed in \autoref{fig:violins} for BDD100k. We see that target coverage of 90\% is satisfied even per class, validating the class-conditional guarantees provided in \autoref{eq:class-cond-validity-od}. The visible coverage variations are explained by the differences in available calibration samples per class (see \autoref{app:emp-cov-distr} and \autoref{tab:datasets-ist}). We further observe that the fixed-width intervals of Box-Std may be large enough to cover small objects, but will fail to account for the magnitude of large ones, resulting in significant undercoverage. In contrast, the adaptive nature of Box-CQR and in particular Box-Ens via its scaling factor can better account for varying magnitudes, achieving higher coverage for large objects at a slight loss in efficiency due to a higher $MPIW$. Whilst coverage across small objects reduces somewhat, it now intuitively aligns with observed prediction difficulty (see \autoref{tab:ap-scores}). That is, objects which are more challenging to predict exhibit a higher variation and chance of miscoverage. We note that the improved coverage balance across object sizes is a purely empirical benefit of our adaptive designs -- the employed conformal procedures only aim to guarantee target coverage per class and do \emph{not} condition on object size.

\textbf{Baseline comparisons.} We further validate our conformal bounding box step by comparing to Andéol \etal~\cite{l.andeol2023a, f.degrancey2022}, modifying our own two-sided interval methods to produce one-sided PIs, and evaluating efficiency via $MPIW$ (see \autoref{app:baseline-impl} for details and their \emph{box stretch} metric). \autoref{tab:baselines} 
demonstrates that we achieve marginally tighter intervals even in their own, more restricted setting, while remaining equally valid. We also evaluate generated uncertainties via deep ensembles (DeepEns) \cite{b.lakshminarayanan2017} and GaussianYOLO \cite{choi2019odbayes}, two popular UQ approaches for object detection. Results confirm the unreliability of produced uncertainties due to their \emph{lack of guarantees}, as seen by severe undercoverage in both settings. Finally, a comparison of the $\max$-rank correction to Bonferroni in \autoref{app:bonf-baseline-results} asserts that substantially tighter PIs can be obtained with our employed correction.

\input{text/fig_metrics}

\subsection{Results for the two-step approach}
\label{subsec:exp_results_label}

After having benchmarked our bounding box step, we next compare the full \emph{two-step} approach by adding the preceeding conformal step for class labels via ClassThr, and comparing to the proposed label set baselines in \autoref{subsec:methods_label_baselines}. Each method's nominal guarantee and expected efficiency is displayed in \autoref{tab:nominal_guarantees_check}. 

In line with expectations, we observe in \autoref{fig:metrics-scatter} that only approaches using ClassThr or Full (the full label set) consistently achieve target coverage for both class labels and box intervals across all three datasets. Notably, while Full results in overly inflated interval widths due to its construction, ClassThr provides surprisingly efficient label sets (with \emph{mean set size} $\leq 4$) which propagate into reasonably tight box intervals. Differences also exist in the efficiency of the three bounding box methods, with Box-Ens performing notably better on BDD100k. Stratifying results by object (mis-)classification also confirms that wrongly classified objects tend to exhibit higher uncertainty (see \autoref{fig:size-misclassif}). Oracle relies on knowing the correct quantile and thus does not require generating a label set. While it provides nominal guarantees assuming correct label prediction and empirically satisfies them with high efficiency (\ie, small $MPIW$), the condition severely limits its practicality. Top consistently undercovers the true label, and also tends towards undercovering boxes as sample sizes increase. Naive is able to consistently maintain box coverage even though label coverage is violated, with surprisingly tight PIs. However, additional experiments in \autoref{app:calibration-ece} showcase its sensitivity to model miscalibration, yielding it less robust than ClassThr. Yet, it is interesting that methods such as Top and Naive perform quite well empirically, even under void nominal guarantees. 

\textbf{Discussion and practitioner's choice.} We end by discussing the choices in selecting a suitable conformal \emph{two-step} approach a practitioner may want to consider. In terms of label set methods, we suggest the following: if the model classifier records a high accuracy and there exists a way to externally validate predicted class labels, then Top may be a highly efficient choice. If the model is strongly calibrated and only empirical assurances are sufficient, then Naive may be a suitable selection. However, if strong safety assurances with both nominal and empirical guarantees are desired, then ClassThr is the only safe choice. We highlight that improvement potential remains: obtained PIs using ClassThr tend to overcover, presumably due to our conservative quantile selection strategy. It may be indeed possible to obtain the same set of safety assurances with higher efficiency under a different strategy. For example, one may consider a weighted quantile construction on the basis of the classifier's confusion matrix.

Regarding the choice of bounding box method, we observe that Box-Std will be the most efficient if the object detection task contains only similarly sized objects. However, if objects vary substantially in size, an adaptive approach such as Box-Ens or Box-CQR will be more suitable. One may also consider designing a conformal approach which explicitly satisfies target coverage across other reasonable strata beyond classes, such as object sizes or shapes. A limiting factor to consider may be the size of available calibration data per partition.

%% file: text/fig_violins.tex
\begin{figure*}[t]
    \centering
    \begin{subfigure}[t]{\textwidth}
        \raisebox{-\height}{
            \includegraphics[width=\textwidth]{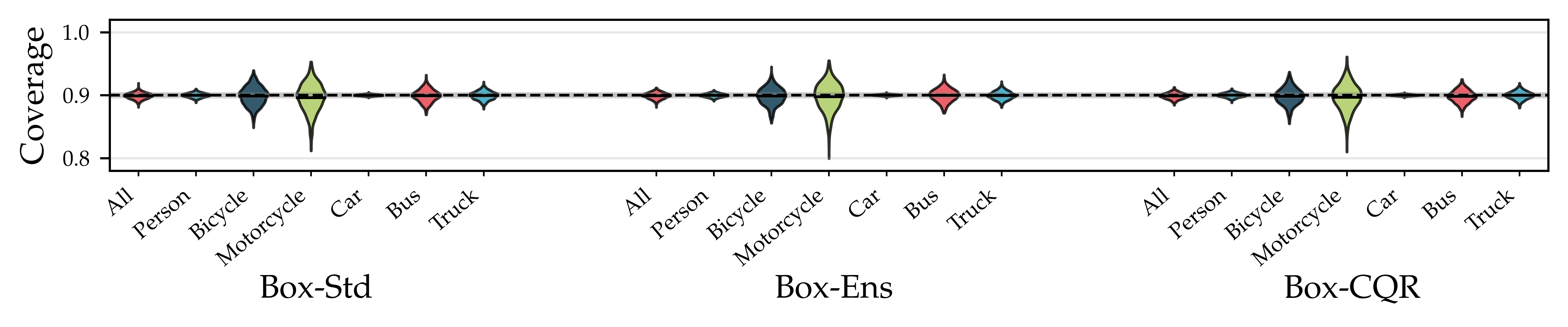}
        }
    \end{subfigure}
    \begin{subfigure}[t]{.72\textwidth}
        \raisebox{-\height}{
            \includegraphics[width=\textwidth]{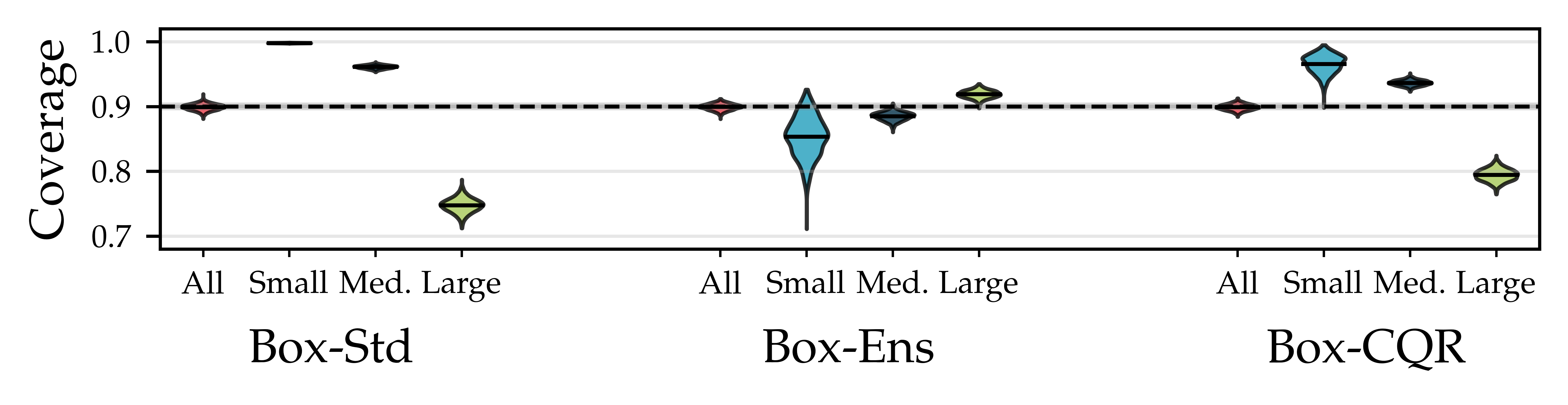}
        }
    \end{subfigure}%
    \hfill
    \begin{subfigure}[t]{.28\textwidth}
        \raisebox{-\height}{
            \includegraphics[width=\textwidth]{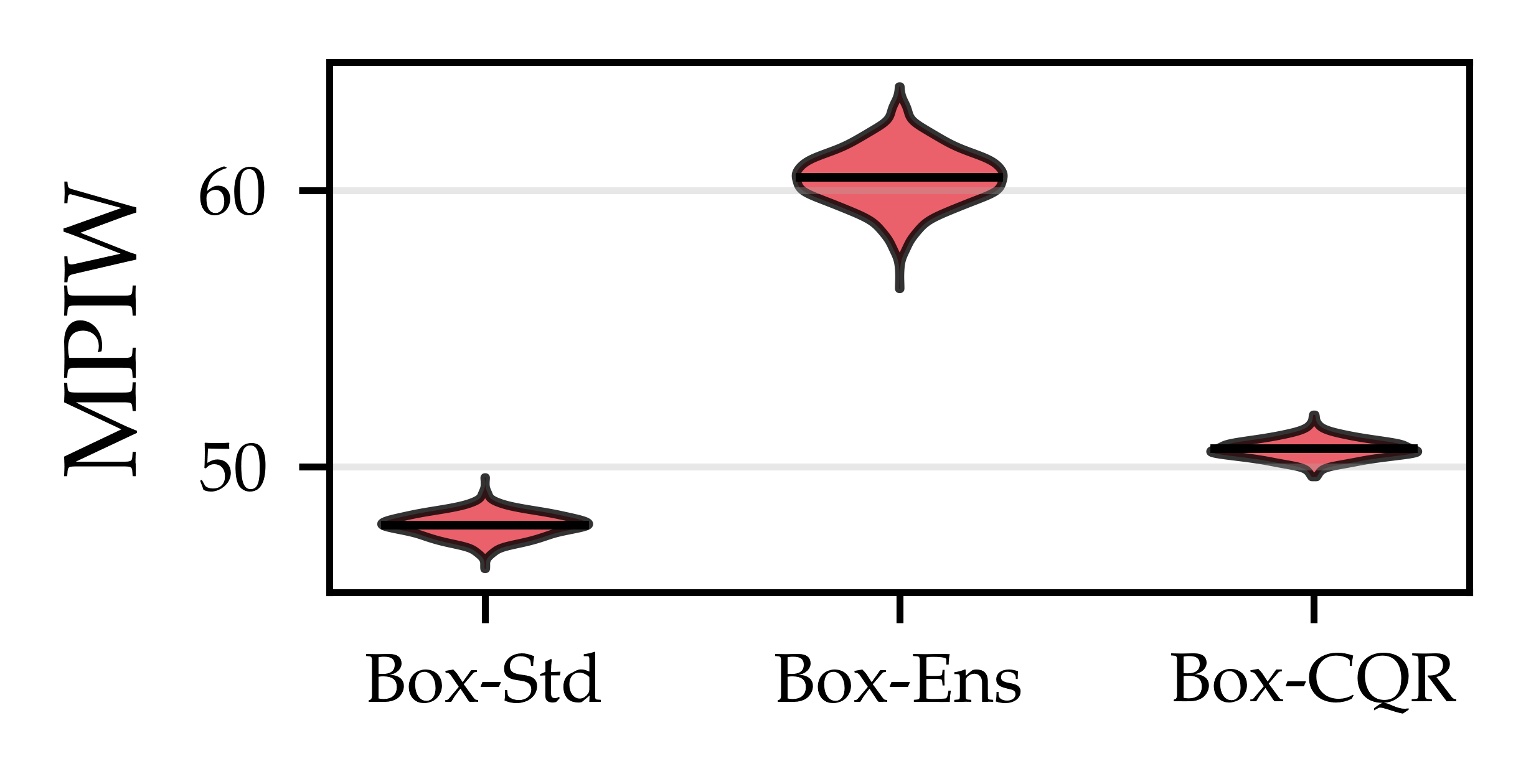}
        }
    \end{subfigure}

    \caption{\emph{Top:} Empirical coverage levels marginally across all objects (All) and across objects from selected classes for the three bounding box methods (\autoref{sec:methods_box}) on the BDD100k dataset. Target coverage is achieved both marginally and for individual classes. \emph{Bottom:} Coverage levels are stratified by object size (Small, Medium, Large), showing that Box-CQR and in particular Box-Ens provide a more balanced empirical coverage across sizes. However, this comes at the cost of slightly larger intervals, as seen when comparing $MPIW$. We also visualize target coverage (\includegraphics[width=2em]{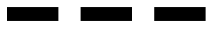}) and the marginal coverage distribution (\includegraphics[width=2em]{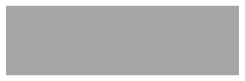}). Displayed densities are results obtained over 1000 trials.}
    \label{fig:violins}
\end{figure*}

%% file: text/table_baselines.tex
\begin{table}[t]
    \centering
    \small
    \setlength{\tabcolsep}{5pt}
    \resizebox{\textwidth}{!}{
    \begin{tabular}{llcccc}
         \multicolumn{1}{c}{\multirow{-1}{*}{}} & \multicolumn{1}{c}{\multirow{-1}{*}{}} & \multicolumn{2}{c}{\textbf{Two-sided box intervals}} & \multicolumn{2}{c}{\textbf{One-sided box intervals}} \\
         \cmidrule(lr){3-4}\cmidrule(lr){5-6}
         \multicolumn{1}{c}{\multirow{-1}{*}{\textbf{Uncertainty method}}} & \multicolumn{1}{c}{\multirow{-1}{*}{\textbf{Object detector}}} & $MPIW$ & $Cov$ & $MPIW$ & $Cov$ \\
         \toprule
         DeepEns & $5 \times$ Faster R-CNN & $12.31 \pm 0.47$ & \cellcolor{gray!25} $0.21 \pm 0.01$ & $74.15 \pm 2.01$ & \cellcolor{gray!25} $0.49 \pm 0.01$ \\
         GaussianYOLO & YOLOv3 & $7.00 \pm 0.14$ & \cellcolor{gray!25} $0.08 \pm 0.01$ & $87.07 \pm 4.25$ & \cellcolor{gray!25} $0.35 \pm 0.01$ \\
         \midrule
         \multirow{4}{*}{Andéol \etal (Best)} & Faster R-CNN & \multicolumn{2}{c}{N/A} & $87.62 \pm 1.79$ & $0.91 \pm 0.01$ \\
         & YOLOv3 & \multicolumn{2}{c}{N/A} & $107.93 \pm 4.85$ & $0.92 \pm 0.02$ \\
         & DETR & \multicolumn{2}{c}{N/A} & $82.21 \pm 1.64$ & $0.90 \pm 0.01$ \\
         & Sparse R-CNN & \multicolumn{2}{c}{N/A} & $79.35 \pm 1.78$ & $0.91 \pm 0.01$ \\
         \midrule
         \multirow{4}{*}{Box-Std (Ours)} & Faster R-CNN & $55.47 \pm 2.97$ & $0.88 \pm 0.02$ & $85.42 \pm 1.99$ & $0.88 \pm 0.02$ \\
         & YOLOv3 & $61.73 \pm 3.66$ & $0.88 \pm 0.02$ & $103.12 \pm 3.95$ & $0.88 \pm 0.02$ \\
         & DETR & $45.34 \pm 3.33$ & $0.88 \pm 0.02$ & $80.57 \pm 1.78$ & $0.88 \pm 0.01$ \\
         & Sparse R-CNN & $41.92 \pm 2.16$ & $0.89 \pm 0.01$ & $77.33 \pm 1.72$ & $0.89 \pm 0.01$ \\
         \bottomrule
         \\
    \end{tabular}
    }
    \caption{
    We compare our simplest method \textbf{Box-Std} to Andéol \etal's best results (see \autoref{app:baseline-impl}) across different object detectors (Faster R-CNN \cite{wu2019detectron2}, YOLOv3 \cite{redmon2018yolov3}, DETR \cite{carion2020detr}, Sparse R-CNN \cite{sun2021sparse}) as well as deep ensembles (DeepEns \cite{b.lakshminarayanan2017}) and GaussianYOLO \cite{choi2019odbayes}, two popular UQ approaches. The former satisfies coverage but is only designed for one-sided intervals, while the latter can heavily undercover in practice (marked \includegraphics[width=2em]{fig/caption_line2.png}). Results are for COCO across classes and 100 trials, for target coverage $(1-\alpha_B) = 0.9$. The key difference between various object detectors are the obtained interval widths ($MPIW$), which relate to predictive performance and are smaller for better models.
    } 
    \label{tab:baselines}
\end{table}

%% file: text/fig_metrics.tex
\begin{figure*}[t]
    \centering
    \begin{subfigure}[t]{0.6\textwidth}
        \centering
        \raisebox{-\height}{
            \includegraphics[width=\textwidth]{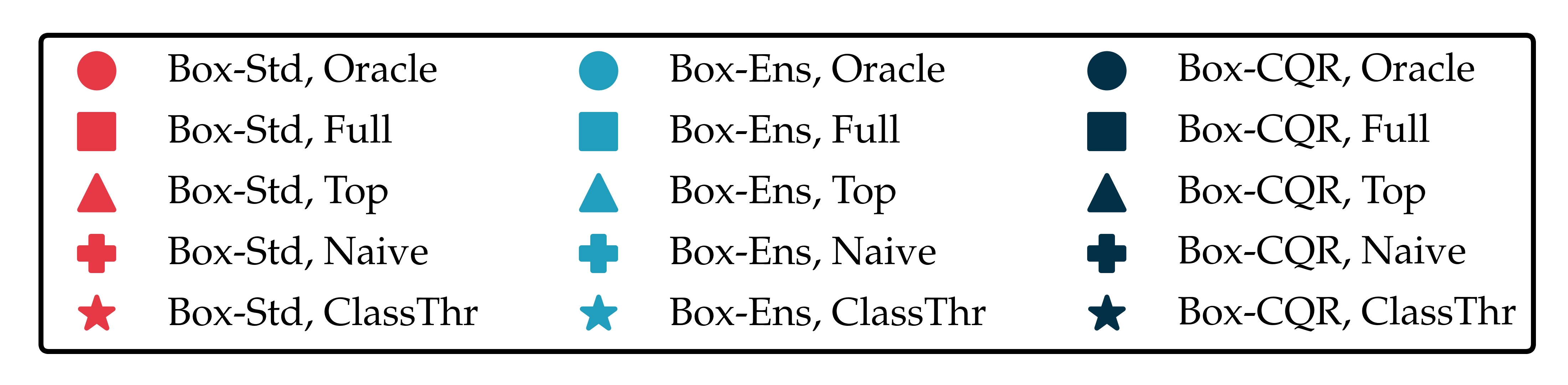}
        }
    \end{subfigure}
    \vspace{-2mm}
    \begin{subfigure}[t]{.48\textwidth}
        \raisebox{-\height}{
            \includegraphics[width=\textwidth]{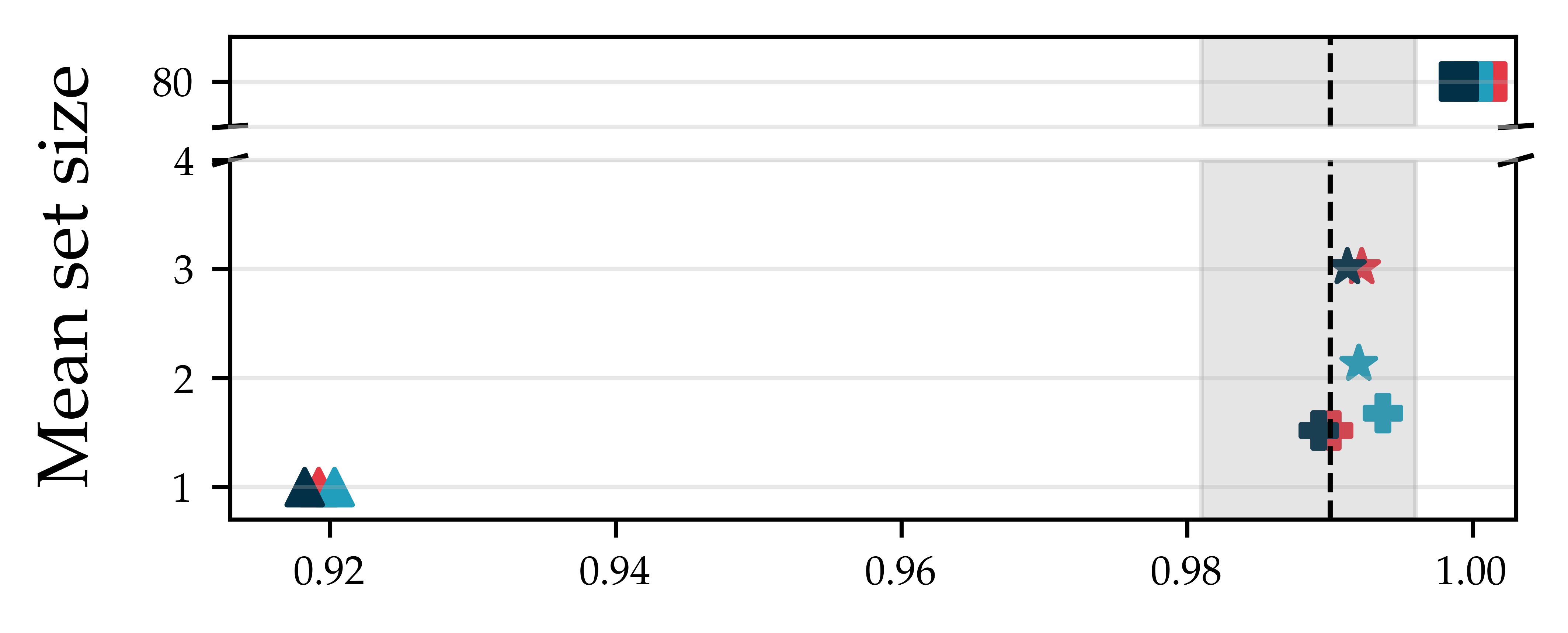}
        }
    \end{subfigure}
    \hfill
    \begin{subfigure}[t]{.48\textwidth}
        \raisebox{-\height}{
            \includegraphics[width=\textwidth]{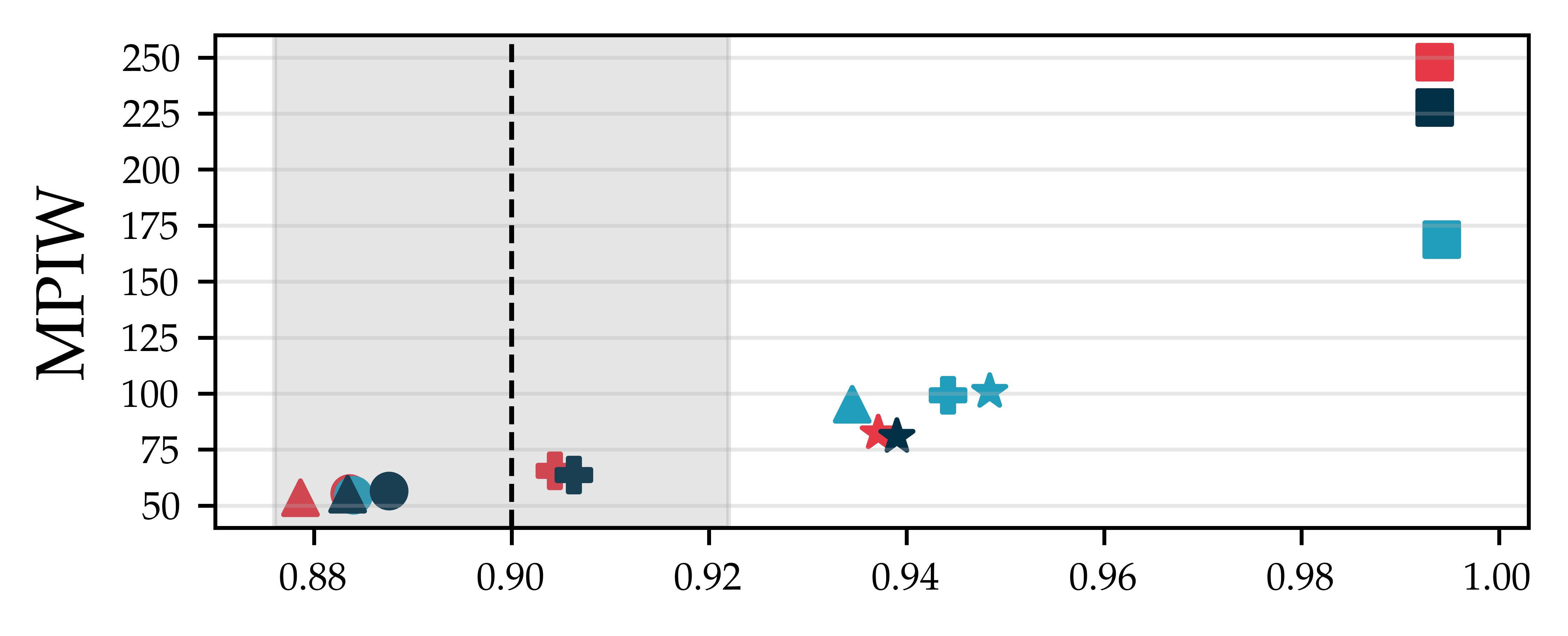}
        }
    \end{subfigure}
    \vspace{-2mm}
    \begin{subfigure}[t]{.48\textwidth}
        \raisebox{-\height}{
            \includegraphics[width=\textwidth]{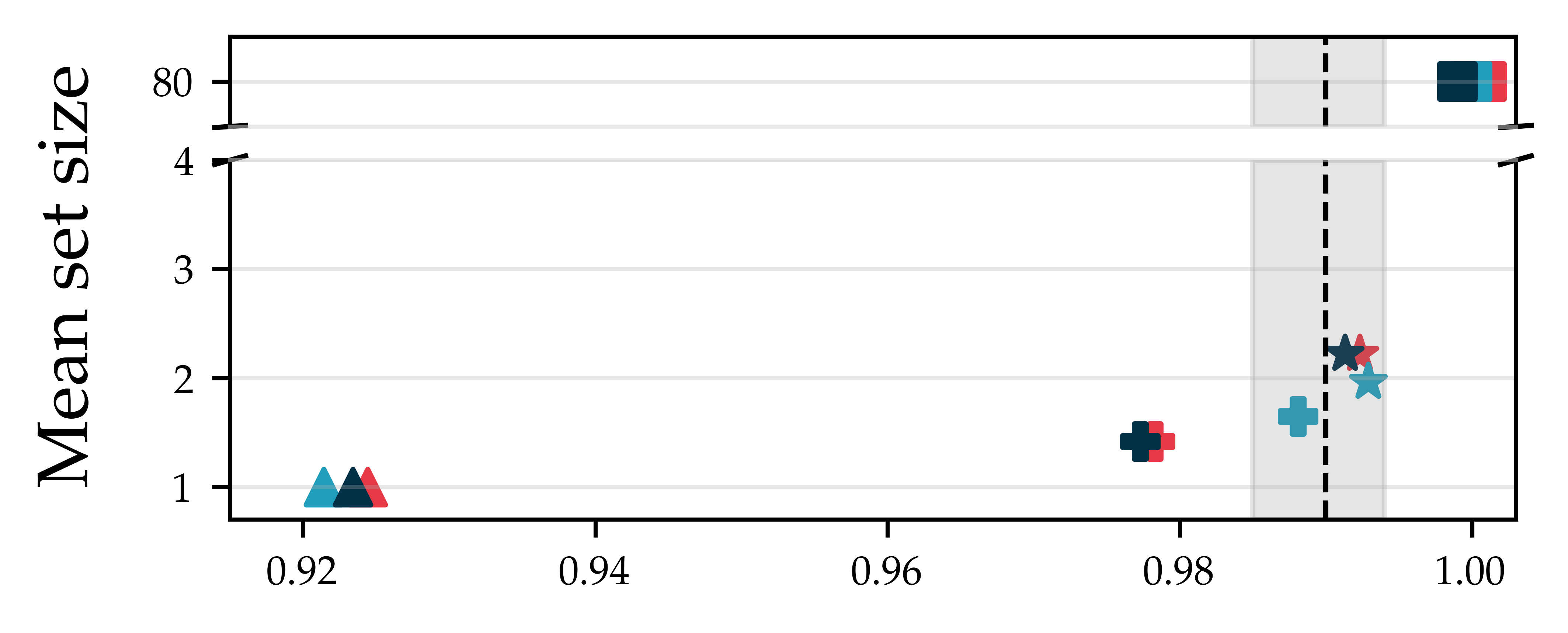}
        }
    \end{subfigure}
    \hfill
    \begin{subfigure}[t]{.48\textwidth}
        \raisebox{-\height}{
            \includegraphics[width=\textwidth]{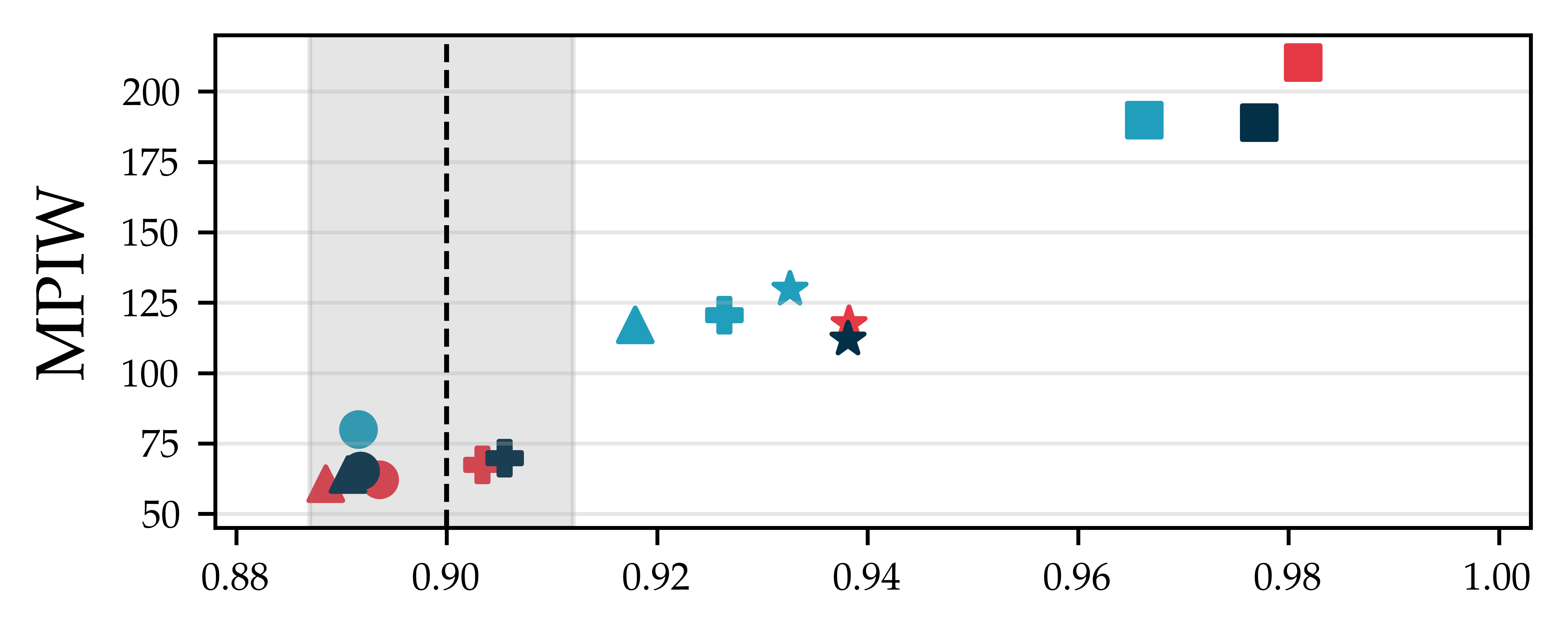}
        }
    \end{subfigure}
    \vspace{-2mm}
    \begin{subfigure}[t]{.48\textwidth}
        \raisebox{-\height}{
            \includegraphics[width=\textwidth]{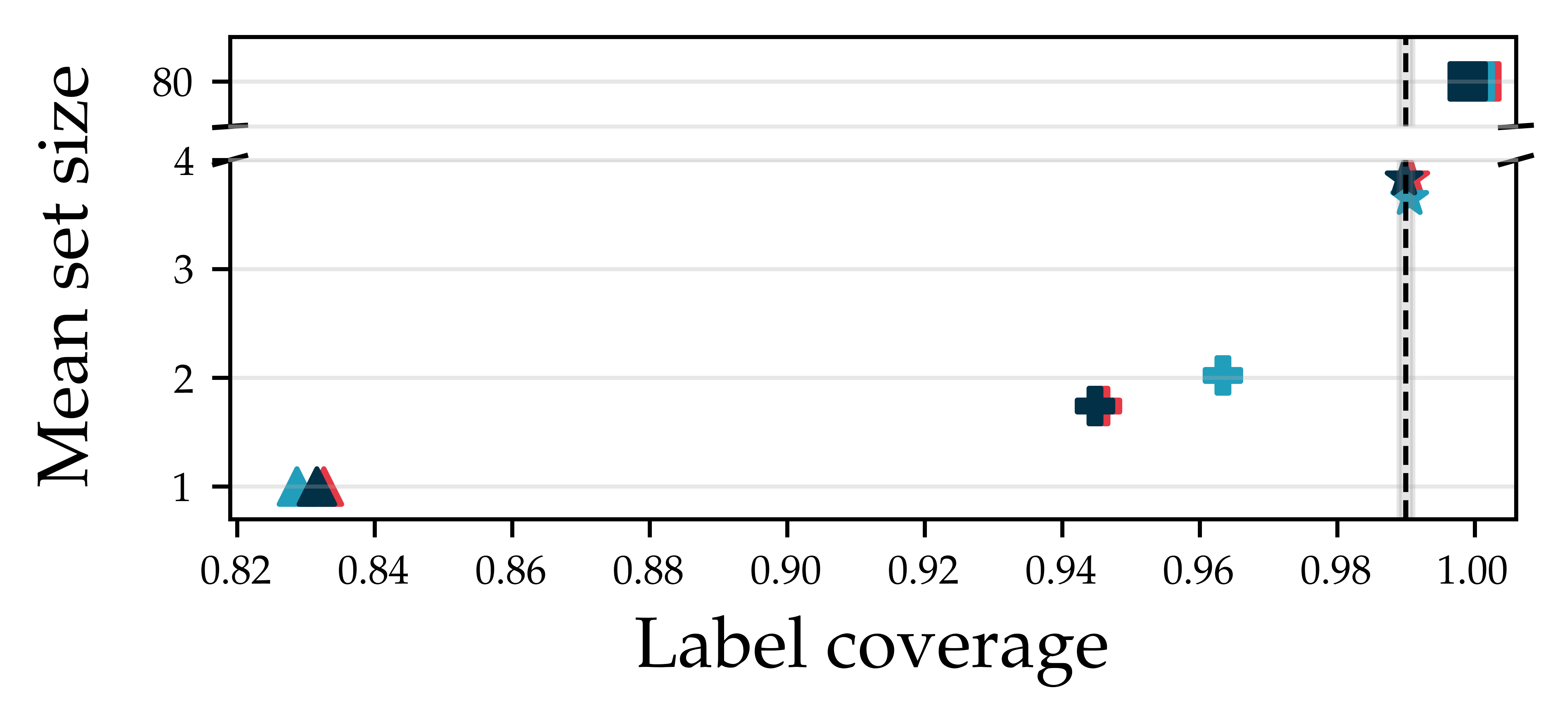}
        }
    \end{subfigure}
    \hfill
    \begin{subfigure}[t]{.48\textwidth}
        \raisebox{-\height}{
            \includegraphics[width=\textwidth]{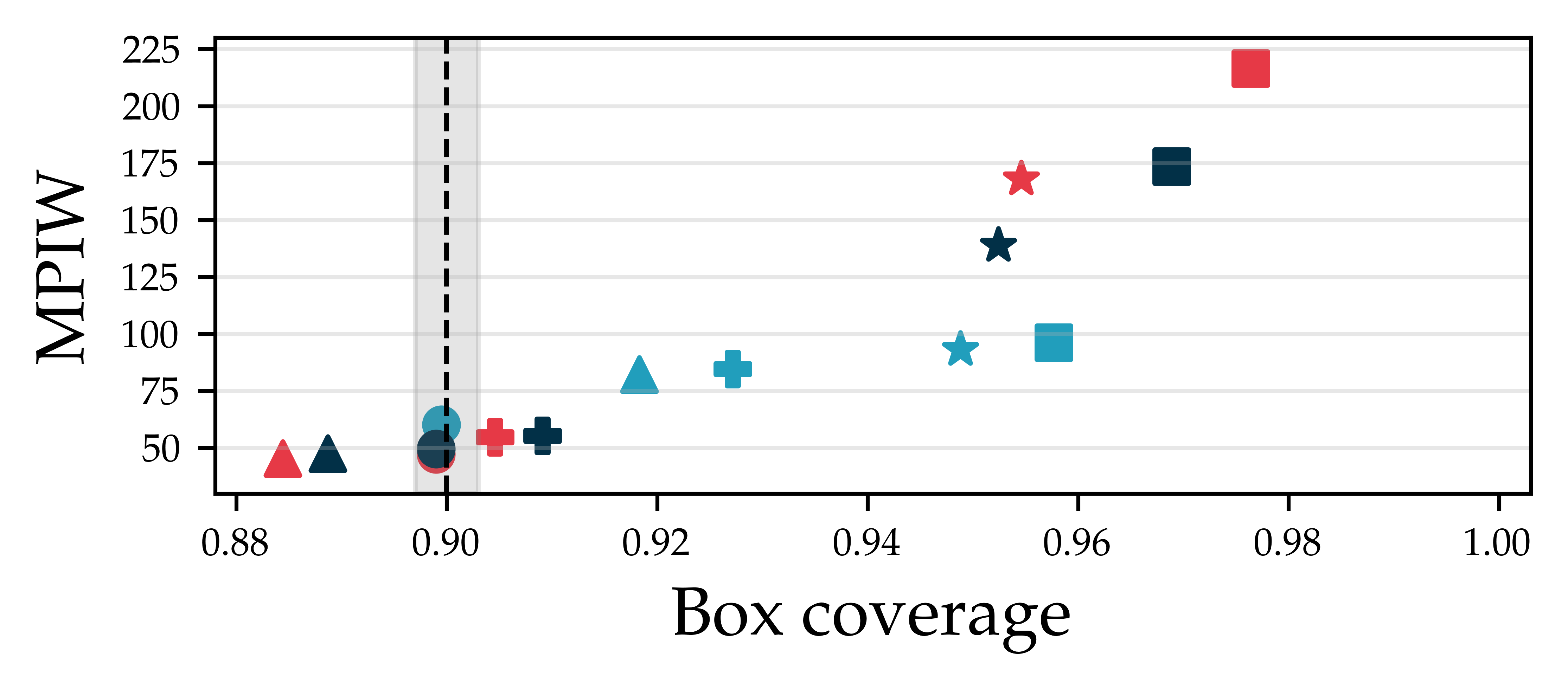}
        }
    \end{subfigure}

    \caption{Every combination of conformal label set and bounding box method is evaluated along two axes for COCO (\emph{top row}), Cityscapes (\emph{middle row}) and BDD100k (\emph{bottom row}). On the vertical axis we display efficiency, \ie, \emph{mean set size} for label sets (\emph{left column}) and $MPIW$ for box intervals (\emph{right column}). On the horizontal axis we display empirical coverage levels. We also draw target coverage (\includegraphics[width=2em]{fig/caption_line1.png}) and marginal coverage distributions (\includegraphics[width=2em]{fig/caption_line2.png}). In line with \autoref{tab:nominal_guarantees_check}, approaches employing ClassThr or Full consistently achieve both label and box target coverage, at the cost of larger prediction sets/intervals. Results are averaged across classes and 100 trials.}
    \label{fig:metrics-scatter}
\end{figure*}

%% file: text/conclusion.tex
\section{Conclusion}
\label{sec:conclusion}

We present a novel procedure to quantify predictive uncertainty for multi-object detection. We leverage CP to generate uncertainty intervals with a per-class coverage guarantee for new samples. Our proposed \emph{two-step} conformal approach provides adaptive bounding box intervals with safety assurances robust to object misclassification. Addressing similar types of guarantees, the procedure can be extended to 3D bounding boxes, object tracking and other detection tasks in future work. Whilst improvements can be made to achieve even narrower intervals, our results are promisingly tight, paving the way for a safer deployment of vision-based systems in scenarios involving decision-making under uncertainty.

%% file: text/X_appendix.tex
\section{Mathematical and algorithmic details}
\label{app:math}

\subsection{Split conformal prediction}
\label{app:split-cp}

The general conformal procedure for split conformal prediction \cite{angelopoulos2023gentle, h.papadopoulos2007, v.vovk2005, g.shafer2008} is provided in Algorithm \ref{algo:split-cp} below.

\begin{algorithm}
\caption{Split conformal prediction}
\label{algo:split-cp}
\begin{algorithmic}[1]
    
    \State \textbf{Input:} data $\mathcal{D} \subset \mathcal{X} \times \mathcal{Y}$, prediction algorithm $\mathcal{A}$, miscoverage rate $\alpha \in (0,1)$.
    
    \State \textbf{Output:} Prediction set/interval $\hat{C}(X_{n+1})$ for test sample $(X_{n+1},Y_{n+1})$.
    
    \State \textbf{Procedure:} 
    
    \State Split data $\mathcal{D}$ into two disjoint subsets: a proper training set $\mathcal{D}_{train}$ and calibration set $\mathcal{D}_{cal} = \{(X_i,Y_i)\}_{i=1}^{n}$.
    
    \State Fit a prediction model on the proper training set: 
    
    $\hat{f}(\cdot) \gets \mathcal{A}(\mathcal{D}_{train})$.
    
    \State Define a scoring function $s:\mathcal{X}\times\mathcal{Y}\rightarrow\mathbb{R}$ applied to $\mathcal{D}_{cal}$, resulting in a set of nonconformity scores 
    
    $S = \{s(\hat{f}(X_i),Y_i)\}_{i=1}^{n} = \{s_i\}_{i=1}^{n}$. 

    \noindent A particular score $s_i$ encodes a notion of dissimilarity (nonconformity) between the predicted value $\hat{f}(X_i)$ and true value $Y_i$.
    
    \State Compute a conformal quantile $\hat{q}$, defined as the 
    
    $\lceil (n+1)(1-\alpha)/n \rceil$-th empirical quantile of $S$.

    \noindent Under exchangeability of $\mathcal{D}_{cal} \cup \{(X_{n+1}, Y_{n+1})\}$, the conformal quantile $\hat{q}$ is a finite sample-corrected quantile ensuring target coverage $(1-\alpha)$ by its construction.
    
    \State For a new test sample $(X_{n+1},Y_{n+1})$, a valid prediction set/interval for $X_{n+1}$ is given by 

    $\hat{C}(X_{n+1})=\{y\in\mathcal{Y}:s(\hat{f}(X_{n+1}),y) \le \hat{q}\}$.

    \noindent Validity refers to satisfying sample coverage with marginal probability $(1-\alpha)$, \ie, the guarantee in \autoref{eq:conf-guarant}. We refer to the relevant literature for the proof.
    
    \State \textbf{End procedure}
\end{algorithmic}
\end{algorithm}

\noindent Rather than a marginal guarantee, we desire the class-conditional guarantee of \autoref{eq:class-cond-validity}, which is re-interpreted as \autoref{eq:class-cond-validity-label} for our object detection setting outlined in \autoref{sec:background}. Thus, we may relate our problem setting to Algorithm \ref{algo:split-cp} as follows:
\begin{itemize}
    \item[$\bullet$] A calibration sample $(X_i, Y_i)$ consists of an input image $X_i$ and an existing object in that image, parametrized as $(c^{1}_i, \dots, c^{m}_i, l_i)$ for an arbitrary amount of coordinates $m$ following \autoref{subsec:background_conf_od}. As we run Algorithm \ref{algo:split-cp} on a per-class basis, the class label $l$ for samples within a class is fixed, and we thus define $Y_i := (c^{1}_i, \dots, c^{m}_i)$ as the object's coordinates.
    \item[$\bullet$] Since Algorithm \ref{algo:split-cp} is applied to samples within each class, the available calibration set $\mathcal{D}_{cal}$ is partitioned into smaller sets of samples matching a specific class label, \ie, we obtain sets $$ \mathcal{D}_{cal, y} = \{ (X_i,Y_i) \in \mathcal{D}_{cal}: l_i = y \}, \forall y \in \mathcal{Y}. $$
    \item[$\bullet$] Finally, the conformal procedure is performed for each of the $m$ coordinates in parallel, therefore necessating a multiple testing correction as motivated in \autoref{subsec:methods_mht} and \autoref{app:mult-testing-problem}. Thus, we define the final set of calibration samples for the $k$-th coordinate of an object belonging to class $y$ as $$ \mathcal{D}_{cal, y, k} = \{ (X_i, c^{k}_i): (X_i, Y_i) \in \mathcal{D}_{cal, y} \wedge c^{k}_i \in Y_i \}. $$
\end{itemize}
In line with our notation in \autoref{sec:methods_box}, the required values to compute a nonconformity score for $(X_i, c^{k}_i) \in \mathcal{D}_{cal, y, k}$ are the true $k$-th coordinate $c^{k}_i$ and the model's prediction $\hat{f}_{B}(X_i) = \hat{c}^{k}_i$; the computed conformal quantile for $\mathcal{D}_{cal, y, k}$ is denoted $\hat{q}_{B}^{k}$; and the obtained prediction interval for a test sample's $k$-th coordinate $(X_{n+1}, c^{k}_{n+1})$ is denoted $\hat{C}^{k}_{B}(X_{n+1})$. The miscoverage rate becomes $\alpha_B$.

Finally, we note that assuming exchangeability of $\mathcal{D}_{cal} \cup \{(X_{n+1}, Y_{n+1})\}$ also implies assuming exchangeability holds for any subsets of $D_{cal}$, such as the considered partitions $D_{cal, y, k} \subset D_{cal, y} \subset D_{cal}$.

\subsection{Conformal class thresholding (ClassThr)}
\label{app:classthr-cp}

The procedure to produce class label predictions sets satisfying a coverage guarantee via conformal class thresholding (ClassThr) \cite{m.sadinle2019a} is outlined in Algorithm \ref{algo:label-cp}. Note that we directly sketch the algorithm for its class-conditional variation which satisfies \autoref{eq:class-cond-validity-label}. We also invert the scores used by Sadinle \etal~\cite{m.sadinle2019a} to better reflect the notion of nonconformity, while being equivalent in procedure.

\begin{algorithm}
\caption{Conformal class thresholding (ClassThr)}
\label{algo:label-cp}
\begin{algorithmic}[1]
    
    \State \textbf{Input:} data $\mathcal{D} \subset \mathcal{X} \times \mathcal{Y}$, prediction algorithm $\mathcal{A}$, miscoverage rate $\alpha_L \in (0,1)$.
    
    \State \textbf{Output:} Prediction set $\hat{C}_{L}(X_{n+1})$ for test sample $(X_{n+1},Y_{n+1})$.
    
    \State \textbf{Procedure:} 
    
    \State Split data $\mathcal{D}$ into two disjoint subsets: a proper training set $\mathcal{D}_{train}$ and calibration set $\mathcal{D}_{cal} = \{(X_i,Y_i)\}_{i=1}^{n}$.
    
    \State Fit a probabilistic classifier on the proper training set: 
    
    $\hat{f}_{L}(\cdot) \gets \mathcal{A}(\mathcal{D}_{train})$.

    \State Further partition $\mathcal{D}_{cal}$ into smaller sets of samples matching a specific class label, \ie, obtain sets 
    
    $\mathcal{D}_{cal, y} = \{ (X_i,Y_i) \in \mathcal{D}_{cal}: Y_i = y \}, \forall y \in \mathcal{Y}$.
    
    \State Define the scoring function 
    
    $s:\mathcal{X} \times \mathcal{Y} \rightarrow [0,1], \, (\hat{f}_L(X), y) \mapsto 1 - \hat{\pi}_{y}(X)$ 

    \noindent for some object in image $X$ belonging to class $y$. That is, compute the object's estimated true class probability $\hat{\pi}_{y}(X)$. Its complement encodes a notion of dissimilarity (nonconformity) between the predicted and true class probabilities.

    \State Define the set of class quantiles $Q = \emptyset$.

    \State \textbf{Begin for} $y \in \mathcal{Y}$:
    
    \State Apply $s$ to $\mathcal{D}_{cal, y}$ to obtain a set of scores 
    
    $S_y = \{s(\hat{f}(X_i),Y_i)\}_{i=1}^{n} = \{s_i\}_{i=1}^{n}$. 
    
    \State Compute a conformal quantile $\hat{q}_{L}^y$, defined as the 
    
    $\lceil (n+1)(1-\alpha_L)/n \rceil$-th empirical quantile of $S_y$.

    \noindent Under exchangeability of $\mathcal{D}_{cal} \cup \{(X_{n+1}, Y_{n+1})\}$, the conformal quantile $\hat{q}_{L}^y$ is a finite sample-corrected quantile ensuring target coverage $(1-\alpha_L)$ by its construction.

    \State Add the quantile to the set: $Q = Q \cup \{ \hat{q}_{L}^y \}$.

    \State \textbf{End for}
    
    \State For a new test sample $(X_{n+1},Y_{n+1})$, a valid prediction set for $X_{n+1}$ is given by 

    $\hat{C}_{L}(X_{n+1}) = \{y \in \mathcal{Y}: \hat{\pi}_{y}(X_{n+1}) \ge 1 - \hat{q}_{L}^y\},$

    \noindent where $\hat{q}_{L}^y \in Q$. Validity refers to satisfying sample coverage with probability $(1-\alpha_L)$ per class, \ie, the guarantee in \autoref{eq:class-cond-validity-label}. We refer to Sadinle \etal~\cite{m.sadinle2019a} for a proof.
    
    \State \textbf{End procedure}
\end{algorithmic}
\end{algorithm}

Since Algorithm \ref{algo:label-cp} is used to obtain prediction sets for class labels, we now define a calibration sample $(X_i, Y_i)$ as consisting of an input image $X_i$ and the class label of an existing object in that image, \ie, we have $Y_i := l_i$ following the object parametrization of \autoref{subsec:background_od}. Similarly, we have $\hat{C}_{L}(X_{n+1}) \subseteq \mathcal{Y} := \{1, \dots, K\}$ for a given classification task with $K$ classes. 

Note that the procedure in Algorithm \ref{algo:label-cp} may lead to empty sets $\hat{C}_{L}(X_{n+1}) = \emptyset$ in some specific cases. We alleviate any empty sets by including the highest probability label, \ie, creating a singleton set according to our baseline Top: 
$$ \hat{C}_{L}(X_{n+1}) = \{y^*: \hat{\pi}_{y^*}(X_{n+1}) = \max_{y \in \{1,\dots,K\}} \,\hat{\pi}_{y}(X_{n+1})\}, $$ 
and thus $|\hat{C}_{L}(X_{n+1})| = 1$. This heuristic maintains validity -- the guarantee already holds for empty sets, so enlargening any prediction sets does not invalidate it. Sadinle \etal~\cite{m.sadinle2019a} also propose an `Accretive Completion' algorithm to address this issue. The approach iteratively reduces the quantile magnitudes on the basis of minimizing ambiguity, filling in the null regions such that ambiguous feature spaces are better reflected. Since error control is our only concern, we opt for the simpler top-class heuristic.

Finally, ClassThr may also be motivated from a theoretical perspective. Given a perfectly calibrated classifier such that $\hat{\pi}_{y}(X) = \pi_{y}(X) \,\,\forall y \in \mathcal{Y}$, the approach can be shown to perform optimally in terms of efficiency, \ie, it produces the smallest \emph{mean set size} (Sadinle \etal~\cite{m.sadinle2019a}, Thm. 1).

\subsection{Multiple testing problem}
\label{app:mult-testing-problem}

Conformal prediction can also be interpreted from a hypothesis testing perspective \cite{v.vovk2005, shi2013applications}. A null hypothesis for the $k$-th test sample coordinate is formed as $H_0: c^k_{n+1} = c^k$ for some candidate value $c^k \in \mathbb{R}$. Leveraging the nonconformity scores of the calibration samples as an empirical null distribution, a p-value $p(c^k)$ is computed as the fraction of samples which conform worse than $s(\hat{c}^k_{n+1}, c^k)$ \footnote{A valid p-value $P$ is defined as $\mathbb{P}(P \le \alpha) \le \alpha \quad \forall \alpha \in (0,1).$}. The value $c^k$ is included in the prediction set --- \ie, we cannot reject $H_0$ --- if $p(c^k) > \alpha_B$. Conformalizing a set of $m$ coordinates separately and thus running Algorithm \ref{algo:split-cp} in parallel $m$ times with a \emph{global} miscoverage rate $\alpha_B$ leads to the need for a corrected miscoverage rate on the individual level, as we show next.

Recall that for a test sample $(X_{n+1}, Y_{n+1})$ we have $Y_{n+1} = (c_{n+1}^1, \dots, c_{n+1}^m) \in \mathbb{R}^m$, and for its $k$-th coordinate $c_{n+1}^k$ the value is included in the prediction set $\hat{C}^{k}_{B}(X_{n+1})$ if $p(c_{n+1}^k) > \alpha_B$. Equivalently, we may state that $c_{n+1}^k \notin \hat{C}^{k}_{B}(X_{n+1})$ if $p(c_{n+1}^k) \leq \alpha_B$.

Starting from \autoref{eq:class-cond-validity-od} for a fixed class $y$, we then have that
\begin{equation}
    \begin{split}
        \mathbb{P}(\bigcap_{k=1}^{m}(c_{n+1}^k &\in \hat{C}^{k}(X_{n+1}))) \\
        &= 1-\mathbb{P}(\bigcup_{k=1}^{m}(c_{n+1}^k \notin \hat{C}^{k}(X_{n+1})))\\
        &\geq 1-\sum_{k=1}^{m}\mathbb{P}(c_{n+1}^k \notin \hat{C}^{k}(X_{n+1})) \\
        &= 1-\sum_{k=1}^{m}\mathbb{P}(p(c_{n+1}^k) \le \alpha_B) \\
        &\geq 1-\sum_{k=1}^{m}\alpha_B
        = 1-m \alpha_B.
    \end{split}
    \label{eq:mht-problem}
\end{equation}
We observe that target coverage cannot be guaranteed globally since $1-m\alpha_B \le 1-\alpha_B$ for any $m \in \mathbb{N}^+$. The Bonferroni correction naively selects $\alpha'_B = \alpha_{B}/m$ to ensure $1-m\alpha'_B =  1-\alpha_B$.

\subsection{Multiple testing correction via $\max$-rank}
\label{app:max-rank}

We leverage the $\max$-rank algorithm of Timans \etal~\cite{timans2023maxrank} to correct for multiple testing. Their procedure ensures family-wise error rate control while retaining high statistical power in settings of positively dependent hypothesis tests. For our setting, this translates to ensuring a guarantee at global coverage level $(1-\alpha_B)$ for the full bounding box, whilst surpressing any overcoverage behaviour, \ie, providing maximally tight quantiles for each box coordinate. Again, we harness our knowledge on the positive dependency of obtained conformal p-values \cite{s.bates2022, y.benjamini2001}, thus providing tight prediction intervals whilst maintaining coverage. In particular, Timans \etal~\cite{timans2023maxrank} show that $\max$-rank both theoretically and empirically improves over Bonferroni.

The suggested correction is an adaptation of the Westfall \& Young~\cite{westfall1993resampling} \emph{min-P} permutation correction. Intuitively, $\max$-rank collapses multiple testing dimensions into a single hypothesis test, using a composite empirical null distribution consisting of the maximum rank statistics across testing dimensions. The computed quantile over these maximum rank statistics will then ensure global coverage by design, since it controls coverage over `worst-case' ranks. Timans \etal~\cite{timans2023maxrank} crucially relate their procedure to the hypothesis testing perspective of conformal prediction. In such a setting, each testing dimension's empirical null distribution corresponds to the ranks of values in the computed sets of nonconformity scores $S$, \ie, for each of our box coordinates. It is shown that $\max$-rank satisfies two key requirements for its integration into a CP procedure -- it preserves both exchangeability and validity. These conditions are demonstrated for the two key operations of the procedure, which are 1) operating in the rank domain of nonconformity scores, and 2) applying the $\max$-operator over these ranks. We refer to their work for further details.

\subsection{Bounding box quantile selection}
\label{app:box-quant-selection}

In \autoref{subsec:methods_label_downstream} we argue that a straight-forward quantile selection strategy is $\hat{q}_{B}^k = \max \{\hat{q}_{B}^{k,y}\}_{y \in \hat{C}_{L}(X_{n+1})}$, where $\hat{q}_{B}^{k,y}$ is the conformal quantile of the $k$-th box coordinate for class $y$. Then, for the $k$-th coordinate of a test sample whose true class label is indeed $y$ -- which we ensure is contained in $\hat{C}_{L}(X_{n+1})$ with probability $(1-\alpha_L) \approx 1$ -- we have that $\hat{q}_{B}^k \geq \hat{q}_{B}^{k,y}$ and thus $\mathbb{P}(s(\hat{c}^k_{n+1}, c^k_{n+1}) \leq \hat{q}_{B}^k) \geq \mathbb{P}(s(\hat{c}^k_{n+1}, c^k_{n+1}) \leq \hat{q}_{B}^{k,y})$. That is, using $\hat{q}_{B}^k$ as a thresholding quantile ensures a higher chance of score conformity, and thus is more likely to decide on inclusion into the prediction interval, at the cost of potential overcoverage. 

From a hypothesis testing perspective, rather than using the p-value $p(c^{k,y})$, we base our testing decision on inclusion using $p(c^k) = \max \{p(c^{k,y})\}_{y \in \hat{C}_{L}(X_{n+1})}$, itself a valid but conservative p-value \cite{vovk2022admissible}. Similarly, we then have that $p(c^k) \geq p(c^{k,y})$ and thus $\mathbb{P}(p(c^k) > \alpha_B) \geq \mathbb{P}(p(c^{k,y}) > \alpha_B)$, producing a higher chance of inclusion. The coverage difference between testing with $p(c^k)$ and $p(c^{k,y})$ then marks the incurred cost in terms of inefficiency, \ie, prediction interval width.

\input{text/X_fig_cov}

\subsection{Empirical coverage distribution}
\label{app:emp-cov-distr}

As mentioned in \autoref{subsec:exp_metrics}, empirical coverage $Cov$ is a random quantity parametrized by a coverage distribution. For a specific randomly sampled calibration set $\mathcal{D}_{cal}$, achieved coverage may deviate from the target coverage level $(1-\alpha)$ based on factors such as calibration set size. Specifically, the exact analytical form of the coverage distribution is given by
\begin{equation}
    \mathbb{P}(Y_{n+1} \in \hat{C}(X_{n+1}) | \mathcal{D}_{cal}) \sim \text{Beta($n+1-l, l$)},
\label{eq:cov-distr-beta}
\end{equation}
where $l= \left\lfloor (n+1)\alpha \right\rfloor$ and $n = |\mathcal{D}_{cal}|$ \cite{v.vovk2012, angelopoulos2023gentle}. The spread in empirical coverage is then approx. proportional to $n^{-1/2}$, and will thus shrink as the calibration set size increases. For example, given $\alpha=0.1$ and $n=1000$ we may expect empirical coverage to reasonably fall anywhere in the range of $90 \pm 2 \%$, while for $n=10\,000$ we have $90 \pm 0.5 \%$. 

In \autoref{fig:emp-cov} we display the exact coverage distributions for our three considered datasets on the basis of their calibration set sizes for both label (left) and box (right) target coverage rates. We additionally mark the $1\%$ and $99\%$ quantiles of respective distributions. We see that obtainable empirical coverage levels are in line with our experiments, where results are aggregated over multiple trials of calibration set sampling. For example, in \autoref{fig:metrics-scatter}, where we shade the coverage distribution between the above quantiles in grey (\includegraphics[width=2em]{fig/caption_line2.png}), obtained coverage levels using the Oracle method are within reasonable coverage bounds, even if on the lower end. Similarly, the magnitude of variations in coverage displayed in the violin plots of \autoref{fig:violins} relates directly to respective calibration set sizes. This also holds for any additional results in \autoref{app:results}.

\section{Implementation details}
\label{app:implement}

\subsection{Conformal ensemble (Box-Ens)}
\label{app:ens}

For Box-Ens, we select a number of pre-trained object detectors from \texttt{detectron2} to form a model ensemble. Specifically, we select five object detectors with similar predictive performance but varying model architectures \footnote{we choose models $\{$\texttt{X101-FPN}, \texttt{R101-FPN}, \texttt{R101-DC5}, \texttt{R50-DC5}, \texttt{R50-FPN}$\}$, see \url{https://github.com/facebookresearch/detectron2/blob/main/MODEL_ZOO.md}}.

The conformal scores for Box-Ens are defined as $s(\hat{f}_{B}(X), Y) = |\hat{c}^k - c^k|/\hat{\sigma}(X)$, where $\hat{\sigma}$ is some choice for a heuristic uncertainty estimate. Note that `heuristic' refers to obtained estimates without any nominal guarantee or assurance on quality. Following the deep ensemble approach \cite{b.lakshminarayanan2017}, we quantify $\hat{\sigma}$ as the standard deviation over box coordinate predictions, and produce a joint prediction $\hat{c}^k$ via weighted box fusion \cite{r.solovyev2021}. We detail each computation in the following.

Consider $c^{k}_i$ the $k$-th true coordinate of an $i$-th calibration sample to be predicted. For an ensemble of size $T$, each ensemble member produces a coordinate prediction $\hat{c}^k_{i,t}$ and a sample-level confidence score $\hat{s}_{i,t}$ (\ie, a confidence for the overall object prediction). The fused coordinate prediction $\hat{c}^k_i$ and uncertainty estimate $\hat{\sigma}_i$ are then computed as 
\begin{equation}
    \hat{c}^k_i= \frac{\sum_{t=1}^{T}\hat{s}_{i,t}\,\hat{c}^k_{i,t}}{\sum_{t=1}^{T}\hat{s}_{i,t}} \,\,\text{ and }\,\, \hat{\sigma}_i = \sqrt{\frac{1}{T}\sum_{t=1}^{T}(\hat{c}^k_{i,t} - \bar{c}^k_{i})^2},
\label{eq:box-fusion}
\end{equation}
where $\bar{c}^k_{i,t}$ is the equivalent of $\hat{c}^k_i$ with equal weights $\hat{s}_{i,t} \, \forall t=1,\dots, T$. 

\subsection{Conformal quantile regression (Box-CQR)}
\label{app:cqr}

We modify an object detection model to regress to estimated conditional quantiles of the bounding box coordinates alongside a standard mean prediction. This is achieved by supplementing the model's final regression output layer with additional box prediction heads, freezing all pre-trained weights, and training the additional heads with a \emph{quantile loss} function, also called \emph{pinball loss} \cite{steinwart2011estimating, r.koenker1978}. 

The loss for some quantile estimator $\hat{Q}_{\tau}$ of the $\tau$-th quantile is given by
\begin{equation}
    \mathcal{L}(y, \hat{Q}_{\tau}) = 
    \begin{cases}
        \tau\,(y - \hat{Q}_{\tau}(x)) & \text{if } y - \hat{Q}_{\tau}(x) > 0 \\
        (1 - \tau)\,(\hat{Q}_{\tau}(x) - y) & \text{else.}
    \end{cases}
\end{equation}
It intuitively penalizes both under- and overcoverage weighted by the target quantile $\tau$, and recovers the $L1$-loss for $\tau=0.5$. Since the box heads are architecturally independent, we can train arbitrary many quantile estimators in parallel, where we obtain an individual loss $\mathcal{L}(y, \hat{Q}_{\tau})$ for each $\tau$. The final loss for model updating is the sum of all individual quantile losses. 

For CQR, we require only lower and upper quantiles $\tau_{1}$ and $\tau_{2}$. If we aim for target coverage $(1-\alpha_B)$, a reasonable choice is $\tau_{1} = \alpha_B/2$ and $\tau_{2} = 1 - \alpha_B/2$, since the obtained interval $[\hat{Q}_{\alpha_B/2}, \hat{Q}_{1-\alpha_B/2}]$ will asymptotically achieve target coverage \cite{r.koenker1978, chaudhuri1991global}. However, in practice we require further interval scaling via CQR to obtain valid coverage in finite samples. Note that the choices for $\tau_{1}$ and $\tau_{2}$ are a modelling decision, and can in fact be tuned to produce more efficient PIs without invalidating the conformal coverage guarantee \cite{y.romano2019}. However, we only consider the simple setting with $\tau_{1} = \alpha_B/2$ and $\tau_{2} = 1 - \alpha_B/2$. For a desired target coverage of 90\%, these correspond to $\tau_{1} = 0.05$ and $\tau_{2} = 0.95$.

\subsection{Dataset splits and class mappings}
\label{app:dataset}

We display the distribution of objects per class for the selected set of classes in \autoref{tab:datasets-ist}. Objects are assigned to either calibration or test data based on the assignment to either split for the respective image they belong to. We randomly split the images according to the following calibration fractions of total available data for each dataset: 50\% for COCO, 50\% for Cityscapes and 70\% for BDD100k.

\input{text/X_tab_data}

In \autoref{tab:datasets-ist-by-size} we additionally display object counts in \% stratified by size, as measured via box surface area: small (area $\le 32^2$), medium (area $\in (32^2, 96^2]$) and large (area $> 96^2$). While Cityscapes contains virtually no small objects, we generally observe a mix of different object sizes for various classes. 

\textbf{Class mappings.} Our pre-trained object detection models are trained on the COCO training split and recognize all 80 COCO object classes. In order to permit the use of pre-trained models without further finetuning, as well as find a common intersection of classes across all three datasets, we map relevant classes with available annotations from Cityscapes and BDD100k to equivalent COCO classes. For the considered set of classes $\{$ person, bicycle, motorcycle, car, bus, truck $\}$, we find one-to-one correspondences for most classes. We additionally perform the following mappings: for Cityscapes, we map classes `pedestrian' and `rider' to class `person', while for BDD100k, we map classes `person' and `rider' to class `person'.

\subsection{Model details and parameter settings}
\label{app:model}

Our primary pretrained model from \texttt{detectron2} (model name \texttt{X101-FPN}) consists of a Faster R-CNN backbone with feature pyramid network, region proposal head, and a fully connected final bounding box prediction head, trained for $\sim 37$ epochs on the COCO training split.

\noindent\textbf{Inference parameters.} We identify two key parameters that filter any proposal boxes to produce the final box predictions, which we fix as follows:
\begin{itemize}
    \item[$\bullet$] The score parameter removes box proposals that receive a model confidence score below a specified threshold. We fix this value at a confidence of 0.5.
    \item[$\bullet$] The non-maximum suppression parameter removes any superfluous box proposals that record an IoU overlap above the specified threshold, with exception of the highest confidence box. We fix this IoU threshold at 0.6.
\end{itemize}

\noindent \textbf{Quantile head training.} We freeze all pretrained model weights and only train the new box prediction heads for lower and upper quantiles $\hat{Q}_{\alpha_B/2}$ and $\hat{Q}_{1-\alpha_B/2}$ with a compounded quantile loss. We set the learning rate to 0.02 and train for $\sim 3000$ COCO iterations on the COCO training split with a batch size of 16.

\subsection{Prior work comparison}
\label{app:baseline-impl}

We compare our methods to the conformal bounding box methods presented in Andéol \etal~\cite{l.andeol2023a} -- an extension of \cite{f.degrancey2022} -- which we consider closest prior work. As mentioned in \autoref{sec:related}, they propose conformal scoring functions designed for one-sided, outer prediction intervals. We thus modify our bounding box methods Box-Std and Box-Ens to produce similar one-sided intervals for comparison. We do not modify Box-CQR because it is unclear how a one-sided version of its scoring function should be constructed. Andéol \etal~\cite{l.andeol2023a} only consider a single object class, thus bypassing the need to account for uncertainty in the class label predictions, as we do in \autoref{sec:methods_label}. Since our comparison relates to differences in the bounding box step only, we evaluate both their and our modified approaches across the set of classes using correct class quantiles (\ie, via the Oracle).

Let us reconsider the 2D bounding box setting with explicit coordinate tuples $(c^1, c^2, c^3, c^4)$. To maintain notational consistency with Andéol \etal~\cite{l.andeol2023a}, we rename the coordinates into two pairs denoting box corners as $Y := (x_0, y_0, x_1, y_1)$. We then compare to the following proposed scoring functions:

\begin{itemize}
\item[$\bullet$] \textbf{AddBonf.} We use signed residuals to compute scores as
\begin{equation}
    s(\hat{f}_B(X), Y) = (\hat{x}_0 - x_0, \hat{y}_0 - y_0, x_1 - \hat{x}_1, y_1 - \hat{y}_1)
\label{eq:base-add}
\end{equation}
and obtain outer prediction intervals as
\begin{equation}
    \hat{C}_B(X_{n+1}) = (\hat{x}_0 - \hat{q}_B, \hat{y}_0 - \hat{q}_B, \hat{x}_1 + \hat{q}_B, \hat{y}_1 + \hat{q}_B),
\label{eq:base-add-pi}
\end{equation}
where $\hat{q}_B$ are the respective coordinate-level conformal quantiles at coverage level $(1-\alpha'_B)$, and $\alpha'_B = \alpha_B/4$ is the corrected Bonferroni coverage level.

\item[$\bullet$] \textbf{MultBonf.} We use the scoring function
\begin{equation}
    s(\hat{f}_B(X), Y) = (\frac{\hat{x}_0 - x_0}{\hat{w}}, \frac{\hat{y}_0 - y_0}{\hat{h}}, \frac{x_1 - \hat{x}_1}{\hat{w}}, \frac{y_1 - \hat{y}_1}{\hat{h}}),
\label{eq:base-mult}
\end{equation}
where $\hat{w} = \hat{x}_1 - \hat{x}_0$ and $\hat{h} = \hat{y}_1 - \hat{y}_0$ are predictions for box width and box height. Outer box intervals are then constructed as
\begin{equation}
    \hat{C}_B(X_{n+1}) = (\hat{x}_0 - \hat{w}\,\hat{q}_B, \hat{y}_0 - \hat{h}\,\hat{q}_B, \hat{x}_1 + \hat{w}\,\hat{q}_B, \hat{y}_1 + \hat{h}\,\hat{q}_B),
\label{eq:base-mult-pi}
\end{equation}
where once more $\alpha'_B = \alpha_B/4$ is the Bonferroni correction.

\item[$\bullet$] \textbf{AddMax, MultMax.} Instead of a Bonferroni correction, Andéol \etal~\cite{l.andeol2023a} also suggest a $\max$-operation over coordinate scores in \autoref{eq:base-add} and \autoref{eq:base-mult} respectively, resulting in a set of `maximal' scores. A conformal quantile $\hat{q}^{\max}_B$ is then computed directly at target coverage level over these scores and used across coordinates, alleviating the need for further correction. The principle is related to the $\max$-rank approach \cite{timans2023maxrank}, but instead operates directly in the domain of scores, which can be problematic due to an improper influence of score magnitudes. Outer box intervals are then constructed as in \autoref{eq:base-add-pi} (AddMax) and \autoref{eq:base-mult-pi} (MultMax) by replacing the quantiles with $\hat{q}^{\max}_B$.
\end{itemize}

\noindent The above approaches are compared to our following modified methods:
\begin{itemize}
\item[$\bullet$] \textbf{Box-Std.} We use the conformal scoring approach from \autoref{eq:base-add} in conjunction with the $\max$-rank multiple testing correction. This is the equivalent of a one-sided signed version of the regression residuals employed by Box-Std.

\item[$\bullet$] \textbf{Box-Ens.} We create a one-sided version of Box-Ens by using the conformal scores from \autoref{eq:base-mult}, but using ensemble-fused coordinate predictions and normalizing by the obtained `heuristic' ensemble uncertainty $\hat{\sigma}$.

\item[$\bullet$] \textbf{Box-Mult.} We additionally consider using the conformal scores from \autoref{eq:base-mult} directly with a $\max$-rank correction, a combination approach that can be related to the normalized scores of Box-Ens.
\end{itemize}

\noindent \textbf{Box stretch metric.} We implement the proposed \emph{box stretch} evaluation metric from Andéol \etal~\cite{l.andeol2023a} that assesses the additional box surface area incurred by applying conformal prediction. Formally, we denote the metric as 
\begin{equation}
    Stretch = \frac{1}{n_t}\sum_{j=n+1}^{n+n_t} \sqrt{\frac{\mathcal{A}(\hat{C}_B(X_j))}{\mathcal{A}(\hat{f}_B(X_j))}},
\label{eq:stretch}
\end{equation}
where $\mathcal{A}(\cdot)$ is the computed surface area of the bounding box formed by the respective input, \ie, the outer box interval bounds, and the predicted bounding box. Ideally, we desire $Stretch$ to be close to 1.0.

\textbf{Mean prediction interval width.} The $MPIW$ metric is formally defined for two-sided intervals only. In order to permit comparison using $MPIW$ alongside $Stretch$, we transform any of the above \emph{one-sided} interval constructions into \emph{two-sided} intervals. We do so by considering the distance of the outer interval coordinates to the box center, \ie, placing a lower interval bound at the predicted box center coordinates. Note that we also do the same for our own initially two-sided methods to allow for a fair comparison. 

\section{Additional experimental results}
\label{app:results}

\subsection{Effect of calibration on Naive and ClassThr}
\label{app:calibration-ece}

\begin{figure}[t]
    \centering
    \includegraphics[width=0.8\linewidth]{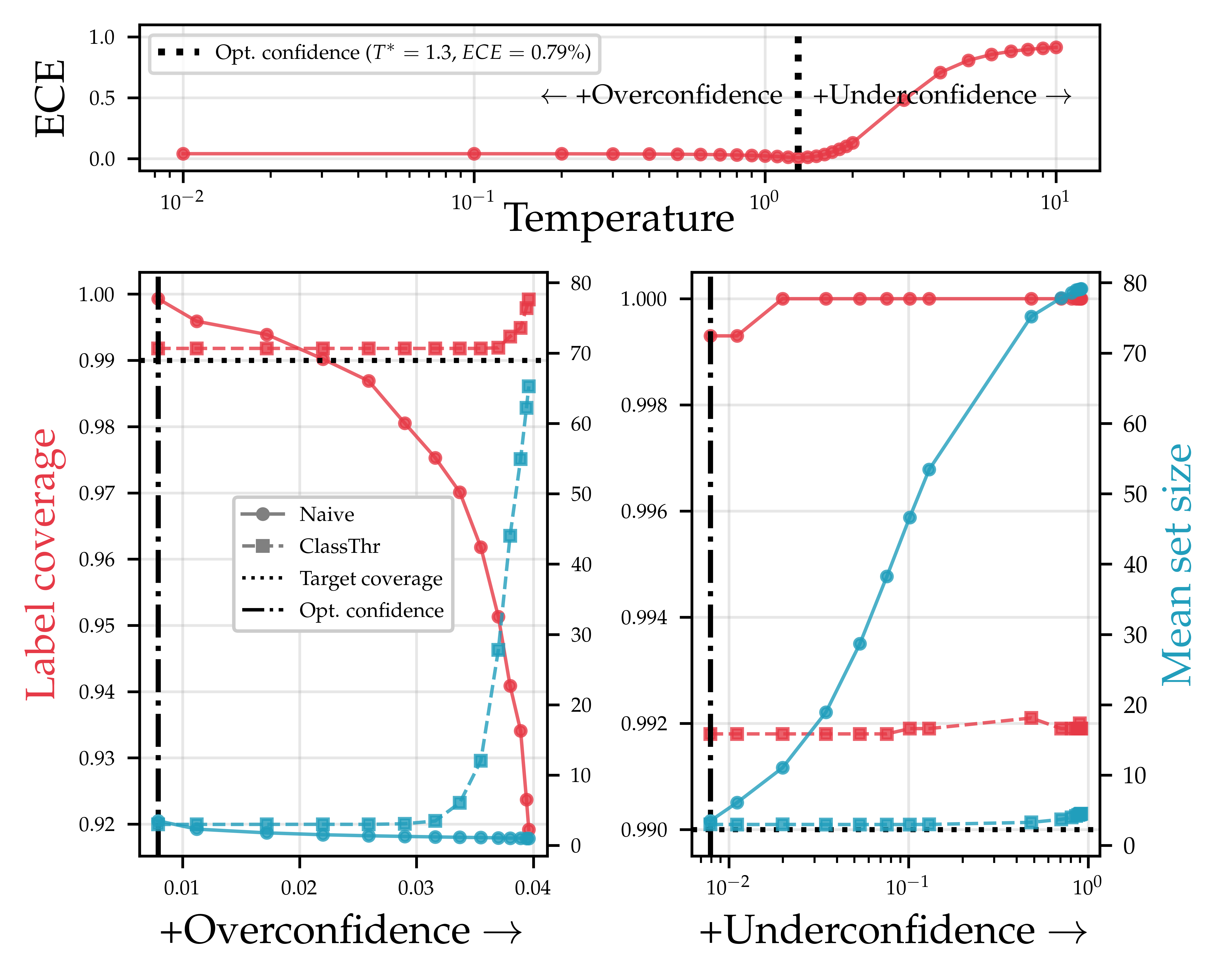}
    \caption{We influence model calibration via temperature scaling \cite{guo2017calibration} to determine regimes of model over- and underconfidence based on $ECE$ (top). We compare Naive and ClassThr for both regimes via label coverage and label set sizes. As miscalibration through overconfidence increases (bottom left), Naive fails to satisfy target coverage. As miscalibration through underconfidence increases (bottom right), the set sizes of Naive explode. In contrast, ClassThr remains stable in respective metrics across regimes.}
    \label{fig:naive-calibration}
\end{figure}

We aim to better understand the strong empirical coverage results observed in \autoref{fig:metrics-scatter} for the Naive label prediction set baseline. Since Naive is theoretically motivated from a model calibration perspective, we empirically assess its sensitivity to changes in calibration, and compare results to ClassThr. We refer to Guo \etal~\cite{guo2017calibration} for more details on calibration and related concepts.

Specifically, we proxy model calibration via top-class calibration, which can be evaluated using the \emph{Expected Calibration Error (ECE)} metric (see \autoref{fig:naive-calibration}). We then influence model calibration via temperature scaling \cite{guo2017calibration} -- a simple logit-scaling approach -- to induce regimes of model miscalibration via over- and underconfidence. We find that our object detector (temperature $T=1.0$) is generally slightly overconfident, and optimal confidence in terms of \emph{ECE} lies at $T^*=1.3$ (\autoref{fig:naive-calibration}, top). We then perform our two-step conformal approach using Box-Std, and compare ClassThr to the use of Naive for multiple severities of over- and underconfidence (\autoref{fig:naive-calibration}, bottom). Both methods are evaluated in regards to label coverage $(1-\alpha_L)$ (left axis) and \emph{mean set size} (right axis). 

\noindent We find that as miscalibration through overconfidence increases (bottom left), Naive gradually fails to satisfy target coverage of 99 \%. In contrast, ClassThr is able to continuously provide desired coverage levels, at the cost of increasing set sizes for high miscalibration. As miscalibration through underconfidence increases (bottom right), the set sizes of Naive explode, tending towards the full domain (\emph{mean set size} $\sim 80$). In contrast, ClassThr provides tighter coverage levels and maintains reasonable set sizes. We conclude that while Naive provides good empirical results under relatively accurate model calibration (as seen in \autoref{fig:metrics-scatter}) it is quite sensitive to miscalibration. In contrast, ClassThr seems to provide a more robust safety assurance even under settings of over- and underconfidence, at the cost of potentially larger set sizes.

\subsection{Prior work comparison and other results}
\label{app:bonf-baseline-results}

\input{text/X_fig_prior}

\autoref{fig:app-base-comp} displays results for the comparison of our bounding box methods to the proposed scoring functions from Andéol \etal~\cite{l.andeol2023a} (see \autoref{app:baseline-impl}). We observe that our modified one-sided approaches are competitive, as measured via efficiency metrics $MPIW$ and $Stretch$, and our best approach even slightly outperforms.

\autoref{tab:app-bonf} displays results for the bounding box methods from \autoref{sec:methods_box} across all three datasets using the Bonferroni correction rather than $\max$-rank to account for multiple testing (see \autoref{subsec:methods_mht}). In comparison to results using $\max$-rank (\autoref{fig:violins}, \autoref{fig:app-violins-city}, and \autoref{fig:app-violins-coco}), performance is generally inferior. This is captured by larger $MPIW$ and a tendency to overcover beyond the target level of $(1-\alpha_B) \approx 0.9$. The effects are particularly visible for datasets COCO and Cityscapes. 

\begin{figure}[t]
    \centering
    \includegraphics[width=0.35\linewidth]{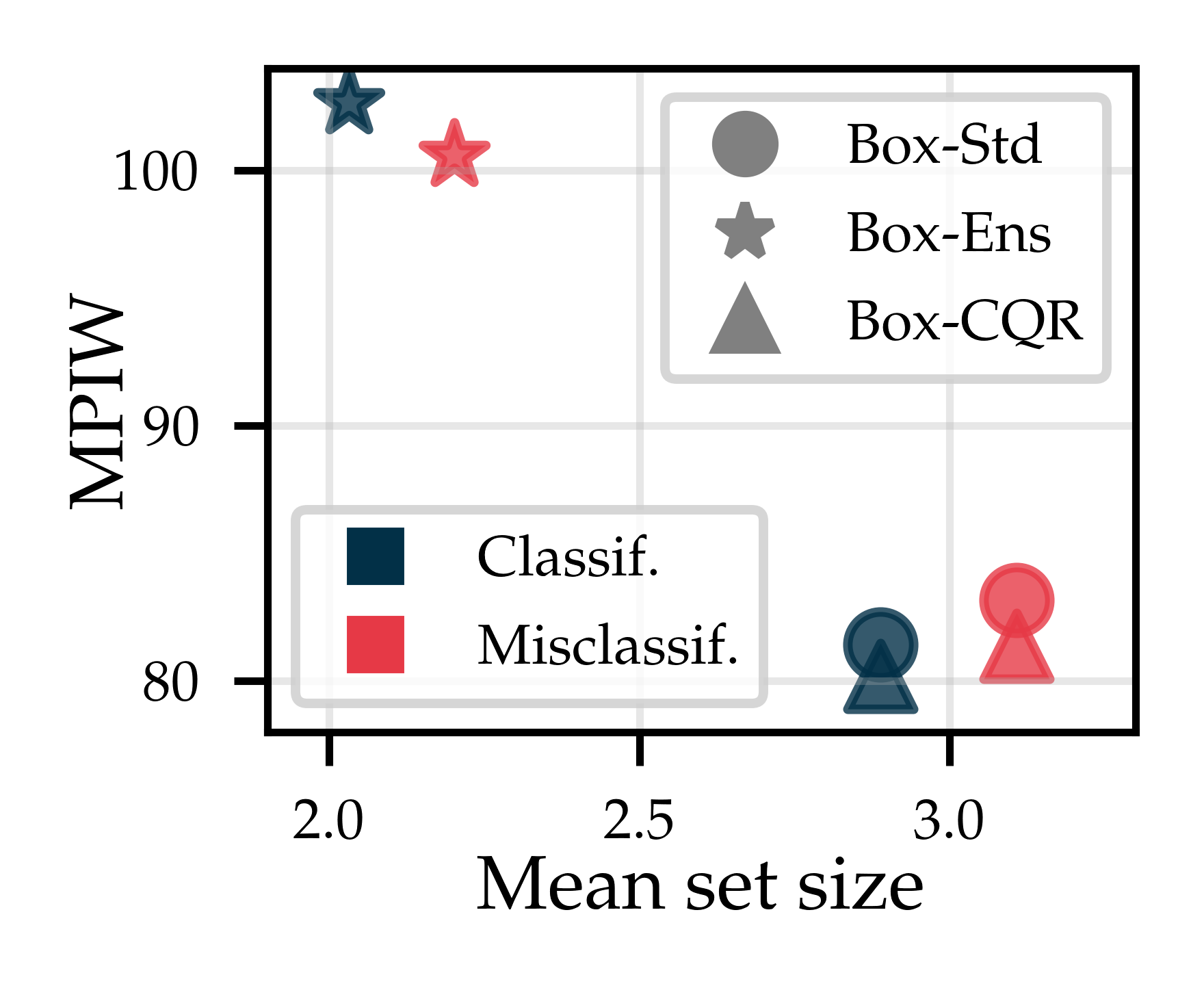}
    \caption{Prediction set sizes for objects stratified by classification on COCO.}
    \label{fig:size-misclassif}
\end{figure}

We further ablate results of our two-step approach using ClassThr by stratifying efficiency metrics across object (mis-)classification in \autoref{fig:size-misclassif} for COCO. Results confirm that the model's higher uncertainty for more ambiguous objects -- as reflected by their misclassification -- tends to result in larger prediction sets.

\input{text/X_tab_bonf}

\subsection{Predictive performance}
\label{app:pred-perf-ap}

We validate the predictive performance -- as measured via average precision (AP) metrics following the COCO detection challenge \footnote{See \url{https://cocodataset.org/\#detection-eval}.} -- for our primary pretrained object detection model \texttt{X101-FPN} across datasets in \autoref{tab:ap-scores}. Obtained scores are similar to results reported by \texttt{detectron2} \footnote{\url{https://github.com/facebookresearch/detectron2/blob/main/MODEL_ZOO.md}}, confirming that the model performs adequately as a base predictor on which we run our conformal procedures. Other pretrained models employed for Box-Ens perform similarly.

\input{text/X_tab_ap}

\subsection{Different target coverage levels}
\label{app:diff-coverage-results}

\input{text/X_fig_abl_cov}

We evaluate a series of different target combinations of box coverage $(1-\alpha_B)$ and label coverage $(1-\alpha_L)$ for our two-step conformal approach using Box-Std and ClassThr on all the datasets in \autoref{fig:abl-cov}. Specifically, we consider all combinations of $(1-\alpha_B) \in \{0.85, 0.9, 0.95\}$ and $(1-\alpha_L) \in \{0.8, 0.9, 0.99, 1.0\}$. Note that the combination $(1-\alpha_B) = 0.9, (1-\alpha_L) = 0.99$ is the primary coverage goal throughout this work (see also \autoref{subsec:methods_label_class}). We compare the obtained empirical coverage levels against desired target levels, denoted by the intersections of dashed lines in the top figure row. Target coverage levels are mostly satisfied with regards to box coverage, albeit the target is reached more clearly as calibration set sizes increase, as we move from COCO (left) to Cityscapes (middle) and BDD100k (right). In contrast, target levels for label coverage are fully satisfied, and in fact tend to overcover. As calibration set sizes increase, this overcoverage tendency decreases. See also \autoref{app:emp-cov-distr} for more details on calibration set size.

We further display obtained efficiency, as measured via $MPIW$ for box intervals and \emph{mean set size} for label sets (bottom row). Two trends are immediately apparent: as we increase the requirement on box coverage, $MPIW$ increases; similarly, as we increase the requirement on label coverage, \emph{mean set size} increases. Interestingly, the gap between permitting minimal label miscoverage and no miscoverage at all, \ie, $\alpha_L=0.01$ and $\alpha_L=0$, tends to increase as the number of calibration samples rises. This observation motivates our choice to select $\alpha_L=0.01$ as a practical approximation ensuring total label coverage whilst minimally affecting the downstream box guarantees.

\input{text/X_fig_violins}

\subsection{Additional visualisations}
\label{app:more-vis}

We display additional examples of conformal bounding box intervals produced using our two-step approach for COCO in \autoref{fig:pi-plots-coco}, for Cityscapes in \autoref{fig:pi-plots-cityscapes} and for BDD100k in \autoref{fig:pi-plots-bdd100k}. We use ClassThr to produce class label prediction sets with 99\% guaranteed coverage and either Box-Std, Box-Ens or Box-CQR (from left to right by image column) to produce box intervals with $\sim$90\% guaranteed coverage. We depict a range of different classes which align with our particular interest in urban driving scenes: person, bicycle, motorcycle, car, bus and truck. Note that for some images we only visualize a filtered amount of classes for clarity. We observe that despite its ability to ensure probabilistic guarantees even under object misclassification, the obtained bounding box intervals are reasonably tight, and could effectively be used for decision-making.

\input{text/X_fig_pi}

%% file: text/X_fig_cov.tex
\begin{figure*}[t]
    \centering

    \begin{subfigure}[t]{.49\textwidth}
        \raisebox{-\height}{
            \includegraphics[width=\textwidth]{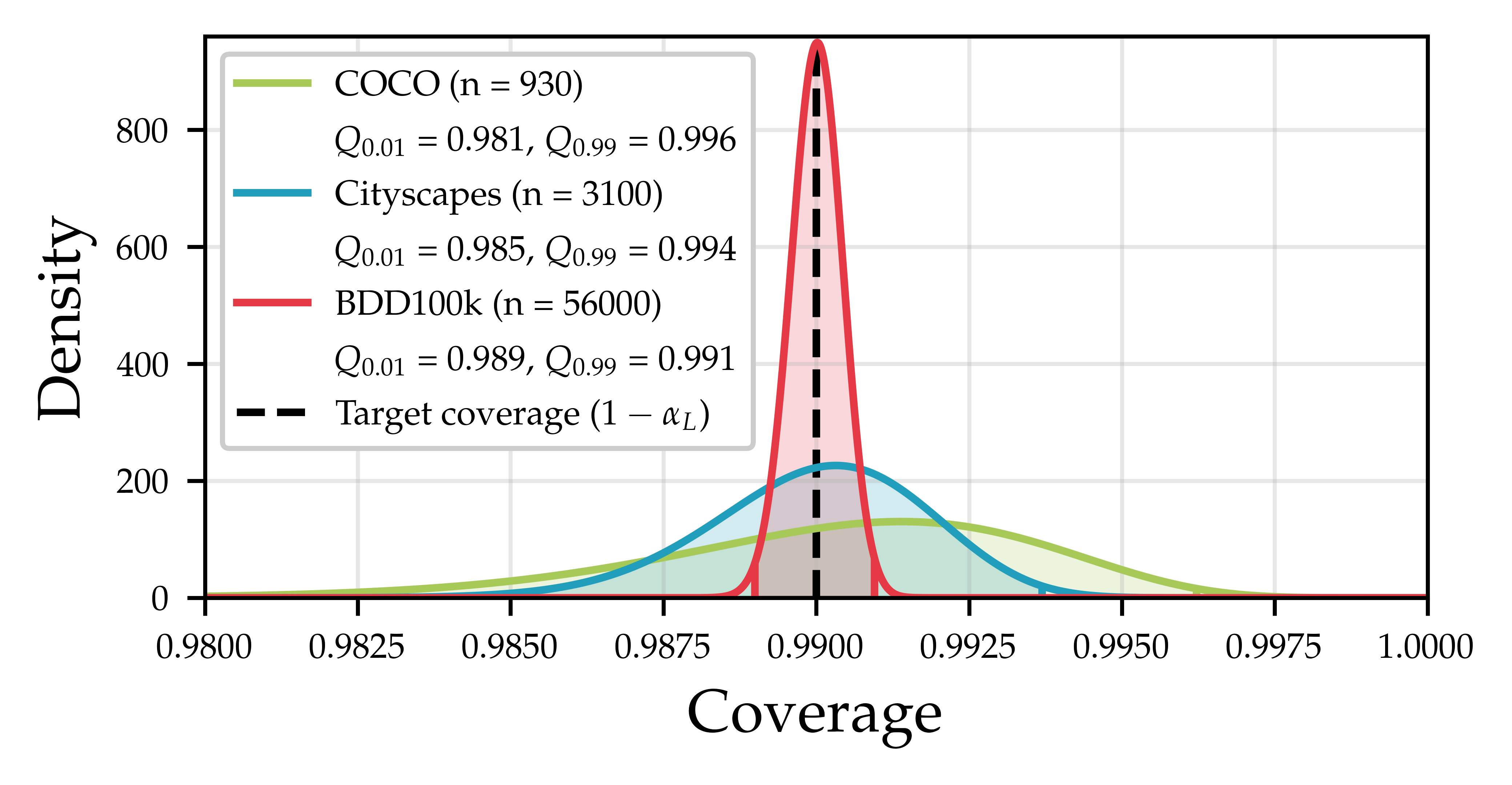}
        }
    \end{subfigure}%
    \hfill
    \begin{subfigure}[t]{.49\textwidth}
        \raisebox{-\height}{
            \includegraphics[width=\textwidth]{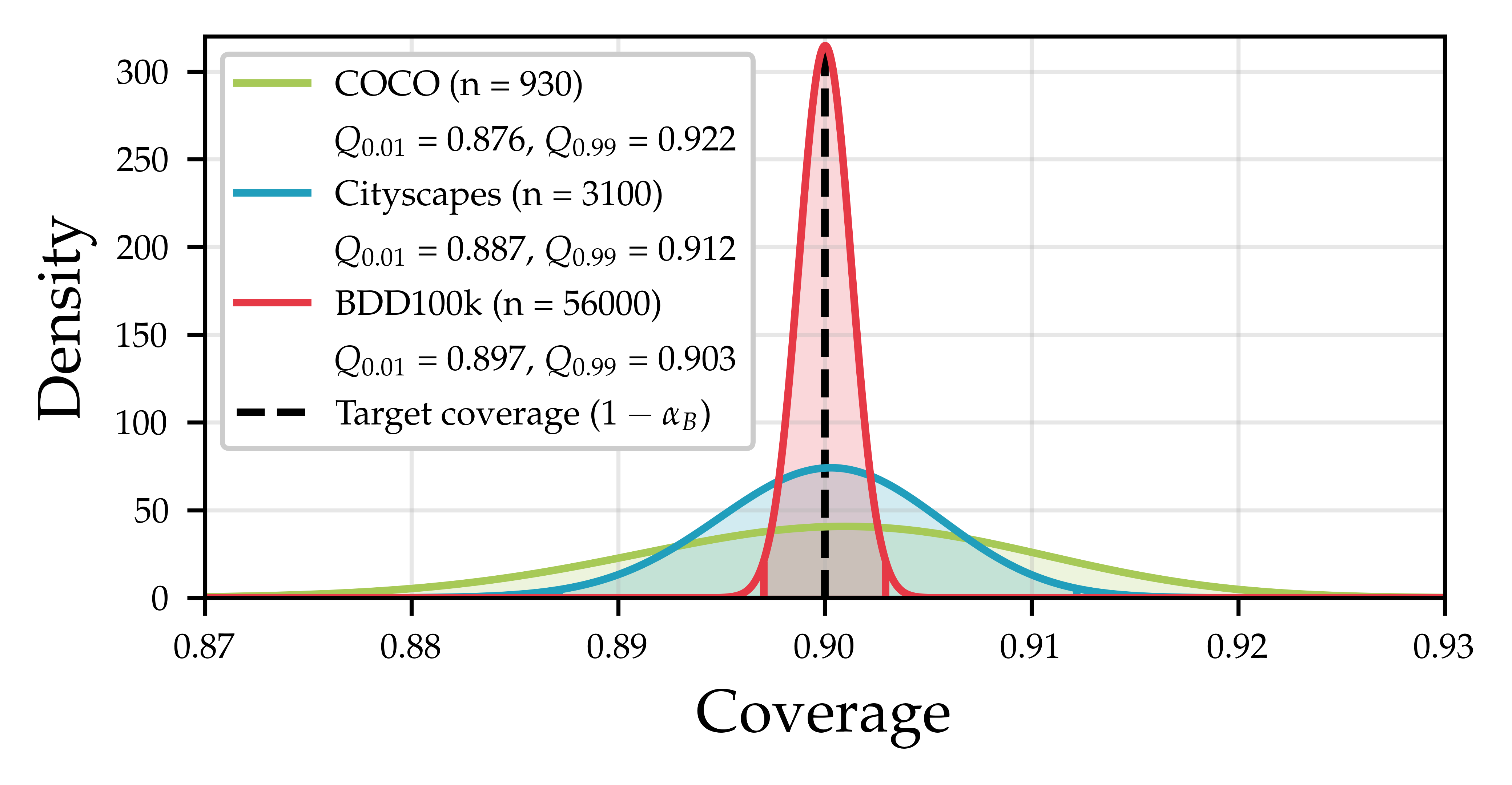}
        }
    \end{subfigure}
    \caption{Exact marginal coverage distributions for our three considered datasets (COCO, Cityscapes, BDD100k) on the basis of their calibration set sizes. For each distribution we additionally mark the $1\%$ and $99\%$ quantiles, as well as the desired target coverage. \emph{Left:} For the conformal class label prediction sets on the basis of a target miscoverage rate $\alpha_L=0.01$. \emph{Right:} For the conformal box prediction intervals on the basis of a target miscoverage rate $\alpha_B=0.1$. Obtained empirical coverage levels across experiments fall within reasonable regions of the derived coverage distributions.}
    \label{fig:emp-cov}
\end{figure*}

%% file: text/X_tab_data.tex
\begin{table*}[t]
    \centering
    \setlength{\tabcolsep}{5pt}
    \begin{tabular}{lrrrrrrr}
        & & \multicolumn{6}{c}{\textbf{Object class}} \\
        \cmidrule(lr){3-8}
        \textbf{Dataset} & \textbf{\# Images} & Person & Bicycle & Motorcycle & Car & Bus & Truck \\
        \toprule
        COCO & 5000 & 10777 & 314 & 367 & 1918 & 283 & 414 \\
        Cityscapes & 5000 & 24713 & 5871 & 895 & 33658 & 477 & 577 \\
        BDD100k & 70000 & 96929 & 7124 & 3023 & 701507 & 11977 & 27963 \\
        \bottomrule
        \\
    \end{tabular}
\caption{Image and object counts per dataset for the selected set of common classes.}
\label{tab:datasets-ist}
\end{table*}

\begin{table*}[t]
    \centering
    \setlength{\tabcolsep}{4pt}
    \begin{tabular}{llrrrrrrr}
        & & \multicolumn{7}{c}{\textbf{Object class}} \\
        \cmidrule(lr){3-9}
        \textbf{Dataset} & \textbf{Object size} & All & Person & Bicycle & Motorcycle & Car & Bus & Truck \\
        \toprule
        \multirow{3}{*}{COCO} & Small & 17.68 & 21.70 & 15.96 & 7.58 & 44.60 & 4.98 & 11.27 \\
        & Medium & 36.88 & 36.29 & 43.09 & 34.09 & 40.70 & 25.34 & 41.78 \\
        & Large & 45.44 & 42.01 & 40.96 & 58.33 & 14.69 & 69.68 & 46.95 \\
        \cmidrule(lr){3-9}
        \multirow{3}{*}{Cityscapes} & Small & 2.90 & 6.56 & 0.49 & 2.99 & 7.34 & 0 & 0 \\
        & Medium & 36.87 & 59.55 & 50.77 & 35.05 & 46.81 & 12.36 & 16.67 \\
        & Large & 60.24 & 33.89 & 48.74 & 61.96 & 45.85 & 87.64 & 83.33 \\
        \cmidrule(lr){3-9}
        \multirow{3}{*}{BDD100k} & Small & 23.40 & 28.87 & 5.87 & 12.51 & 23.13 & 0.66 & 2.73 \\
        & Medium & 49.36 & 60.99 & 66.59 & 53.60 & 46.85 & 28.68 & 39.46 \\
        & Large & 38.34 & 10.14 & 27.54 & 33.89 & 30.02 & 70.66 & 57.81 \\
        \bottomrule
        \\
    \end{tabular}
\caption{Objects stratified by size as a fraction (in \%) of total object counts (\autoref{tab:datasets-ist}) for the selected set of common classes.}
\label{tab:datasets-ist-by-size}
\end{table*}

%% file: text/X_fig_prior.tex
\begin{figure*}[t]
    \centering

    \begin{subfigure}[t]{\textwidth}
        \raisebox{-\height}{
            \includegraphics[width=\textwidth]{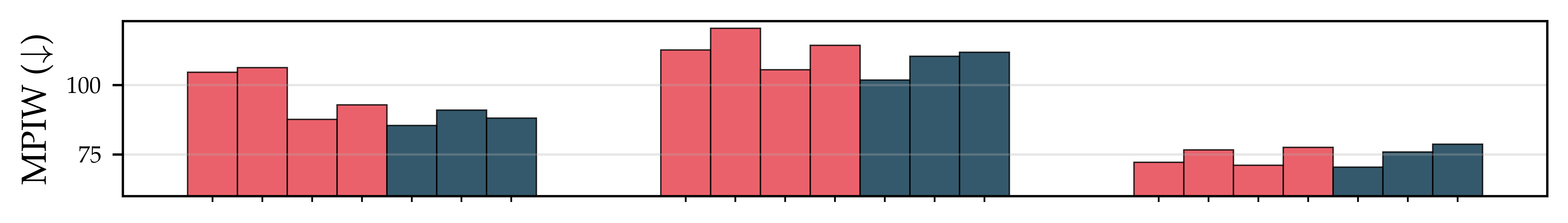}
        }
    \end{subfigure}%
    \hfill
    \begin{subfigure}[t]{\textwidth}
        \raisebox{-\height}{
            \includegraphics[width=\textwidth]{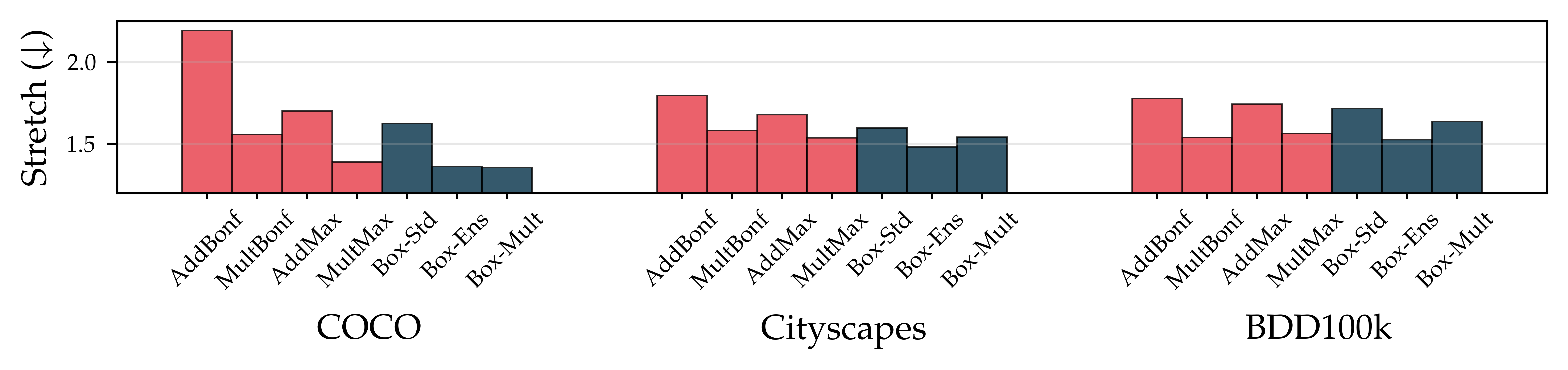}
        }
    \end{subfigure}
    \caption{Comparison of our modified one-sided approaches to conformal scoring functions proposed by \cite{l.andeol2023a} (see \autoref{app:baseline-impl}) in terms of $MPIW$ (top) and $Stretch$ (bottom). We find that our approaches are competitive with regards to both evaluation metrics.}
    \label{fig:app-base-comp}
\end{figure*}

%% file: text/X_tab_bonf.tex
\begin{table*}[t]
    \centering
    \setlength{\tabcolsep}{5pt}
    \begin{tabular}{llrrrrrr}
        \textbf{Dataset} & \textbf{Method} & \textbf{$MPIW$} & \textbf{$Stretch$} & \textbf{$Cov$} & \textbf{$Cov_S$} & \textbf{$Cov_M$} & \textbf{$Cov_L$} \\
        \toprule
        \multirow{3}{*}{COCO} & Box-Std & 102.6767 & 2.6543 & 0.9400 & 0.9992 & 0.9792 & 0.8595 \\
        & Box-Ens & 90.8789 & 1.7573 & 0.9358 & 0.9291 & 0.9270 & 0.9418 \\
        & Box-CQR & 90.8073 & 2.3136 & 0.9374 & 0.9925 & 0.9604 & 0.8764 \\
        \cmidrule(lr){3-8}
        \multirow{3}{*}{Cityscapes} & Box-Std & 84.7931 & 1.9156 & 0.9279 & -- & 0.9779 & 0.8790 \\
        & Box-Ens & 102.2210 & 1.7700 & 0.9233 & -- & 0.9208 & 0.9243 \\
        & Box-CQR & 82.5204 & 1.8376 & 0.9240 & -- & 0.9545 & 0.8910 \\
        \cmidrule(lr){3-8}
        \multirow{3}{*}{BDD100k} & Box-Std & 51.4646 & 1.9352 & 0.9134 & 0.9986 & 0.9684 & 0.7656 \\
        & Box-Ens & 63.6889 & 1.7657 & 0.9084 & 0.8735 & 0.8994 & 0.9297 \\
        & Box-CQR & 51.7637 & 1.8438 & 0.9087 & 0.9782 & 0.9449 & 0.8004 \\
        \bottomrule
        \\
    \end{tabular}
    \caption{Metrics comparison of proposed conformal methods for box coordinates (see \autoref{sec:methods_box}) across the three datasets using the Bonferroni correction for multiple testing. Values are means over trials and the selected set of classes.}
\label{tab:app-bonf}
\end{table*}

%% file: text/X_tab_ap.tex
\begin{table*}[t]
    \centering
    \begin{tabular}{lrrrrrr}
        & \multicolumn{6}{c}{\textbf{Predictive performance metrics}} \\
        \cmidrule(lr){2-7}
        \textbf{Dataset} & AP@IoU=.50:.05:.95 & AP@IoU=.75 & AP@IoU=.50 & AP-small & AP-med & AP-large \\
        \toprule
        COCO & 0.4521 & 0.4937 & 0.6655 & 0.2184 & 0.2781 & 0.4281 \\
        Cityscapes & 0.4320 & 0.4641 & 0.6637 & 0.0270 & 0.0459 & 0.2782 \\
        BDD100k & 0.3098 & 0.3141 & 0.5256 & 0.0745 & 0.1400 & 0.3055 \\
        \bottomrule
        \\
    \end{tabular}
\caption{Average precision (AP) scores following the COCO detection challenge metrics for our primarily employed pretrained object detection model \texttt{X101-FPN} (see \autoref{app:model}). Results are the mean over our selected set of classes. The primary metric AP@IoU=.50:.05:.95 averages AP scores for 10 different IoU thresholds in $[0.5, 0.95]$ with step size $0.05$. AP-small, AP-med and AP-large compute scores stratified across object sizes (see our adaptivity metric in \autoref{subsec:exp_metrics}).}
\label{tab:ap-scores}
\end{table*}

%% file: text/X_fig_abl_cov.tex
\begin{figure*}[t]
    \centering  
    \begin{subfigure}[t]{0.9\textwidth}
        \centering
        \raisebox{-\height}{
            \includegraphics[width=\textwidth]{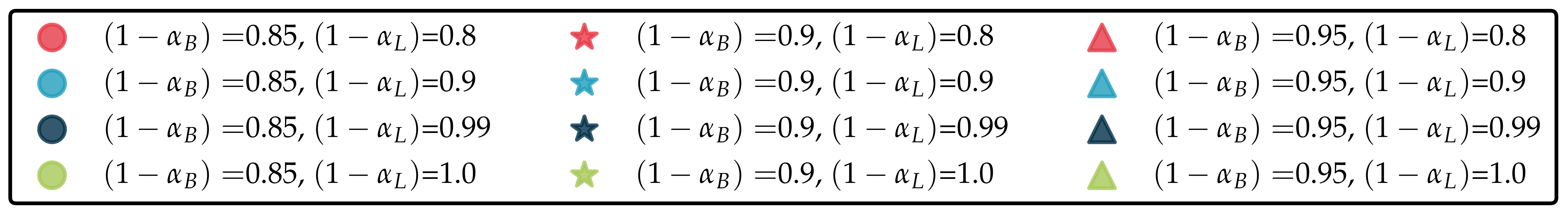}
        }
    \end{subfigure}
    \vspace{-2mm}
    \begin{subfigure}[t]{.33\textwidth}
        \raisebox{-\height}{
            \includegraphics[width=\textwidth]{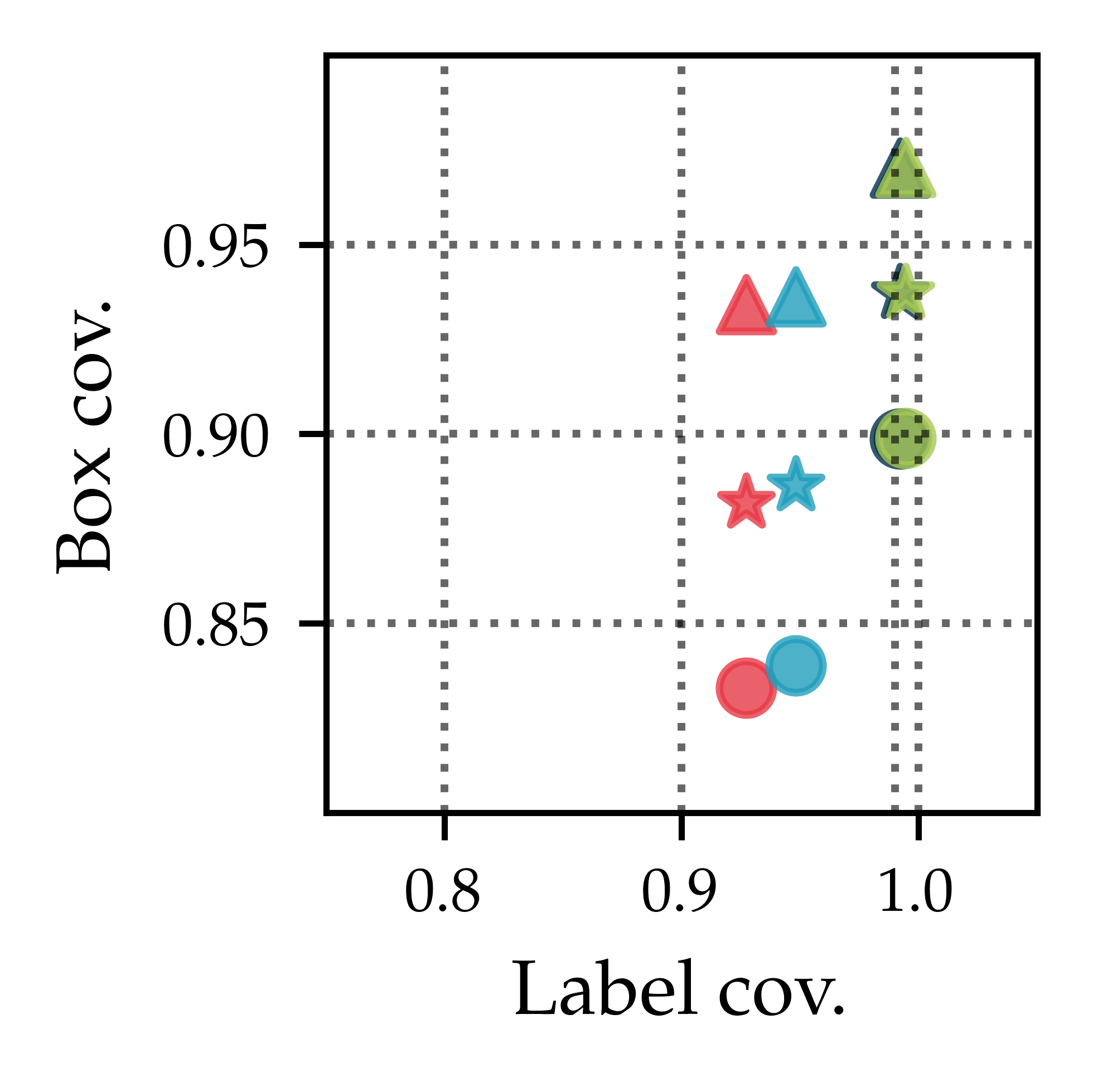}
        }
    \end{subfigure}%
    \hfill
    \begin{subfigure}[t]{.33\textwidth}
        \raisebox{-\height}{
            \includegraphics[width=\textwidth]{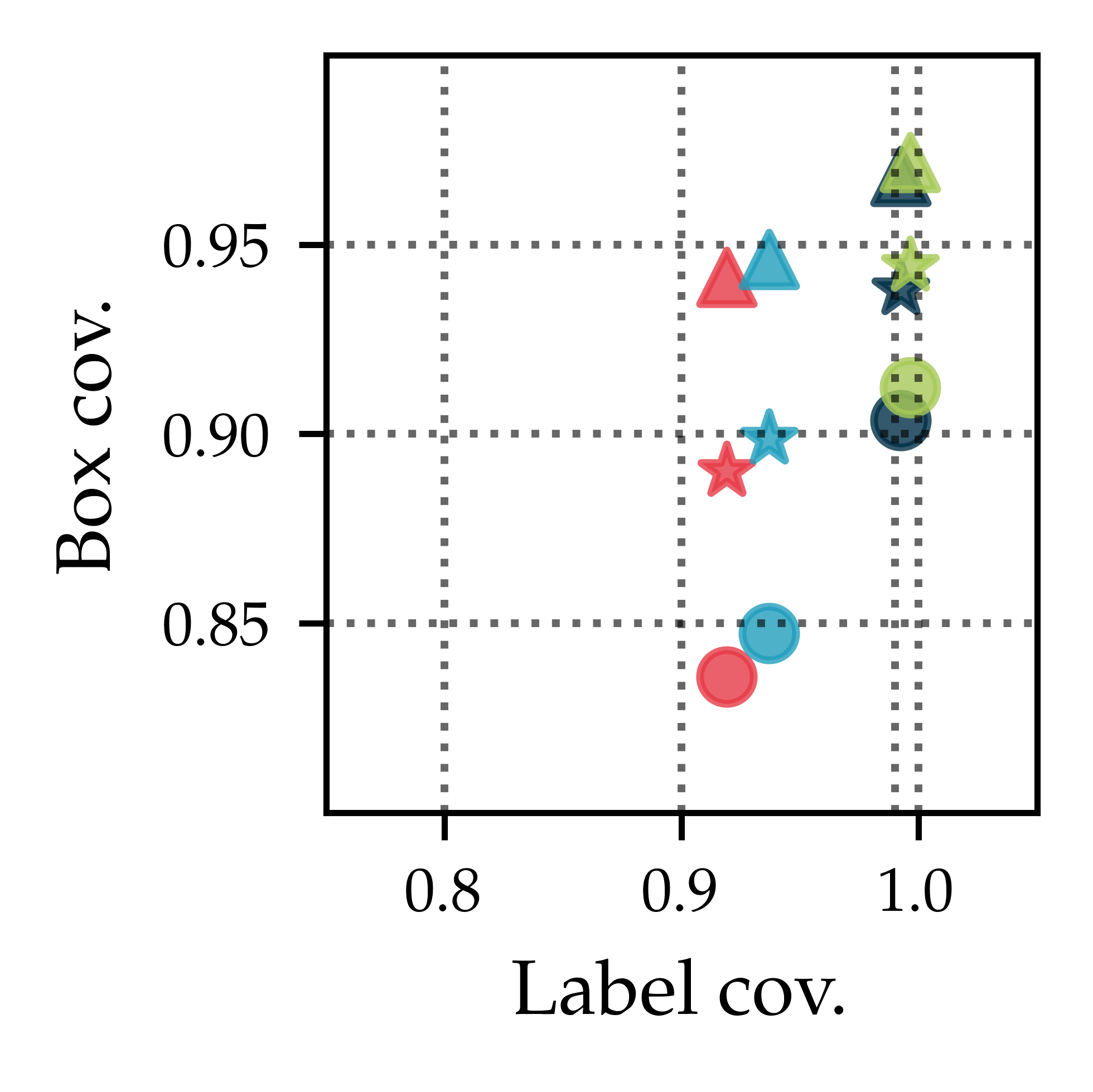}
        }
    \end{subfigure}
    \hfill
    \begin{subfigure}[t]{.33\textwidth}
        \raisebox{-\height}{
            \includegraphics[width=\textwidth]{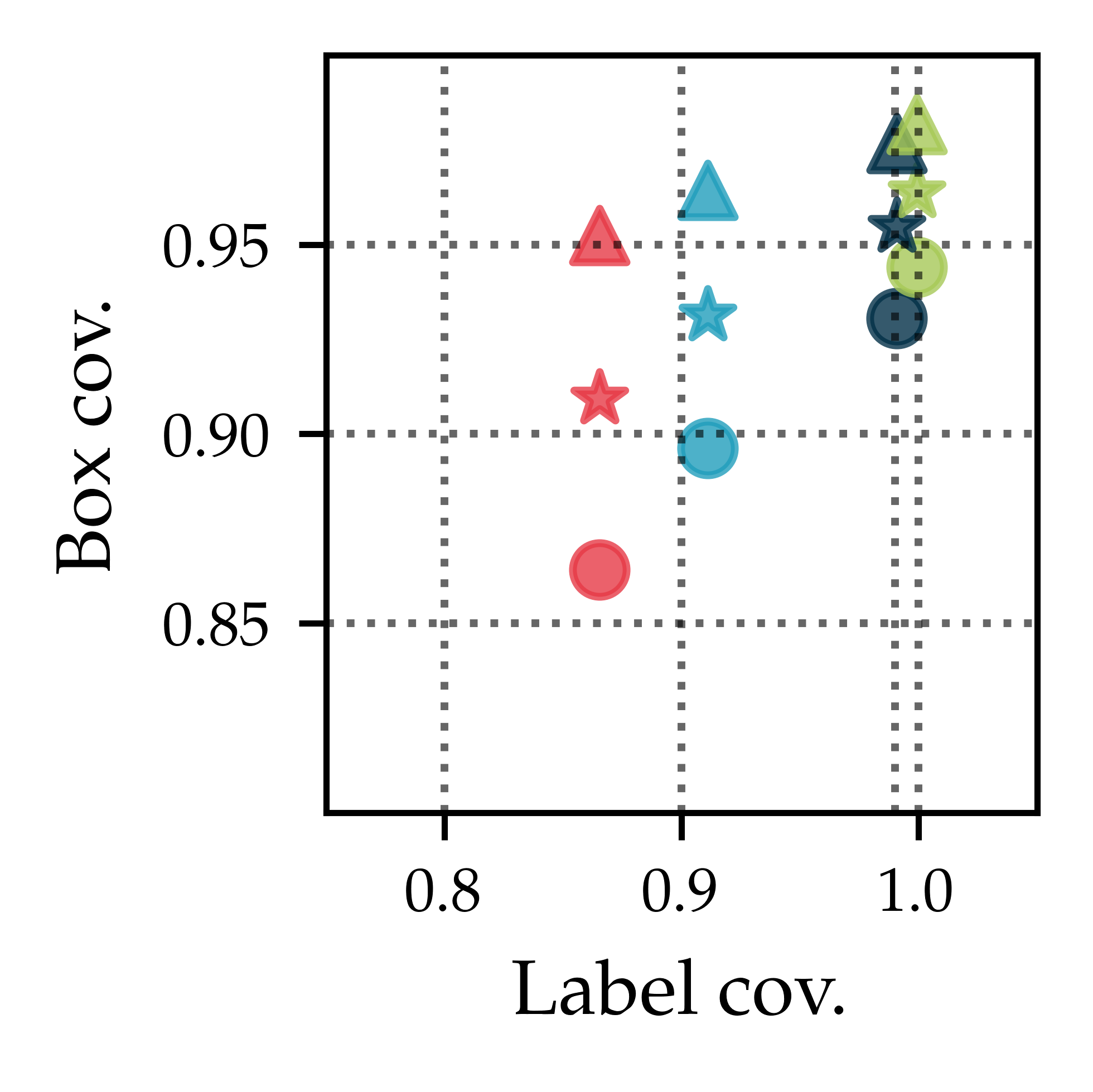}
        }
    \end{subfigure}
    \vspace{-2mm}
    \begin{subfigure}[t]{.33\textwidth}
        \raisebox{-\height}{
            \includegraphics[width=\textwidth]{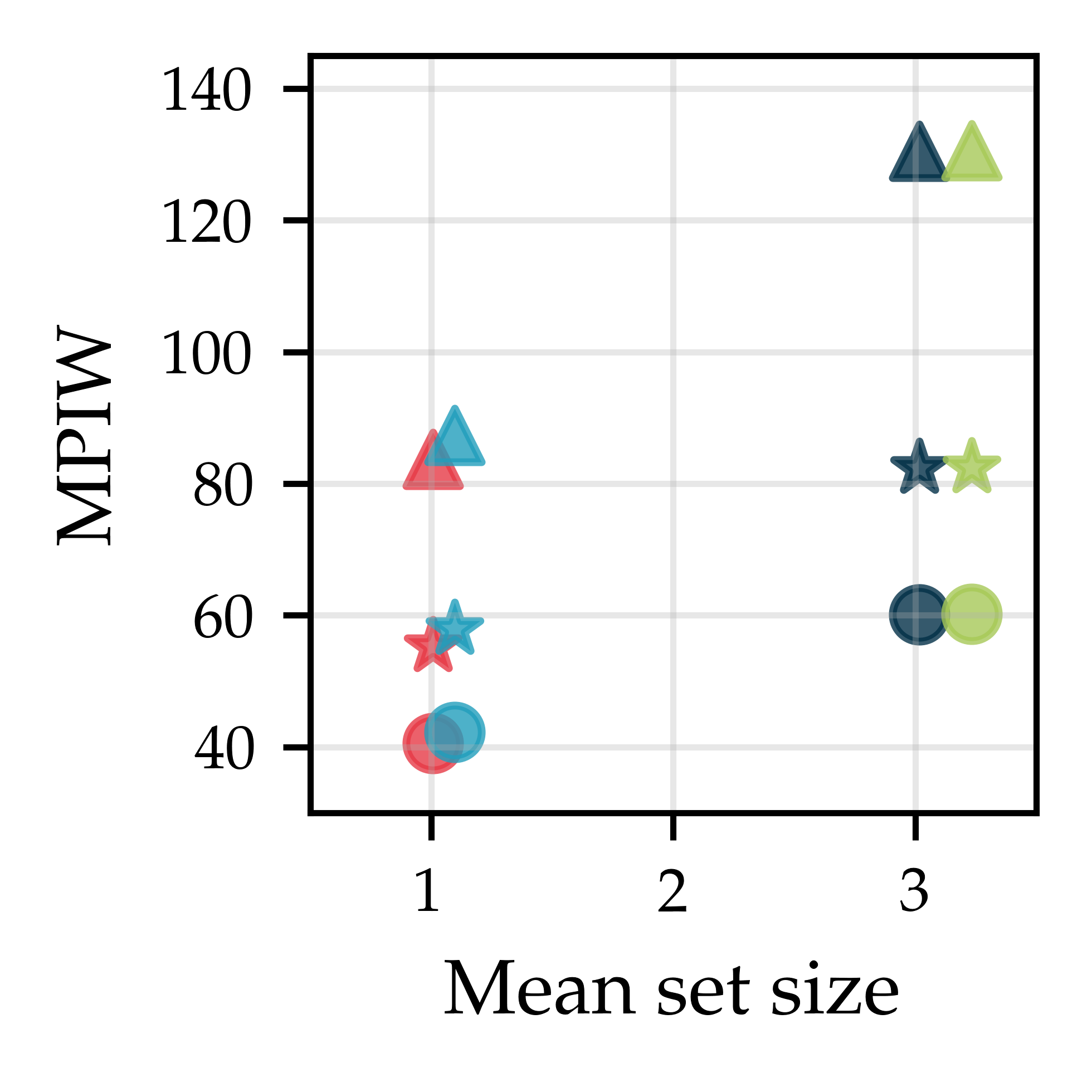}
        }
    \end{subfigure}%
    \hfill
    \begin{subfigure}[t]{.33\textwidth}
        \raisebox{-\height}{
            \includegraphics[width=\textwidth]{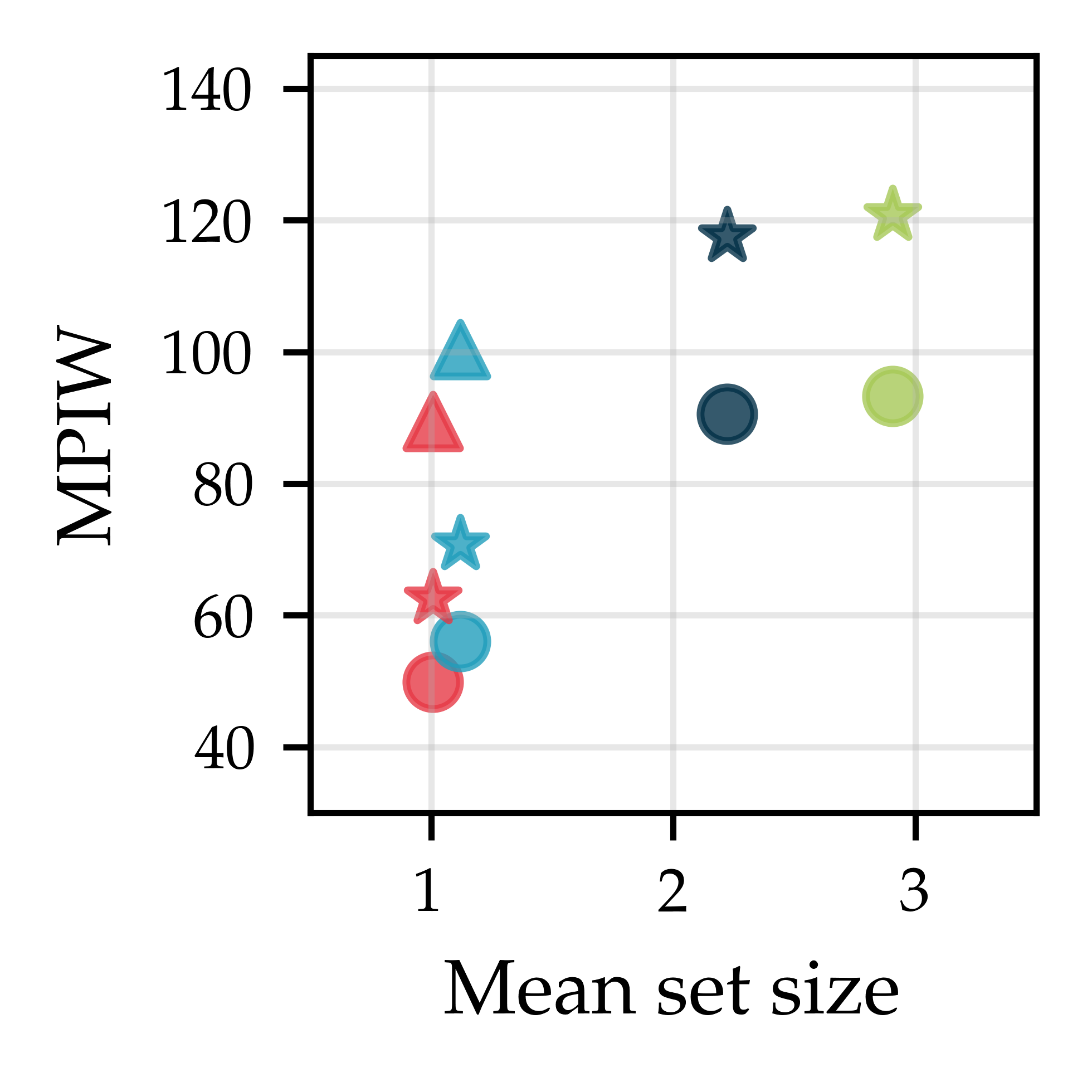}
        }
    \end{subfigure}
    \hfill
    \begin{subfigure}[t]{.33\textwidth}
        \raisebox{-\height}{
            \includegraphics[width=\textwidth]{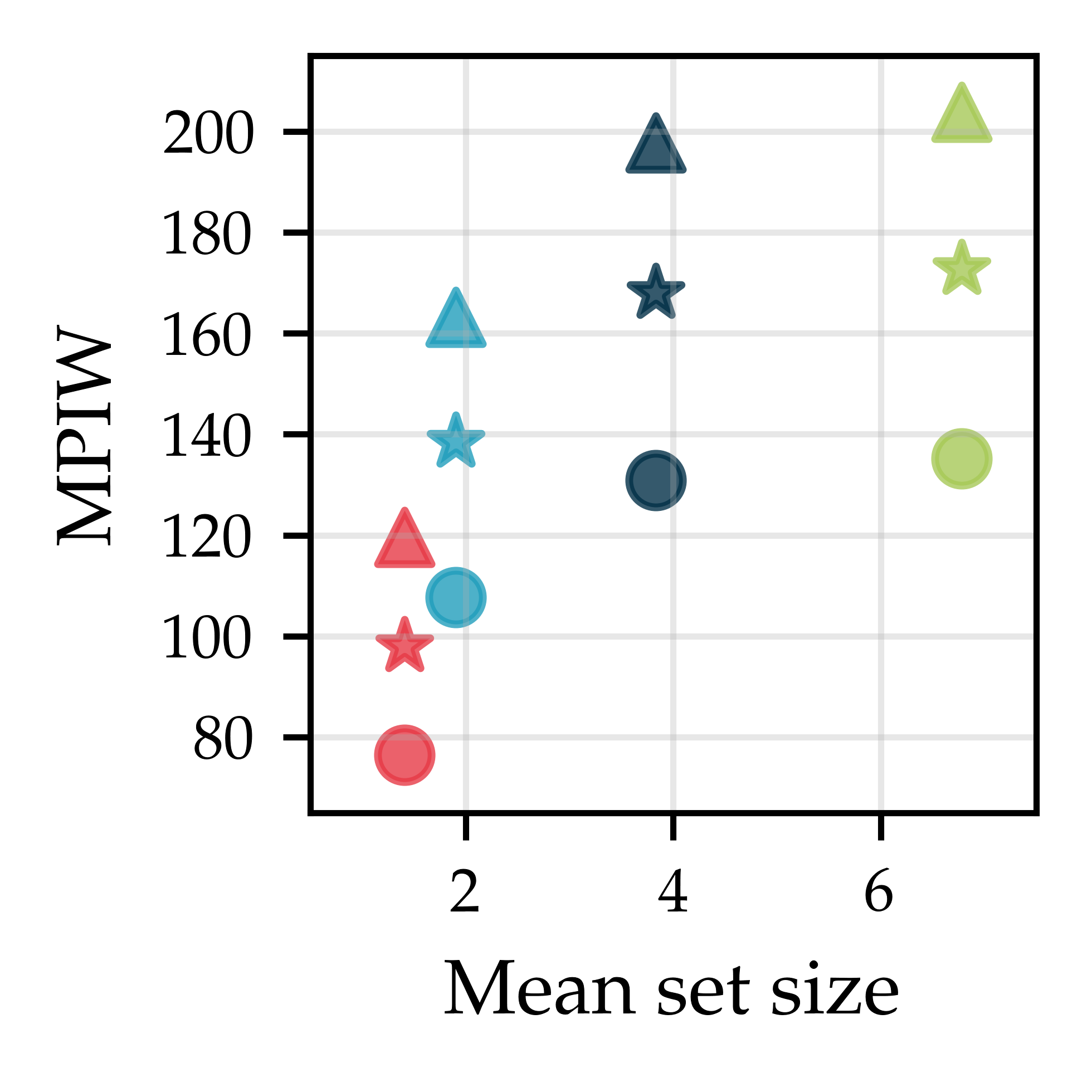}
        }
    \end{subfigure}

    \caption{We ablate different combinations of box coverage $(1-\alpha_B)$ and label coverage $(1-\alpha_L)$ for a two-step conformal approach using Box-Std and ClassThr on COCO (left), Cityscapes (middle) and BDD100k (right). We denote at each intersection of dashed lines (\includegraphics[width=2em]{fig/caption_line1.png}) the desired target combination. We also display obtained efficiencies, as measured via $MPIW$ for conformal box intervals and \emph{mean set size} for conformal label sets. 
    Results are averaged across the selected set of classes and 100 trials.}
    \label{fig:abl-cov}
\end{figure*}

%% file: text/X_fig_violins.tex
\begin{figure*}[t]
    \centering  

    \begin{subfigure}[t]{\textwidth}
        \raisebox{-\height}{
            \includegraphics[width=\textwidth]{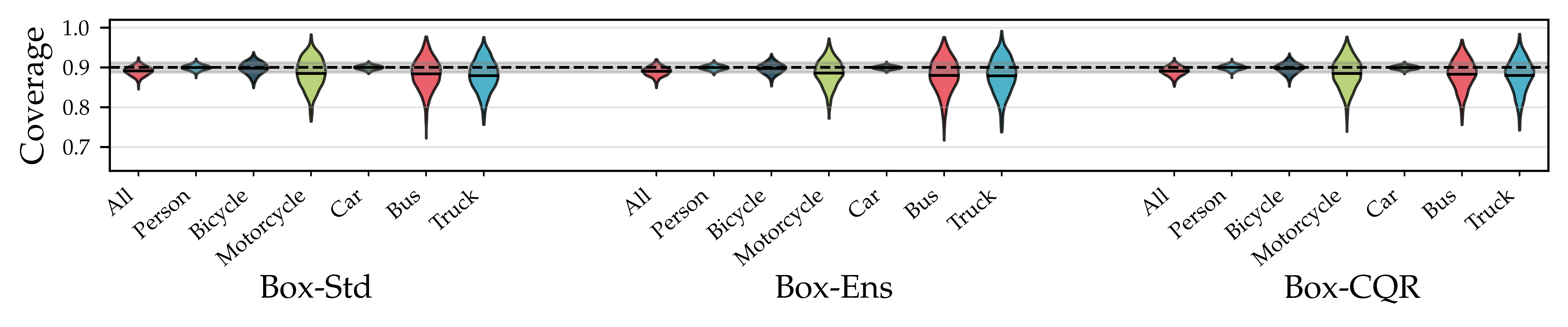}
        }
    \end{subfigure}%

    \begin{subfigure}[t]{.72\textwidth}
        \raisebox{-\height}{
            \includegraphics[width=\textwidth]{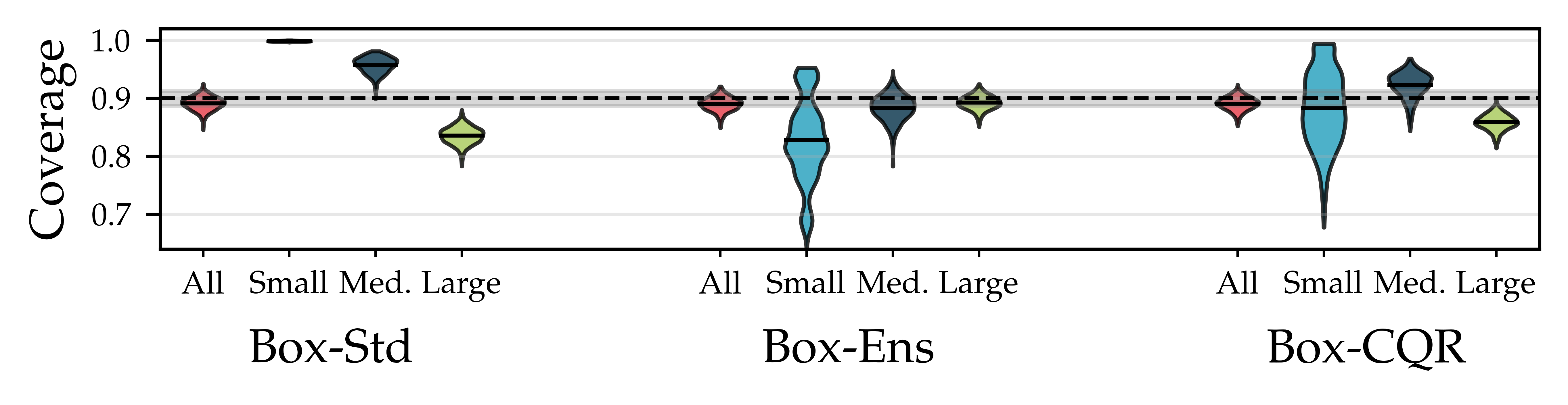}
        }
    \end{subfigure}%
    \hfill
    \begin{subfigure}[t]{.28\textwidth}
        \raisebox{-\height}{
            \includegraphics[width=\textwidth]{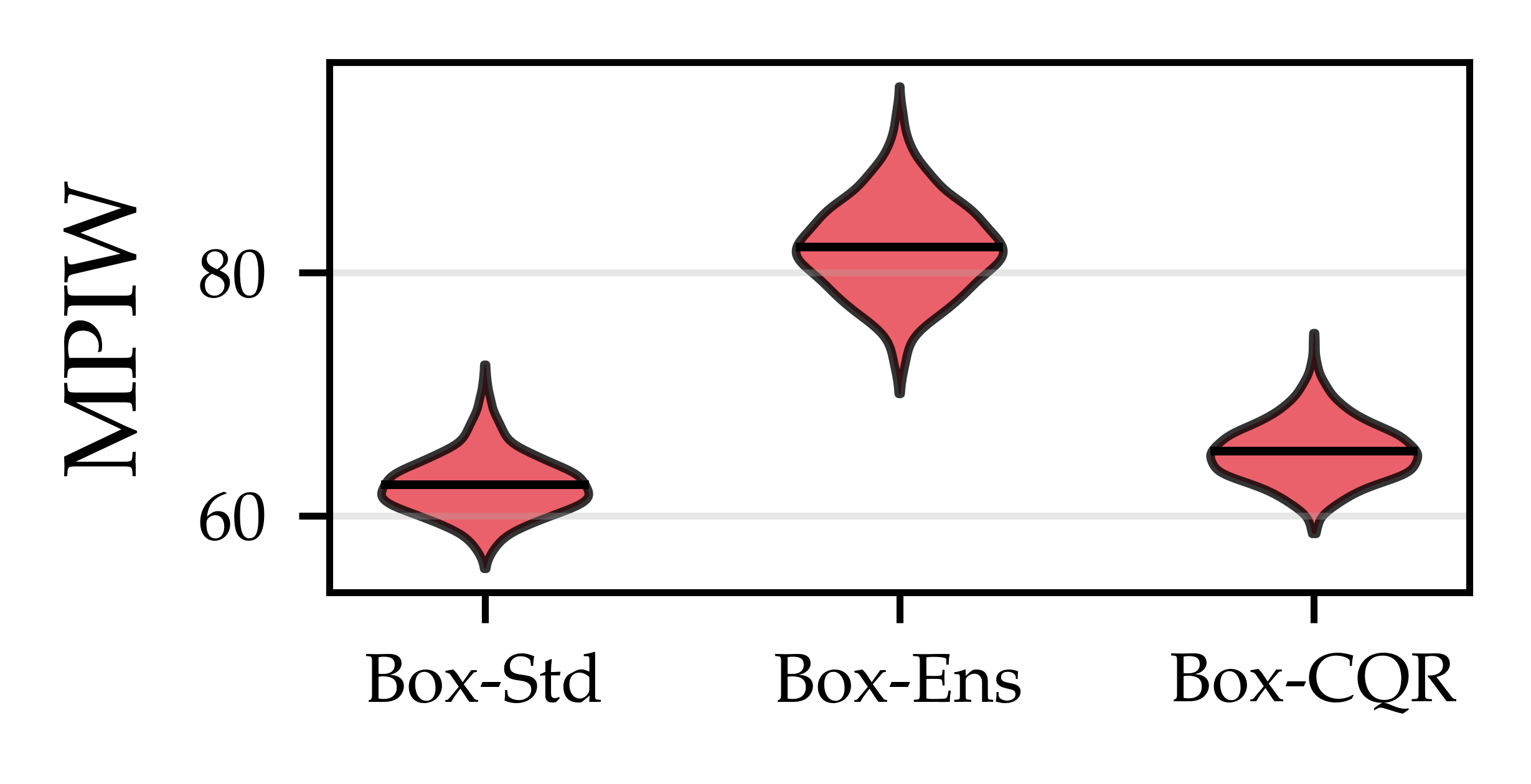}
        }
    \end{subfigure}

    \caption{\emph{Top:} Empirical coverage levels marginally across all objects (All) and across objects from selected classes for the three conformal bounding box methods on the Cityscapes dataset. \emph{Bottom:} Coverage levels stratified by object size (Small, Medium, Large) and $MPIW$. We also visualize target coverage (\includegraphics[width=2em]{fig/caption_line1.png}) and the marginal coverage distribution (\includegraphics[width=2em]{fig/caption_line2.png}). Displayed densities are results obtained over 1000 trials. For interpretation see \autoref{subsec:exp_results}.}
    \label{fig:app-violins-city}
\end{figure*}

\begin{figure*}[!ht]
    \centering  

    \begin{subfigure}[t]{\textwidth}
        \raisebox{-\height}{
            \includegraphics[width=\textwidth]{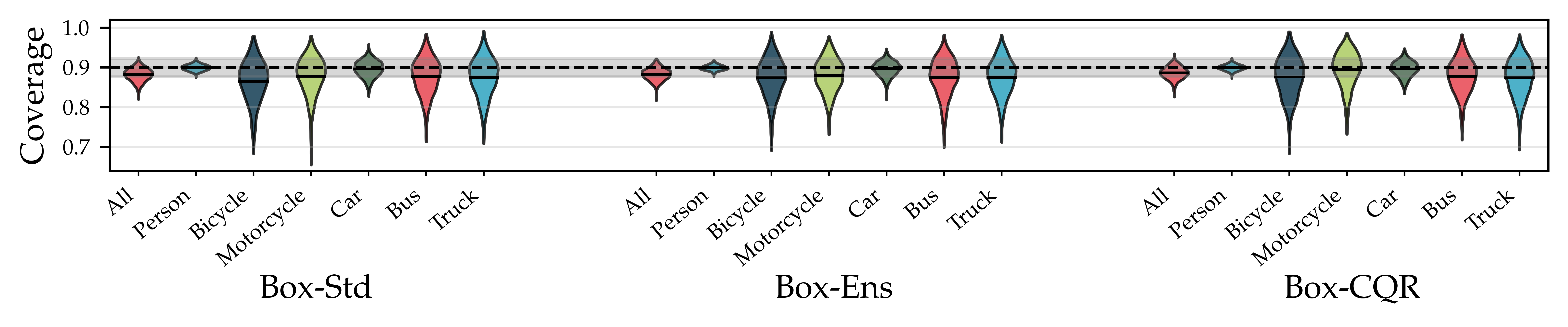}
        }
    \end{subfigure}%

    \begin{subfigure}[t]{.72\textwidth}
        \raisebox{-\height}{
            \includegraphics[width=\textwidth]{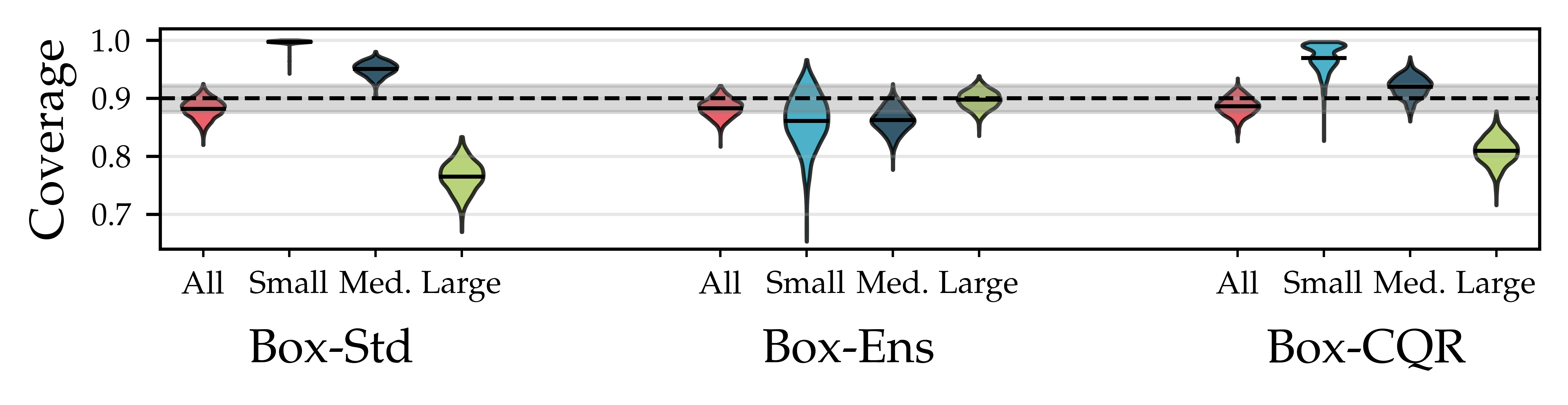}
        }
    \end{subfigure}%
    \hfill
    \begin{subfigure}[t]{.28\textwidth}
        \raisebox{-\height}{
            \includegraphics[width=\textwidth]{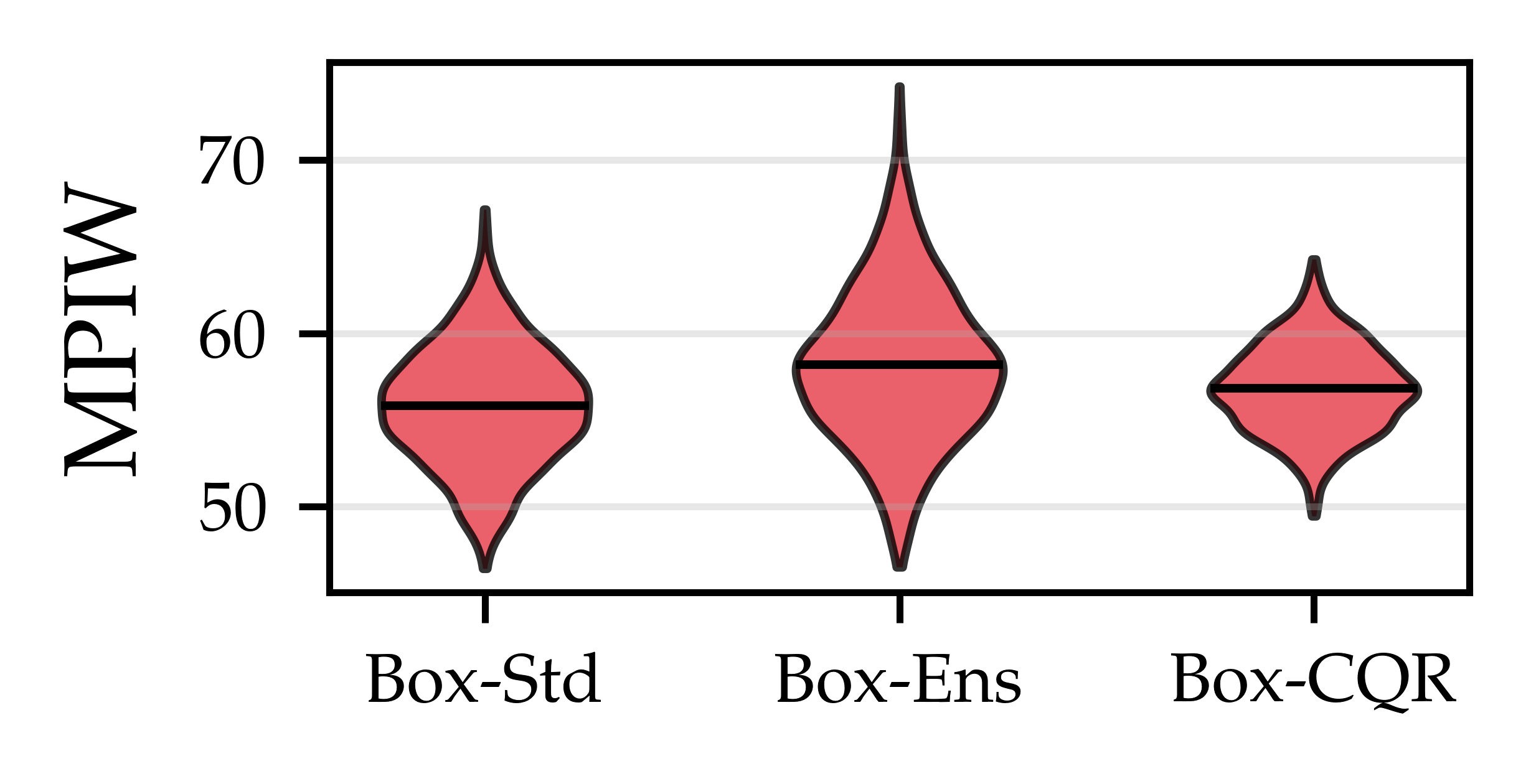}
        }
    \end{subfigure}

    \caption{\emph{Top:} Empirical coverage levels marginally across all objects (All) and across objects from selected classes for the three conformal bounding box methods on the Cityscapes dataset. \emph{Bottom:} Coverage levels stratified by object size (Small, Medium, Large) and $MPIW$. We also visualize target coverage (\includegraphics[width=2em]{fig/caption_line1.png}) and the marginal coverage distribution (\includegraphics[width=2em]{fig/caption_line2.png}). Displayed densities are results obtained over 1000 trials. For interpretation see \autoref{subsec:exp_results}.}
    \label{fig:app-violins-coco}
\end{figure*}

%% file: text/X_fig_pi.tex
\begin{figure*}
    
    \begin{subfigure}{.31\textwidth}
        \centering
        \resizebox{\textwidth}{!}{\includegraphics{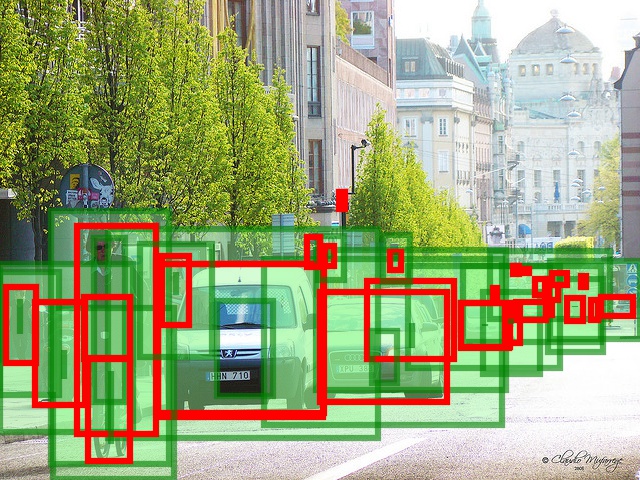}}
    \end{subfigure}%
    \hspace{.005\textwidth}%
    \begin{subfigure}{.31\textwidth}
        \centering
        \resizebox{\textwidth}{!}{\includegraphics{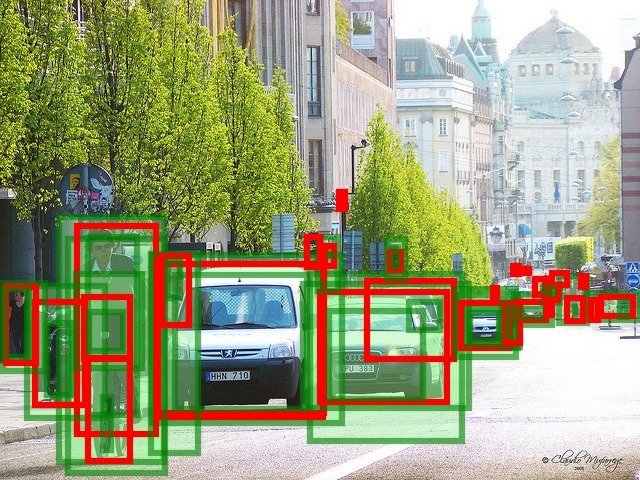}}
    \end{subfigure}%
    \hspace{.005\textwidth}%
    \begin{subfigure}{.31\textwidth}
        \centering
        \resizebox{\textwidth}{!}{\includegraphics{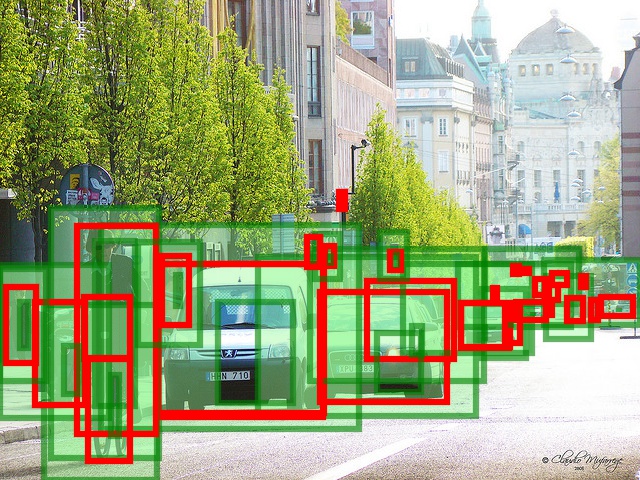}}
    \end{subfigure}
    \vspace{.005\textwidth}%
    \newline
    \begin{subfigure}{.31\textwidth}
        \centering
        \resizebox{\textwidth}{0.3\height}{\includegraphics[width=\textwidth, height=10cm]{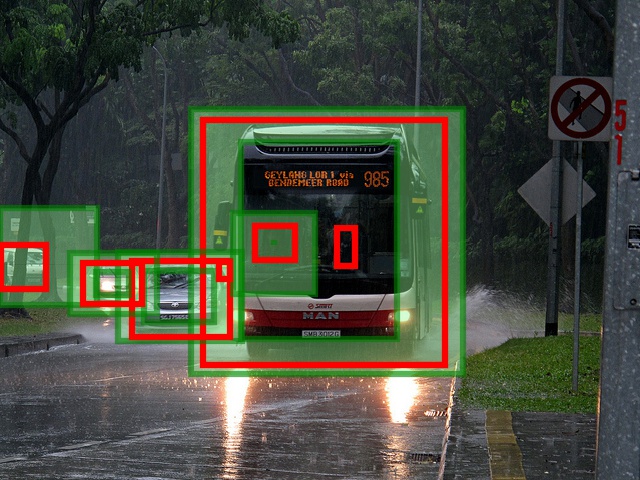}}
    \end{subfigure}%
    \hspace{.005\textwidth}%
    \begin{subfigure}{.31\textwidth}
        \centering
        \resizebox{\textwidth}{0.3\height}{\includegraphics[width=\textwidth, height=10cm]{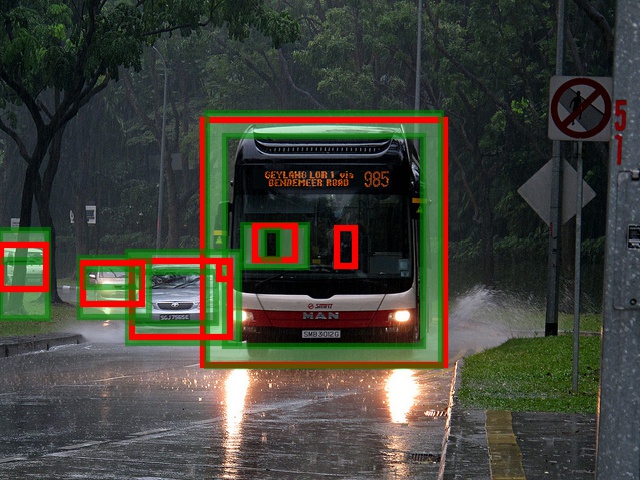}}
    \end{subfigure}%
    \hspace{.005\textwidth}%
    \begin{subfigure}{.31\textwidth}
        \centering
        \resizebox{\textwidth}{0.3\height}{\includegraphics[width=\textwidth, height=10cm]{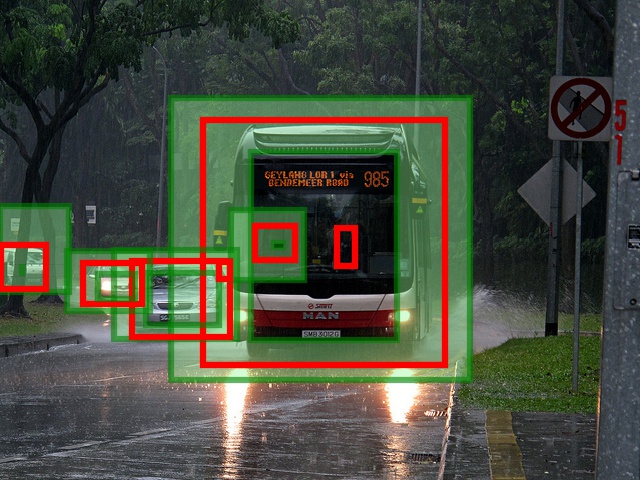}}
    \end{subfigure}
    \vspace{.005\textwidth}%
    \newline  
    \begin{subfigure}{.31\textwidth}
        \centering
        \resizebox{\textwidth}{!}{\includegraphics{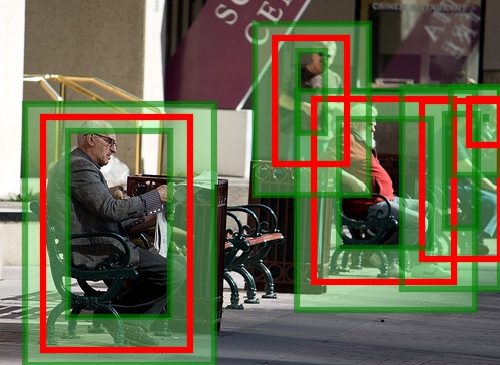}}
    \end{subfigure}%
    \hspace{.005\textwidth}%
    \begin{subfigure}{.31\textwidth}
        \centering
        \resizebox{\textwidth}{!}{\includegraphics{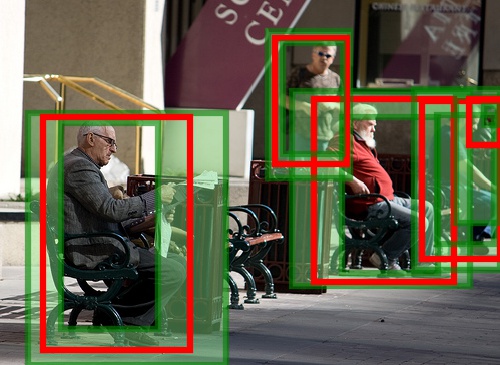}}
    \end{subfigure}%
    \hspace{.005\textwidth}%
    \begin{subfigure}{.31\textwidth}
        \centering
        \resizebox{\textwidth}{!}{\includegraphics{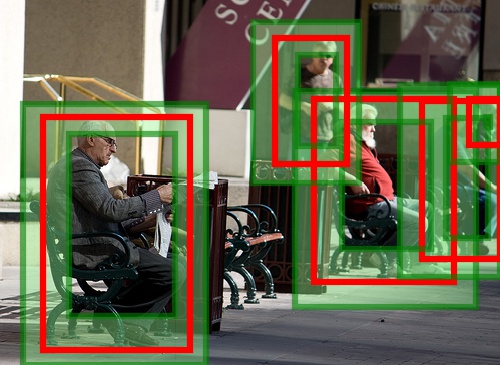}}
    \end{subfigure}
    \vspace{.005\textwidth}%
    \newline  
    \begin{subfigure}{.31\textwidth}
        \centering
        \resizebox{\textwidth}{!}{\includegraphics{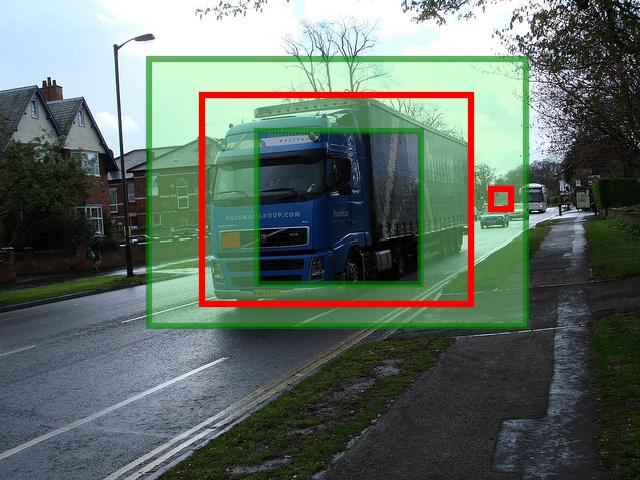}}
    \end{subfigure}%
    \hspace{.005\textwidth}%
    \begin{subfigure}{.31\textwidth}
        \centering
        \resizebox{\textwidth}{!}{\includegraphics{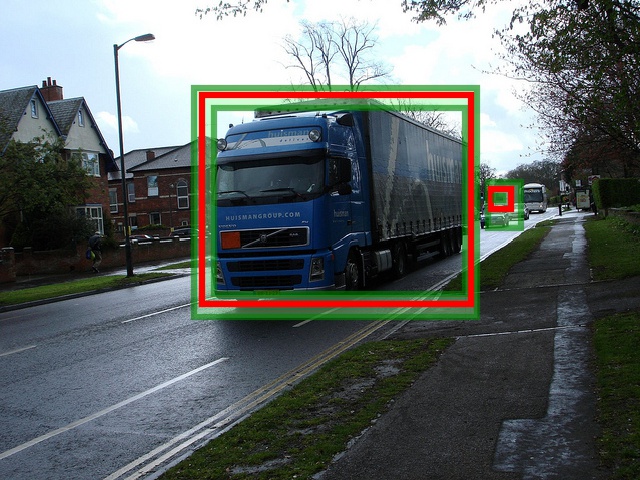}}
    \end{subfigure}%
    \hspace{.005\textwidth}%
    \begin{subfigure}{.31\textwidth}
        \centering
        \resizebox{\textwidth}{!}{\includegraphics{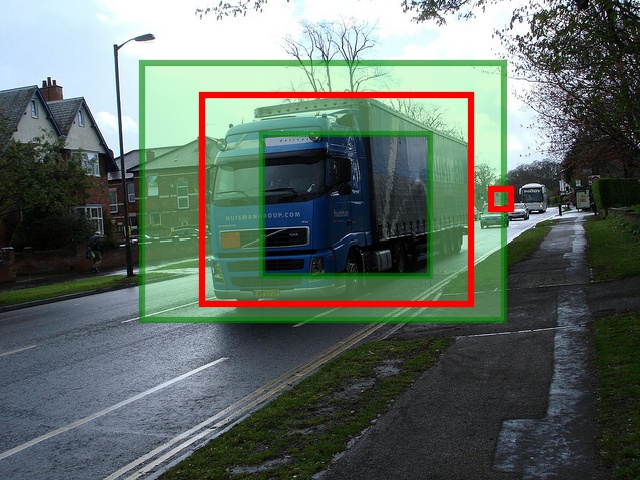}}
    \end{subfigure}
    
    \caption{Examples of conformal bounding box intervals produced by our two-step approach on COCO for a mixed set of classes. \emph{Left to right by column}: using ClassThr in combination with Box-Std, Box-Ens or Box-CQR. True bounding boxes are in red, two-sided prediction interval regions are shaded in green.}
    \label{fig:pi-plots-coco}
\end{figure*}

\begin{figure*}
    
    \begin{subfigure}{.31\textwidth}
        \centering
        \resizebox{\textwidth}{!}{\includegraphics{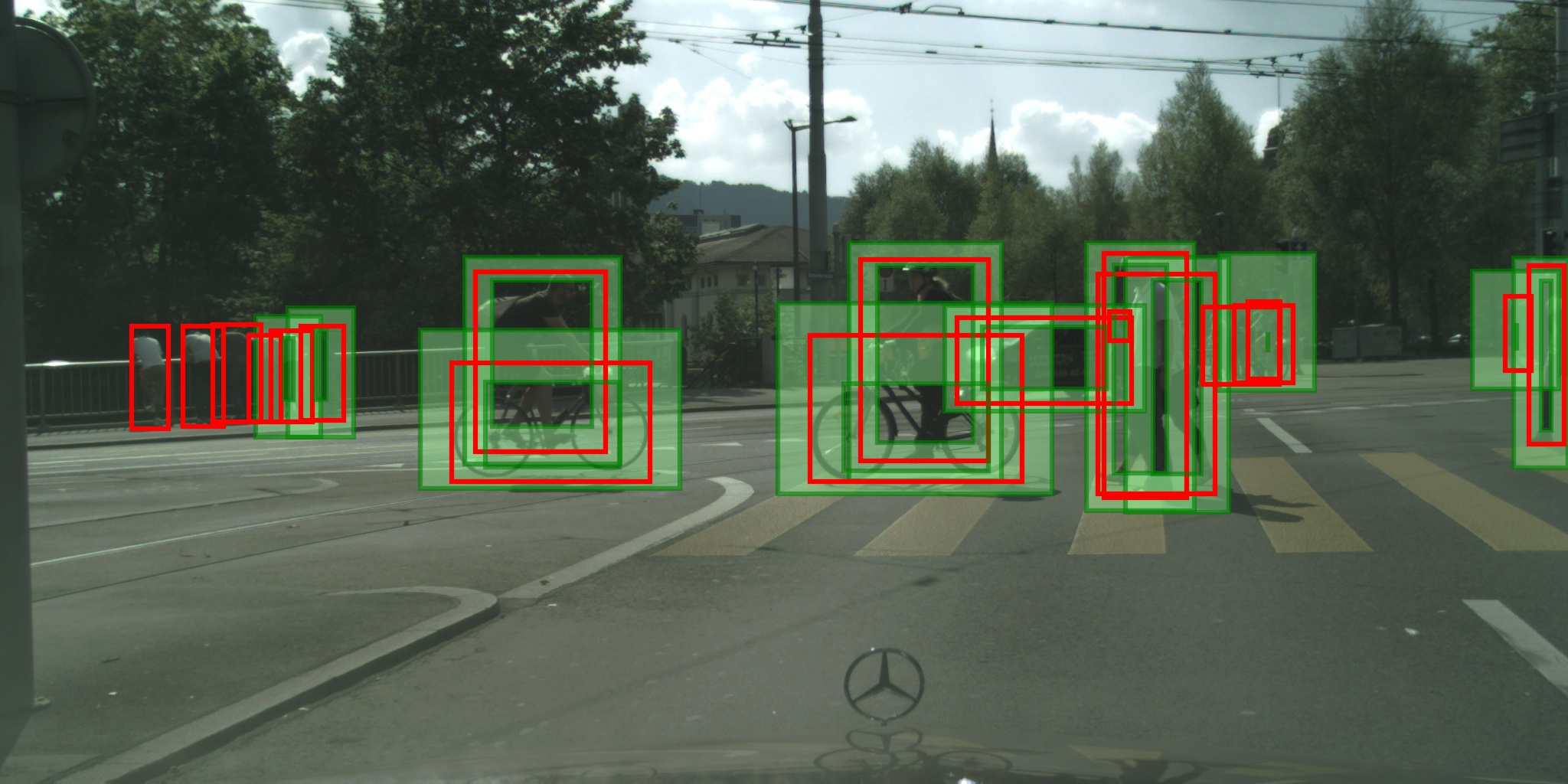}}
    \end{subfigure}%
    \hspace{.005\textwidth}%
    \begin{subfigure}{.31\textwidth}
        \centering
        \resizebox{\textwidth}{!}{\includegraphics{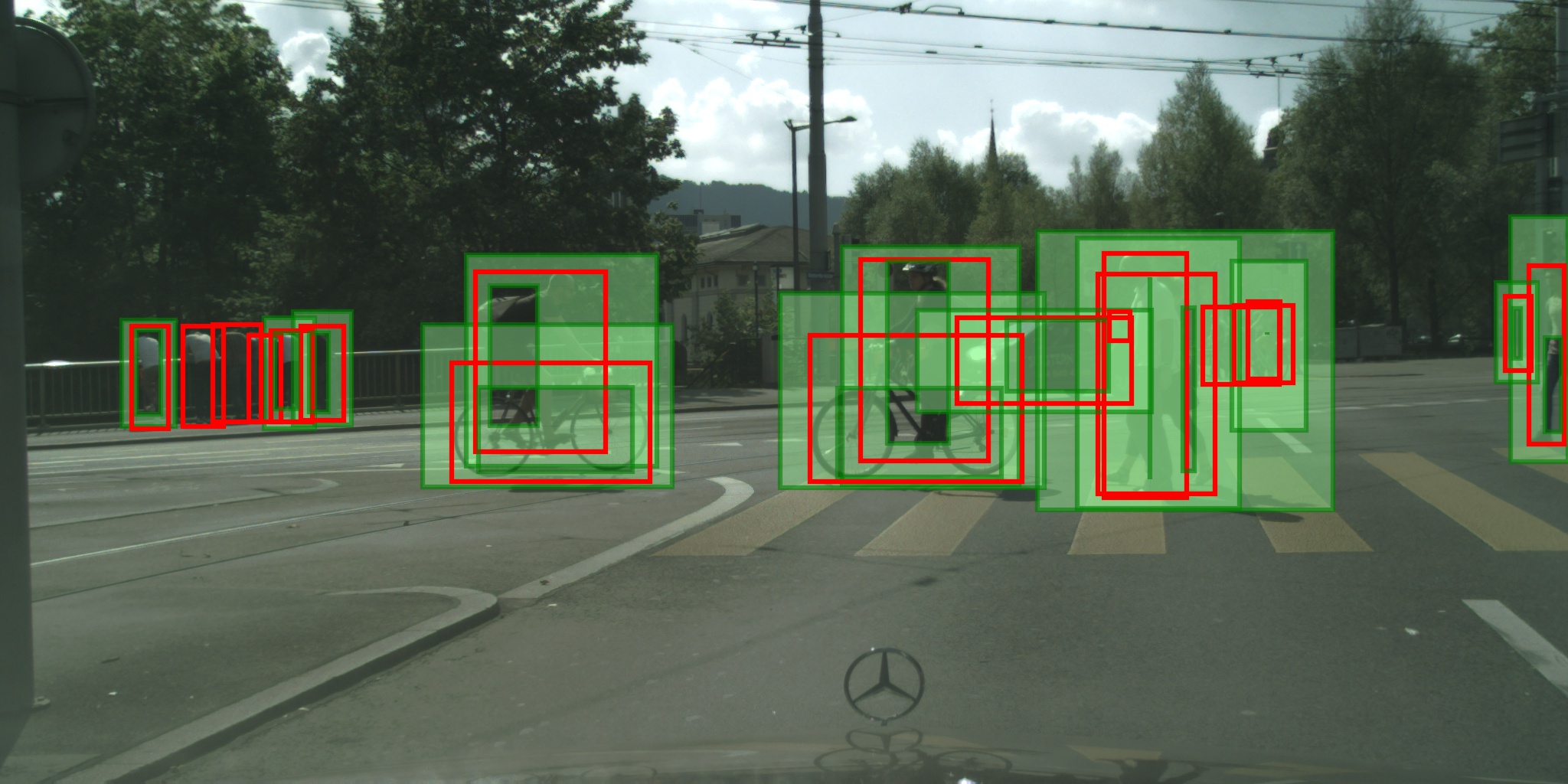}}
    \end{subfigure}%
    \hspace{.005\textwidth}%
    \begin{subfigure}{.31\textwidth}
        \centering
        \resizebox{\textwidth}{!}{\includegraphics{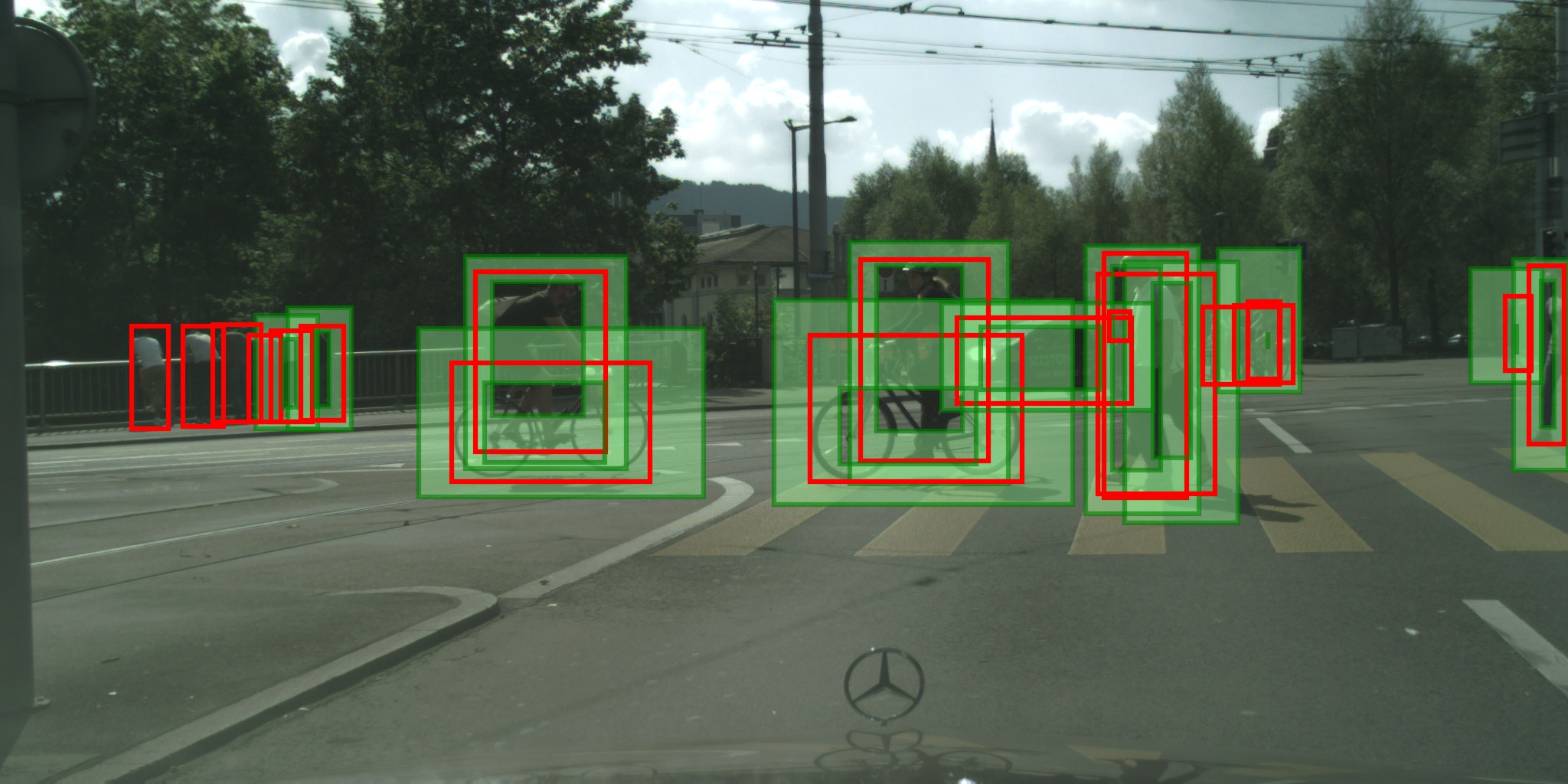}}
    \end{subfigure}
    \vspace{.005\textwidth}%
    \newline
    \begin{subfigure}{.31\textwidth}
        \centering
        \resizebox{\textwidth}{!}{\includegraphics{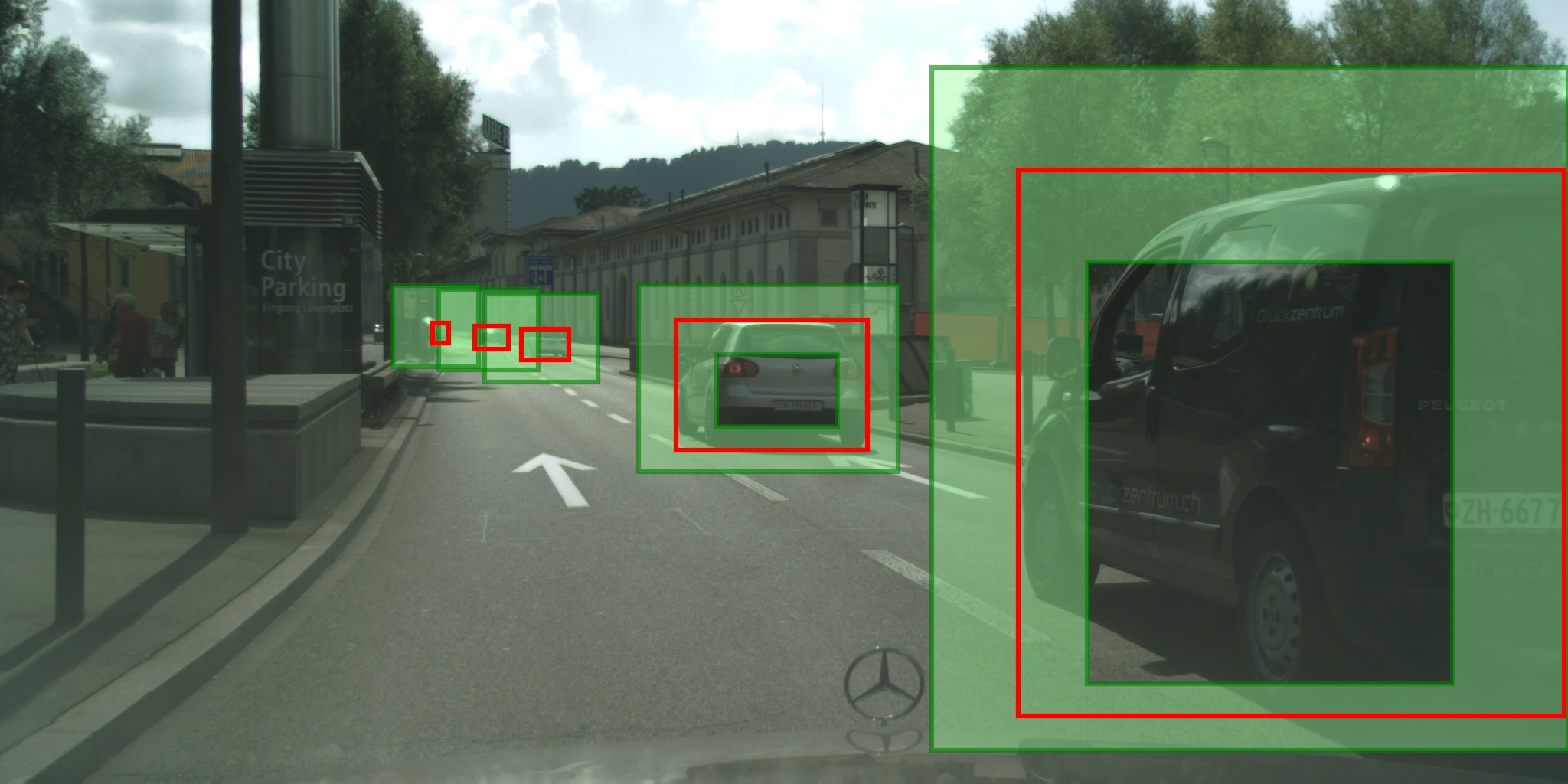}}
    \end{subfigure}%
    \hspace{.005\textwidth}%
    \begin{subfigure}{.31\textwidth}
        \centering
        \resizebox{\textwidth}{!}{\includegraphics{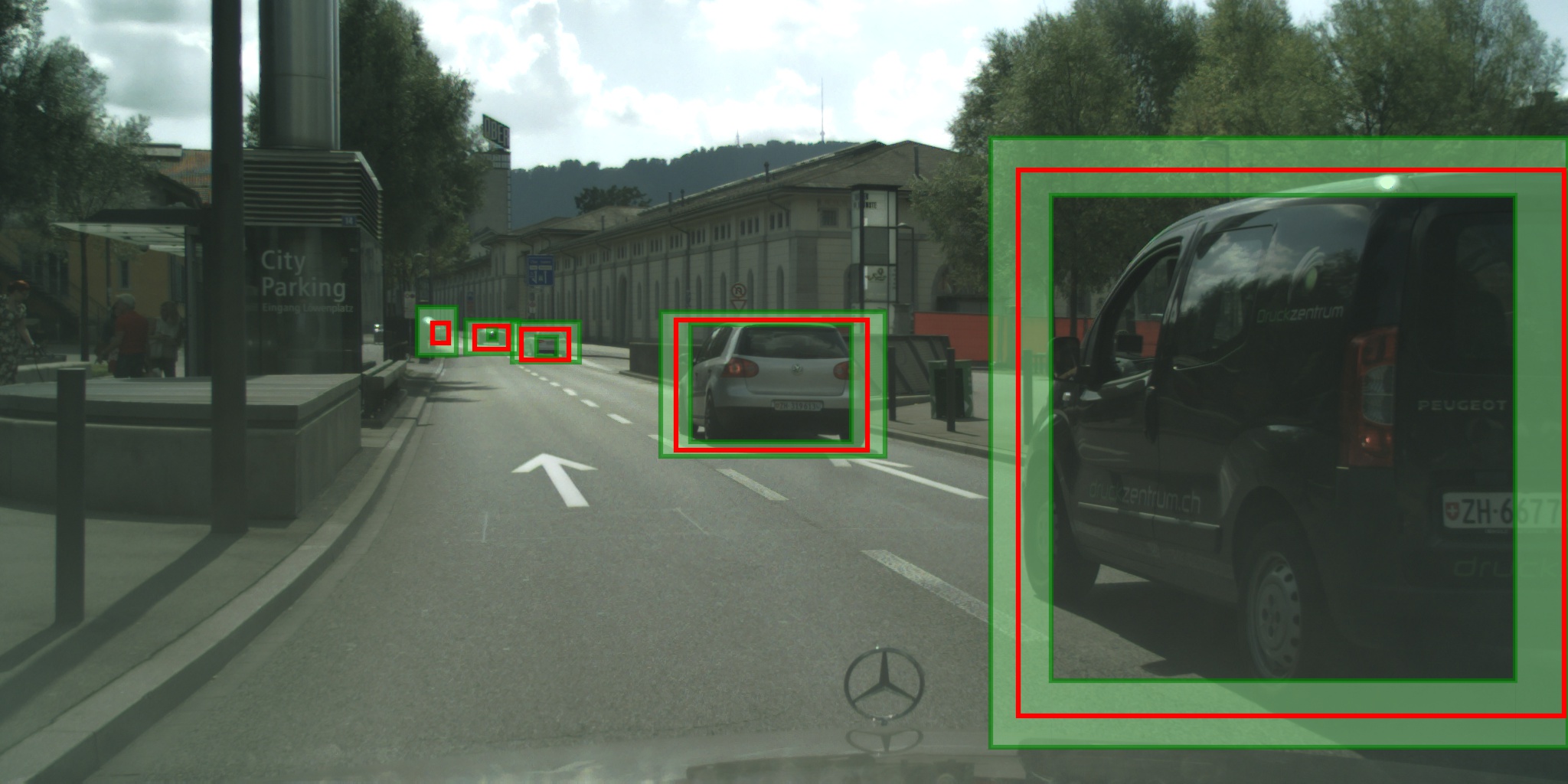}}
    \end{subfigure}%
    \hspace{.005\textwidth}%
    \begin{subfigure}{.31\textwidth}
        \centering
        \resizebox{\textwidth}{!}{\includegraphics{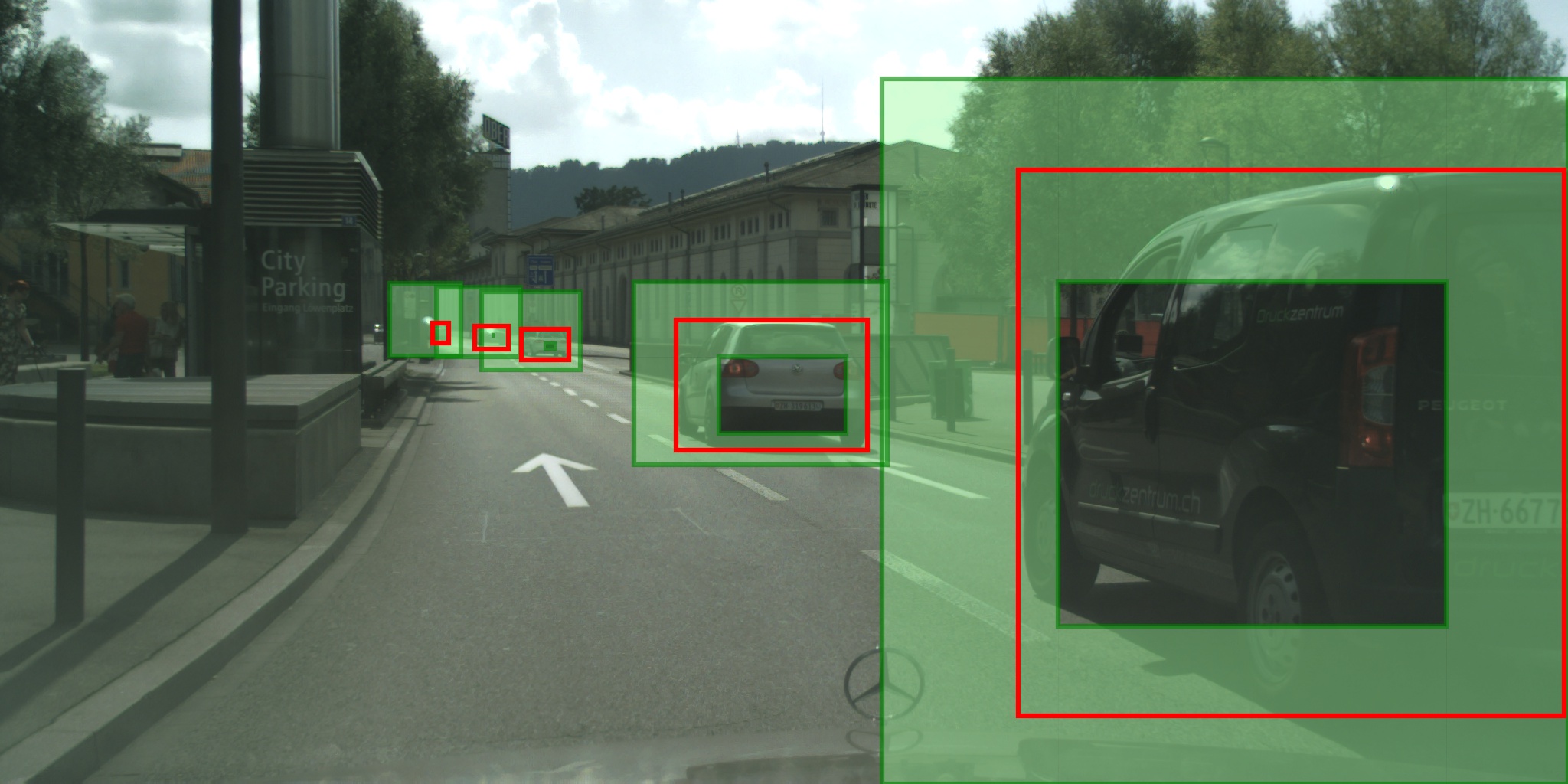}}
    \end{subfigure}
    \vspace{.005\textwidth}%
    \newline  
    \begin{subfigure}{.31\textwidth}
        \centering
        \resizebox{\textwidth}{!}{\includegraphics{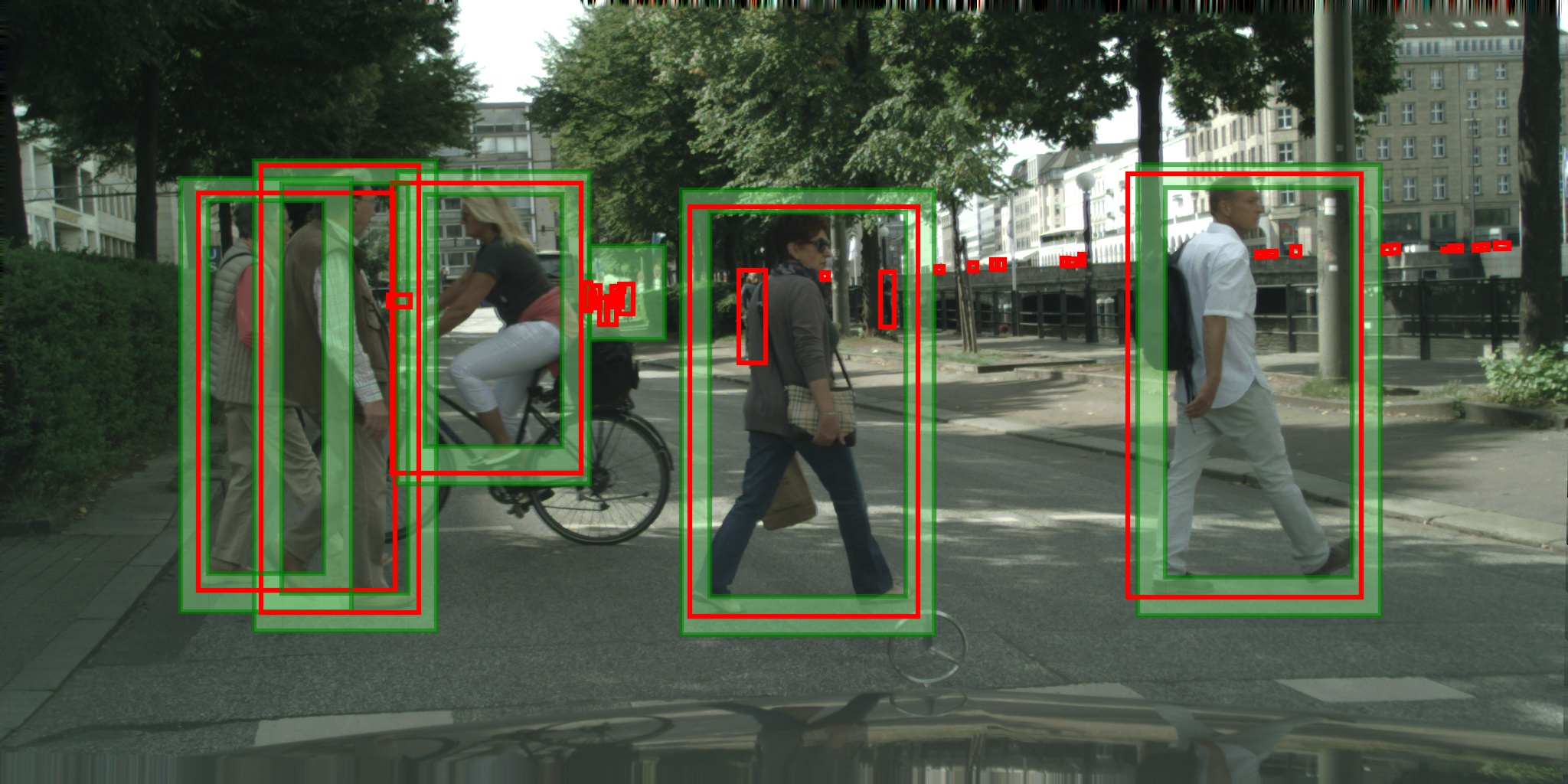}}
    \end{subfigure}%
    \hspace{.005\textwidth}%
    \begin{subfigure}{.31\textwidth}
        \centering
        \resizebox{\textwidth}{!}{\includegraphics{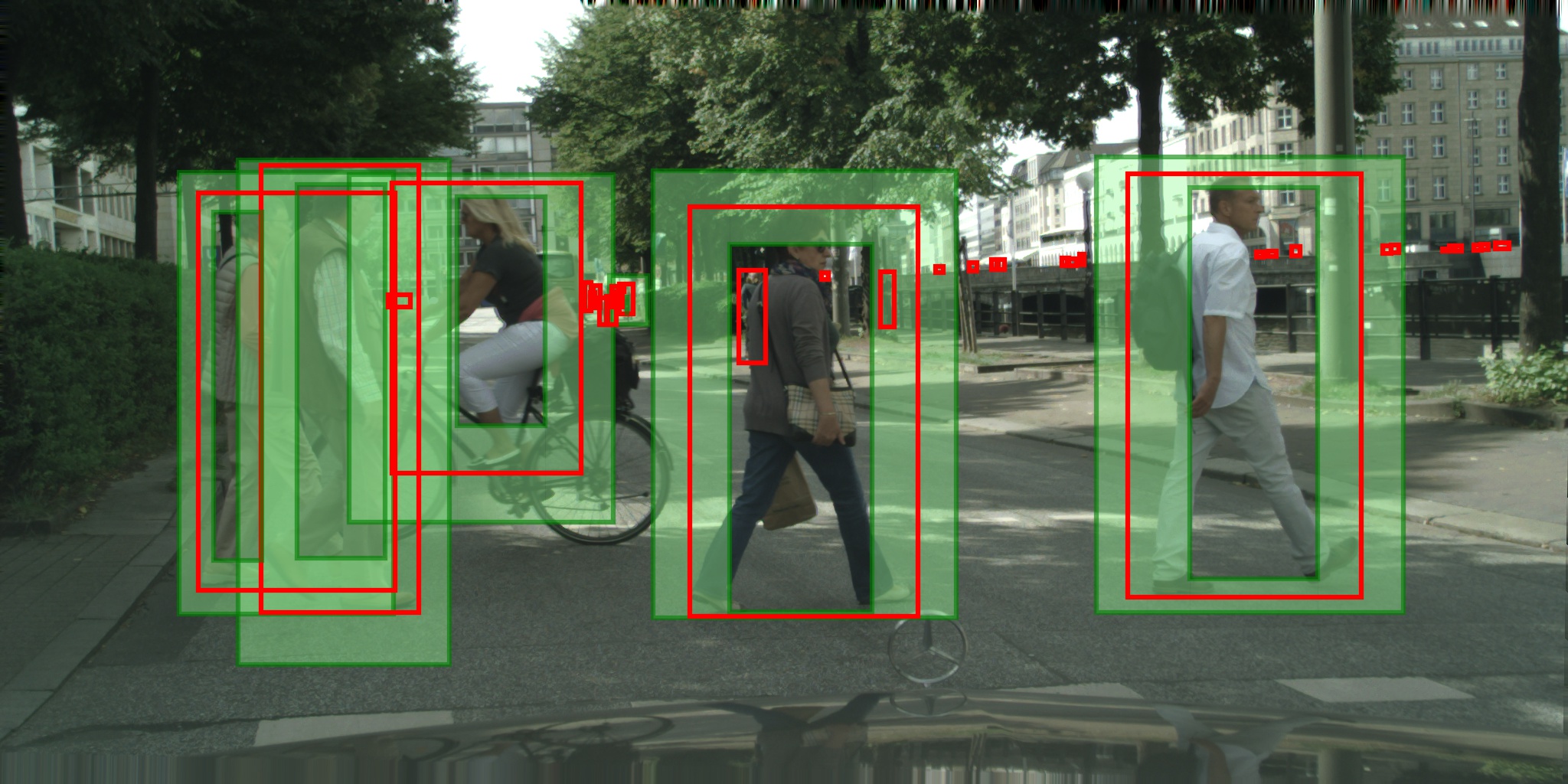}}
    \end{subfigure}%
    \hspace{.005\textwidth}%
    \begin{subfigure}{.31\textwidth}
        \centering
        \resizebox{\textwidth}{!}{\includegraphics{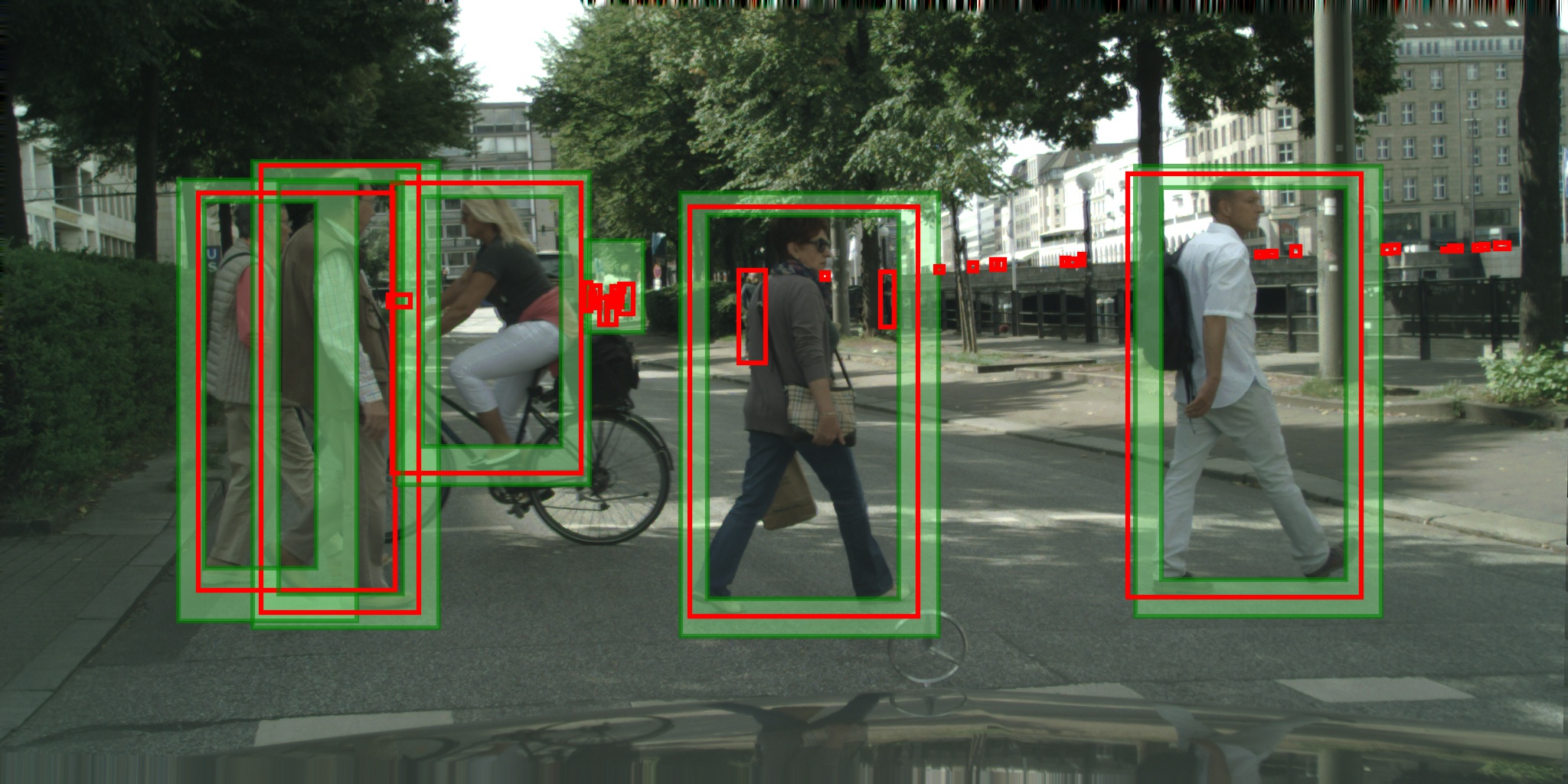}}
    \end{subfigure}
    \vspace{.005\textwidth}%
    \newline  
    \begin{subfigure}{.31\textwidth}
        \centering
        \resizebox{\textwidth}{!}{\includegraphics{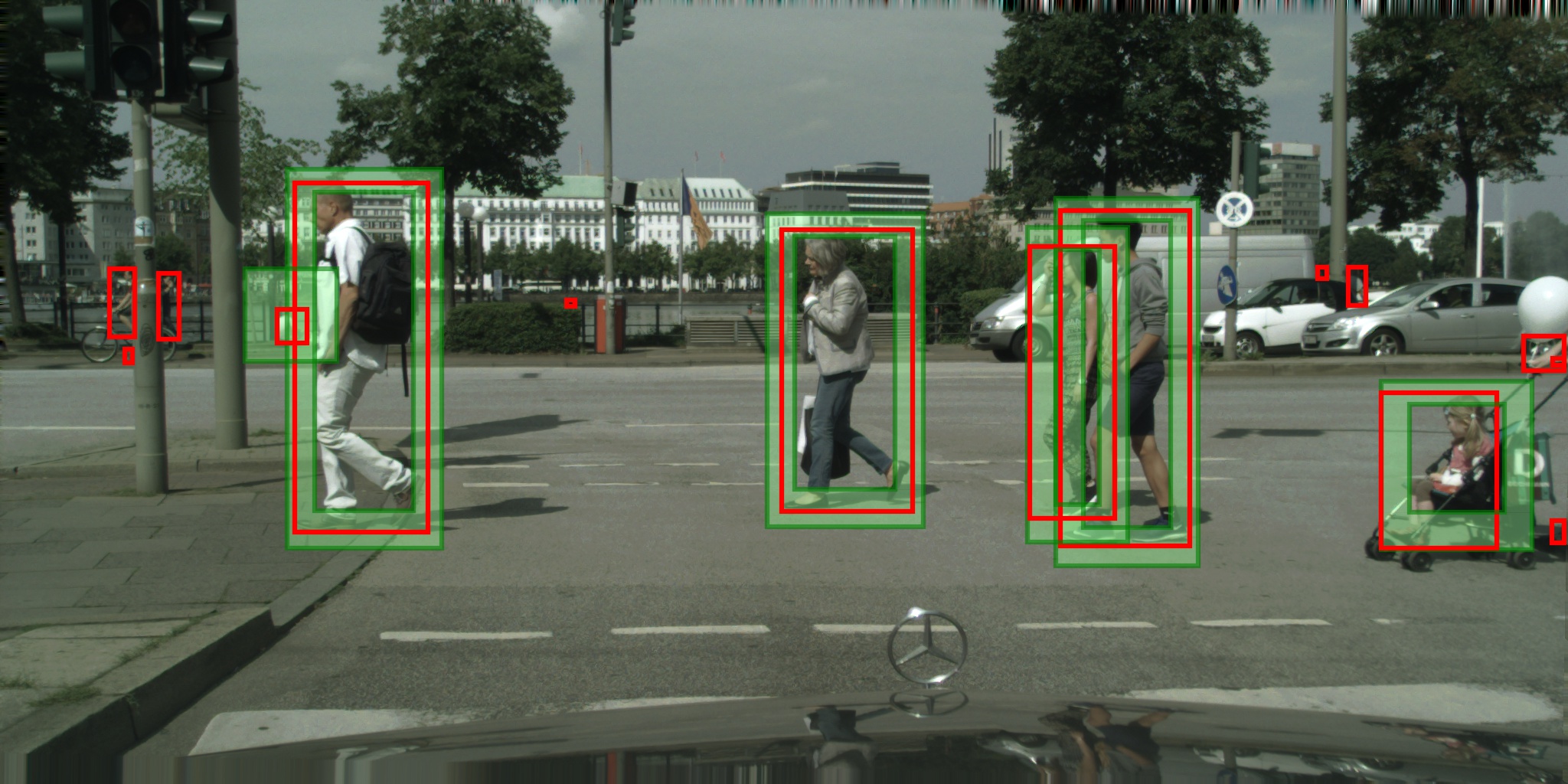}}
    \end{subfigure}%
    \hspace{.005\textwidth}%
    \begin{subfigure}{.31\textwidth}
        \centering
        \resizebox{\textwidth}{!}{\includegraphics{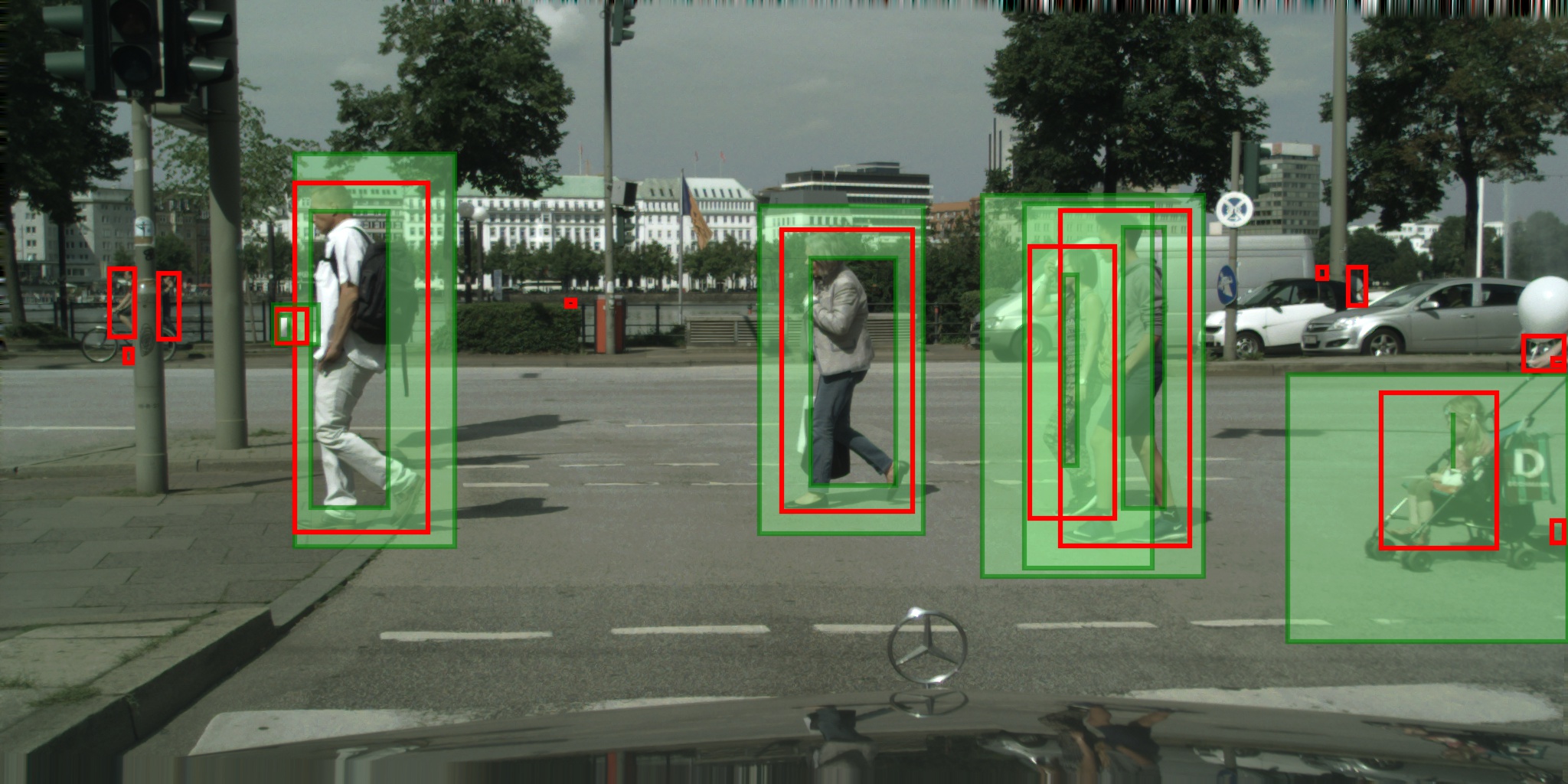}}
    \end{subfigure}%
    \hspace{.005\textwidth}%
    \begin{subfigure}{.31\textwidth}
        \centering
        \resizebox{\textwidth}{!}{\includegraphics{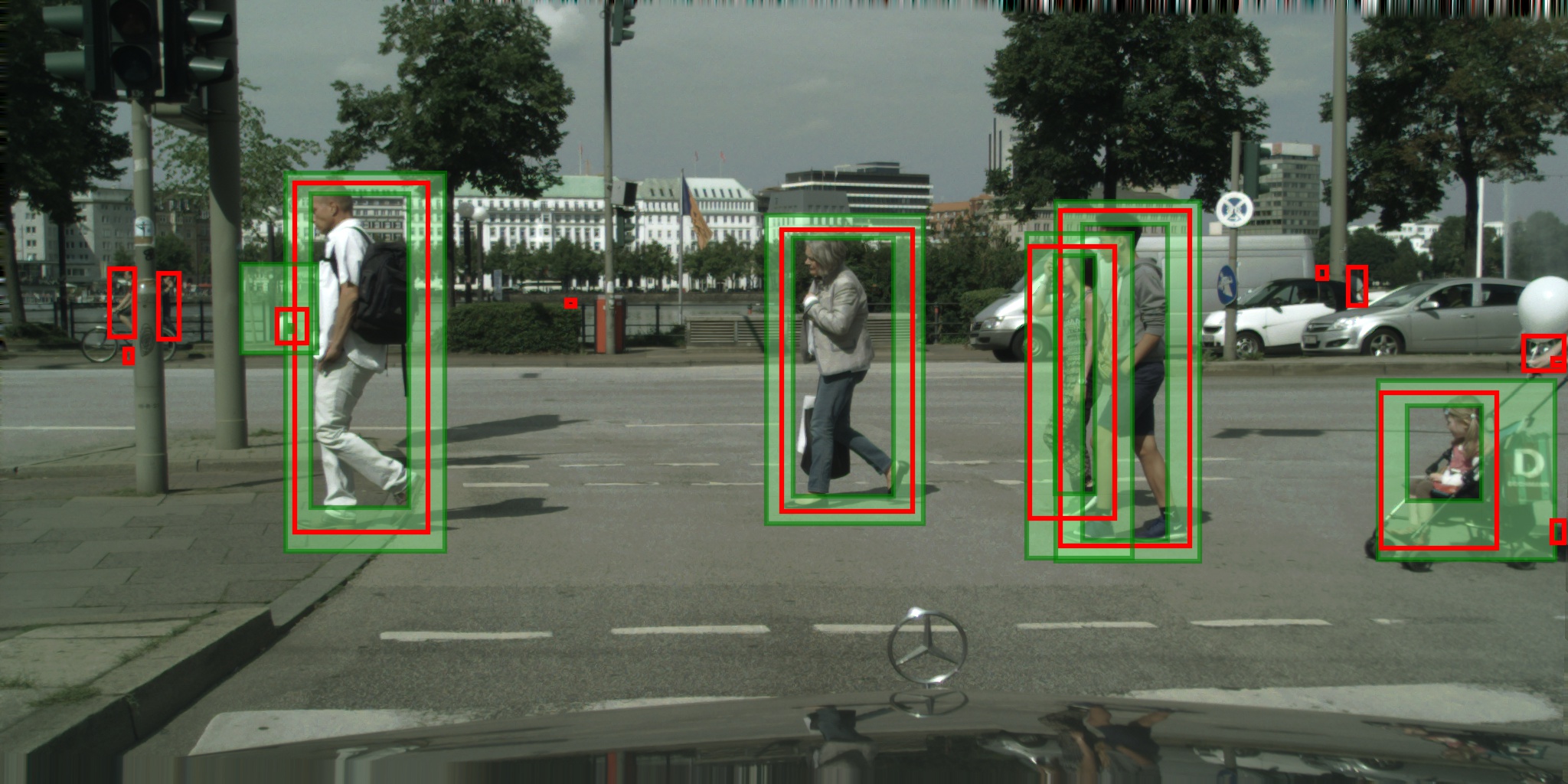}}
    \end{subfigure}
    
    \caption{Examples of conformal bounding box intervals produced by our two-step approach on Cityscapes for a mixed set of classes. \emph{Left to right by column}: using ClassThr in combination with Box-Std, Box-Ens or Box-CQR. True bounding boxes are in red, two-sided prediction interval regions are shaded in green.}
    \label{fig:pi-plots-cityscapes}
\end{figure*}

\begin{figure*}
    
    \begin{subfigure}{.31\textwidth}
        \centering
        \resizebox{\textwidth}{!}{\includegraphics{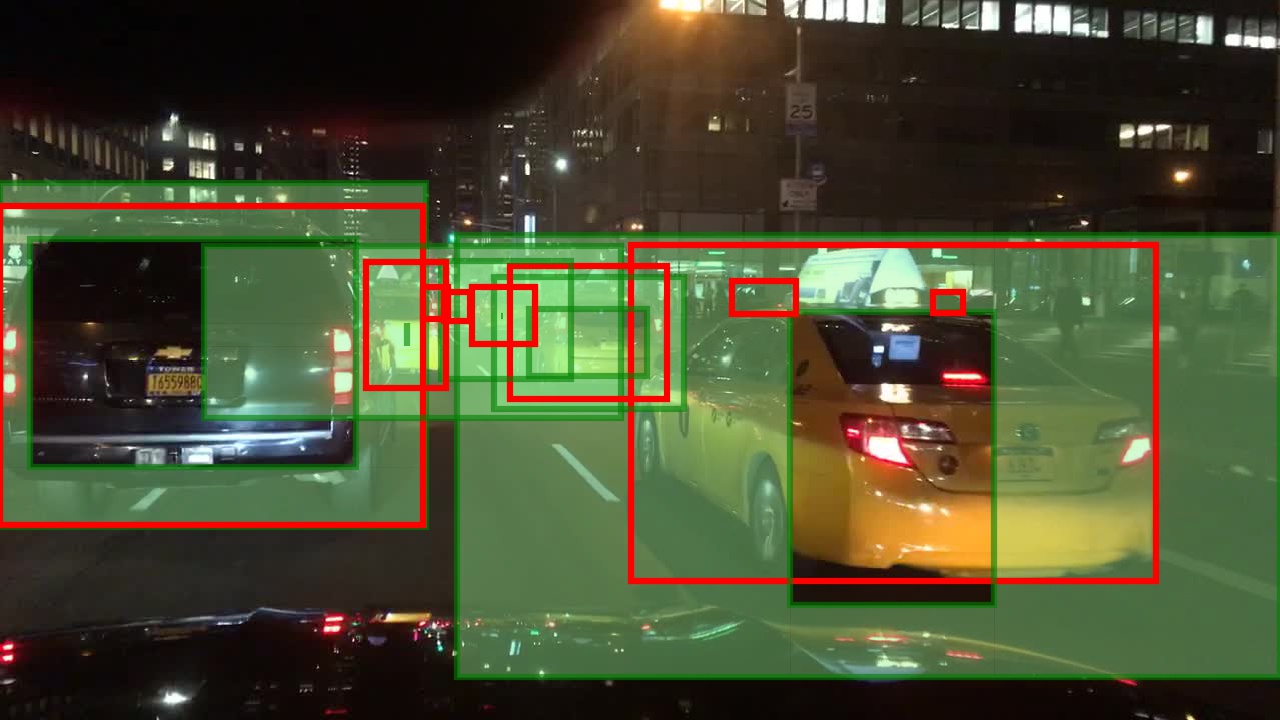}}
    \end{subfigure}%
    \hspace{.005\textwidth}%
    \begin{subfigure}{.31\textwidth}
        \centering
        \resizebox{\textwidth}{!}{\includegraphics{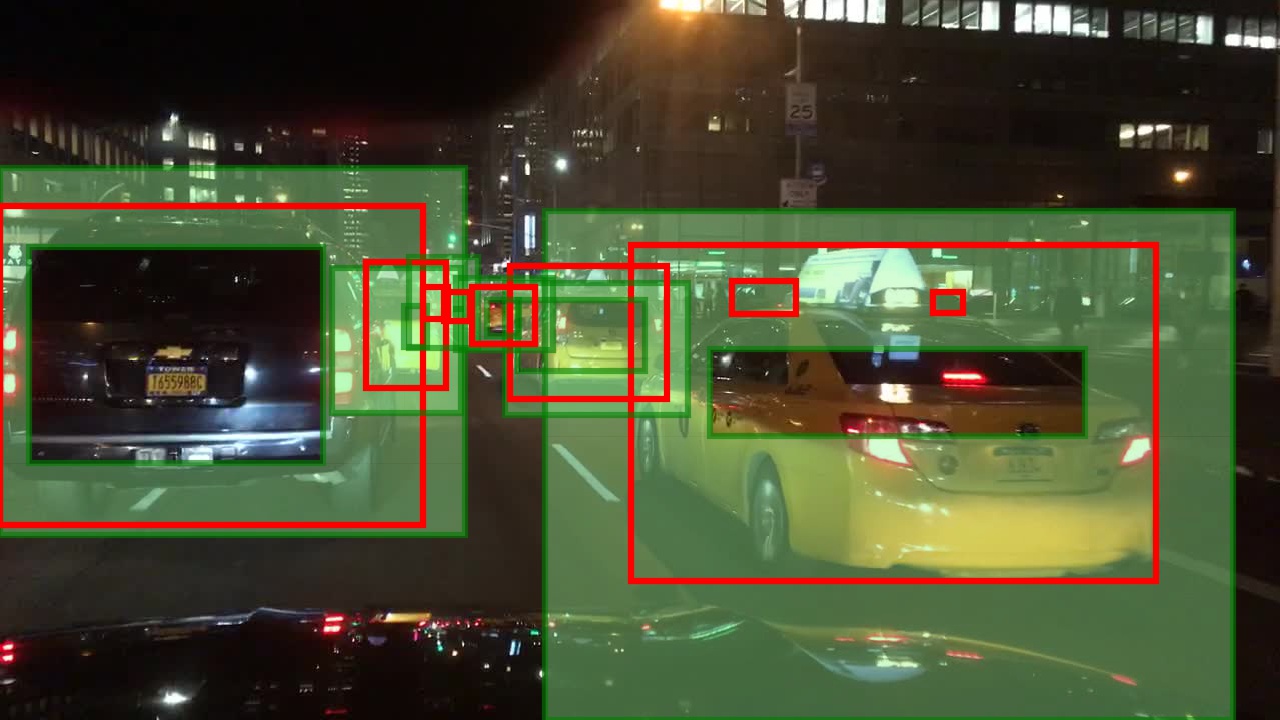}}
    \end{subfigure}%
    \hspace{.005\textwidth}%
    \begin{subfigure}{.31\textwidth}
        \centering
        \resizebox{\textwidth}{!}{\includegraphics{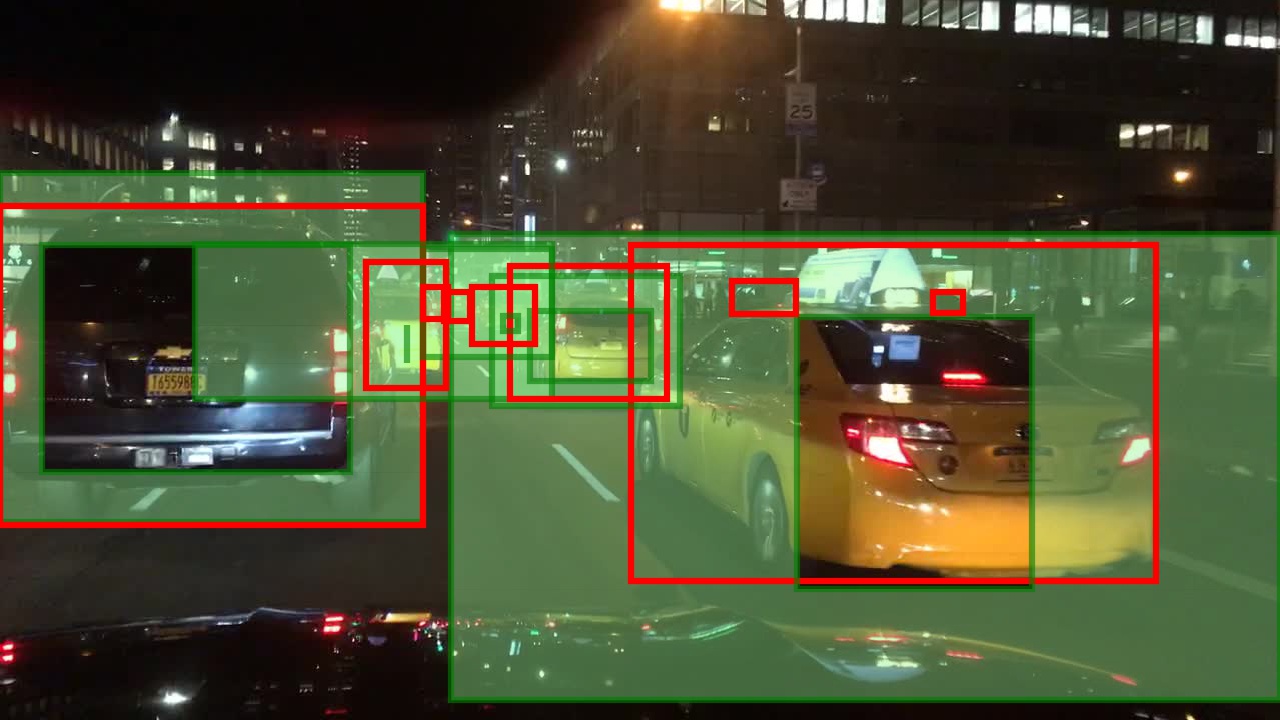}}
    \end{subfigure}
    \vspace{.005\textwidth}%
    \newline
    \begin{subfigure}{.31\textwidth}
        \centering
        \resizebox{\textwidth}{!}{\includegraphics{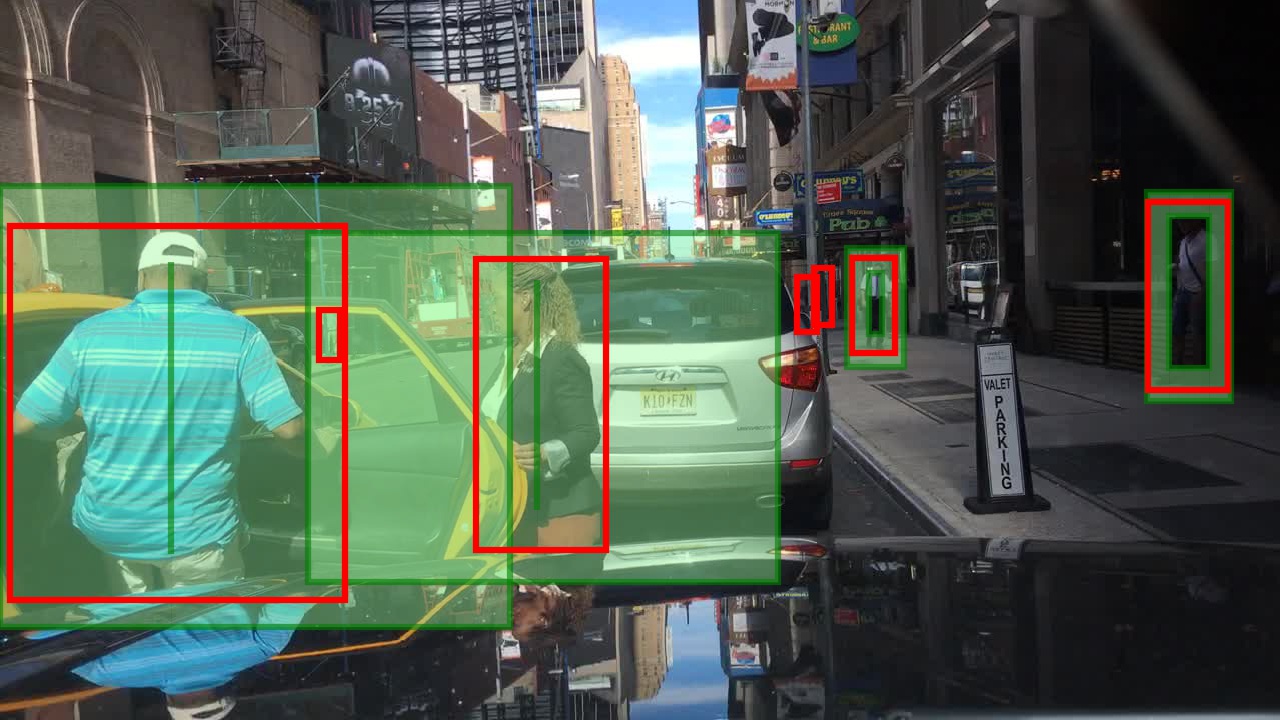}}
    \end{subfigure}%
    \hspace{.005\textwidth}%
    \begin{subfigure}{.31\textwidth}
        \centering
        \resizebox{\textwidth}{!}{\includegraphics{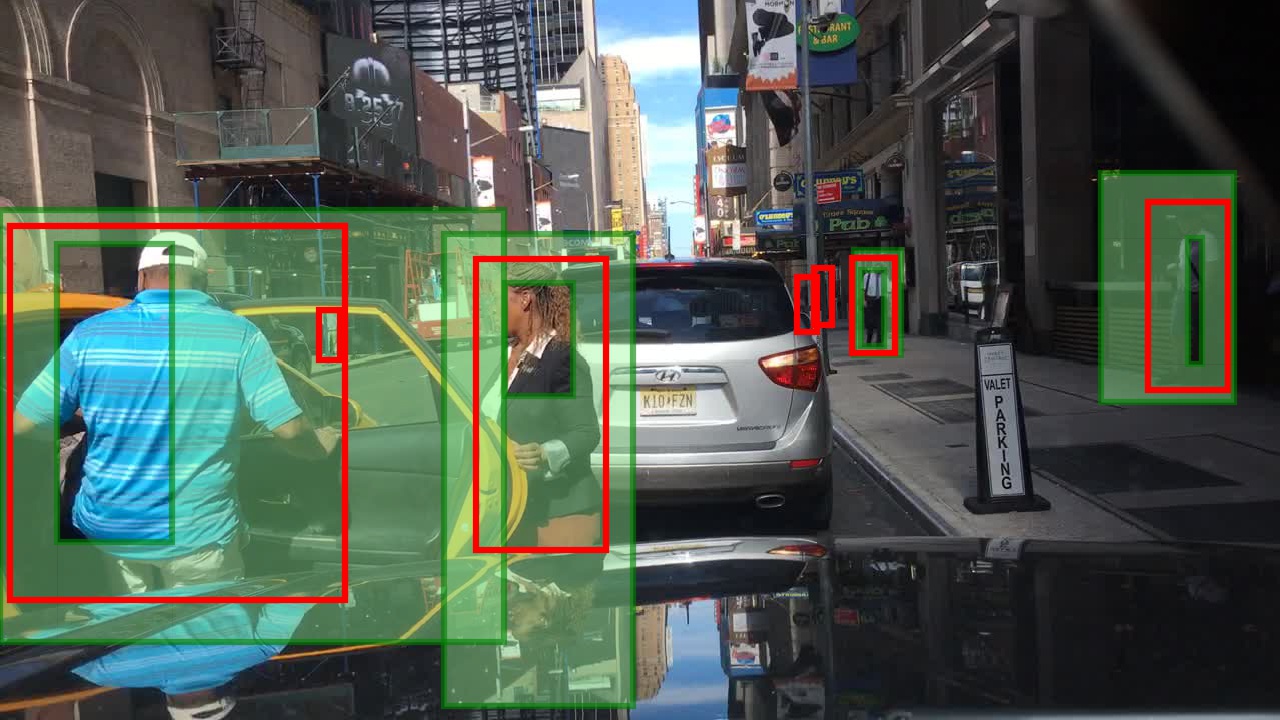}}
    \end{subfigure}%
    \hspace{.005\textwidth}%
    \begin{subfigure}{.31\textwidth}
        \centering
        \resizebox{\textwidth}{!}{\includegraphics{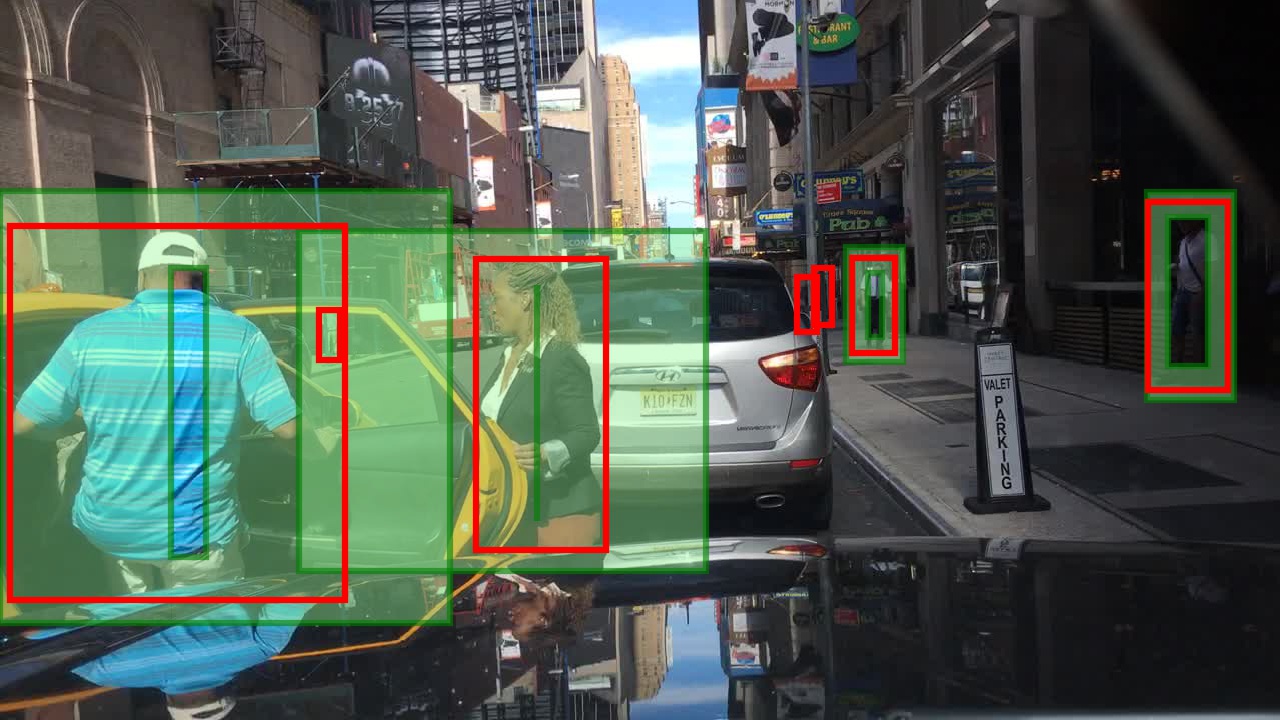}}
    \end{subfigure}
    \vspace{.005\textwidth}%
    \newline  
    \begin{subfigure}{.31\textwidth}
        \centering
        \resizebox{\textwidth}{!}{\includegraphics{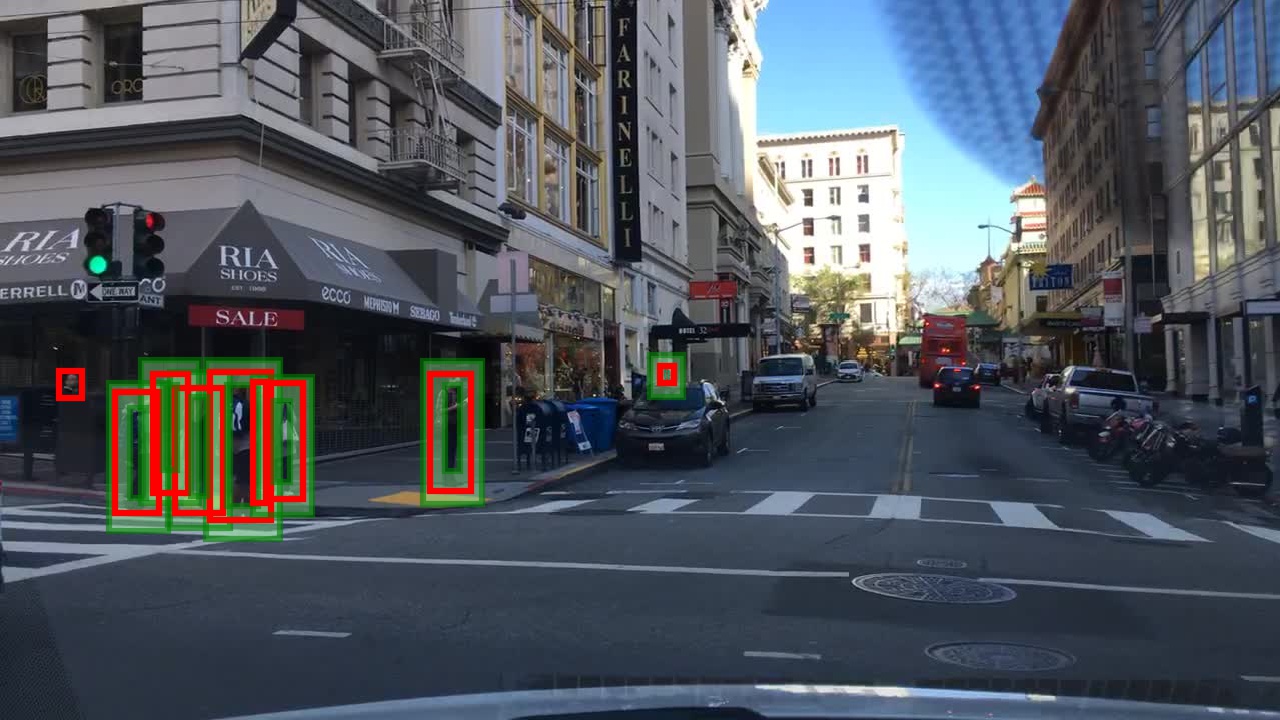}}
    \end{subfigure}%
    \hspace{.005\textwidth}%
    \begin{subfigure}{.31\textwidth}
        \centering
        \resizebox{\textwidth}{!}{\includegraphics{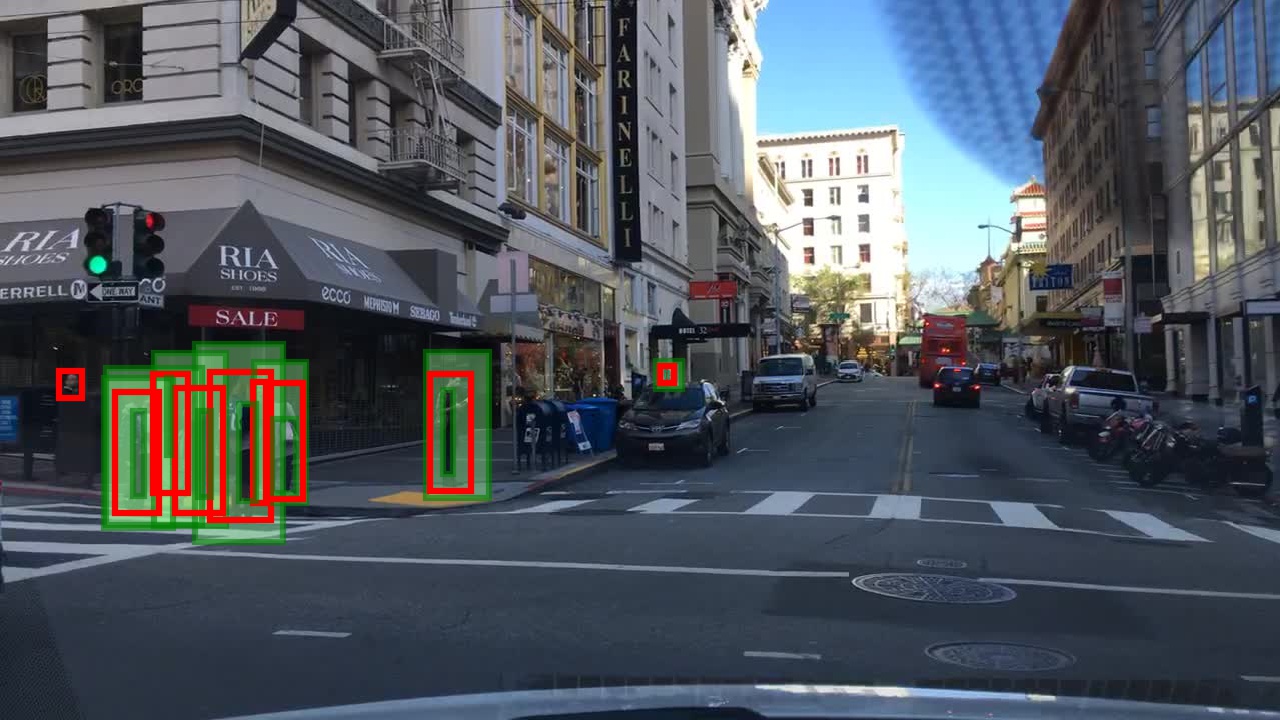}}
    \end{subfigure}%
    \hspace{.005\textwidth}%
    \begin{subfigure}{.31\textwidth}
        \centering
        \resizebox{\textwidth}{!}{\includegraphics{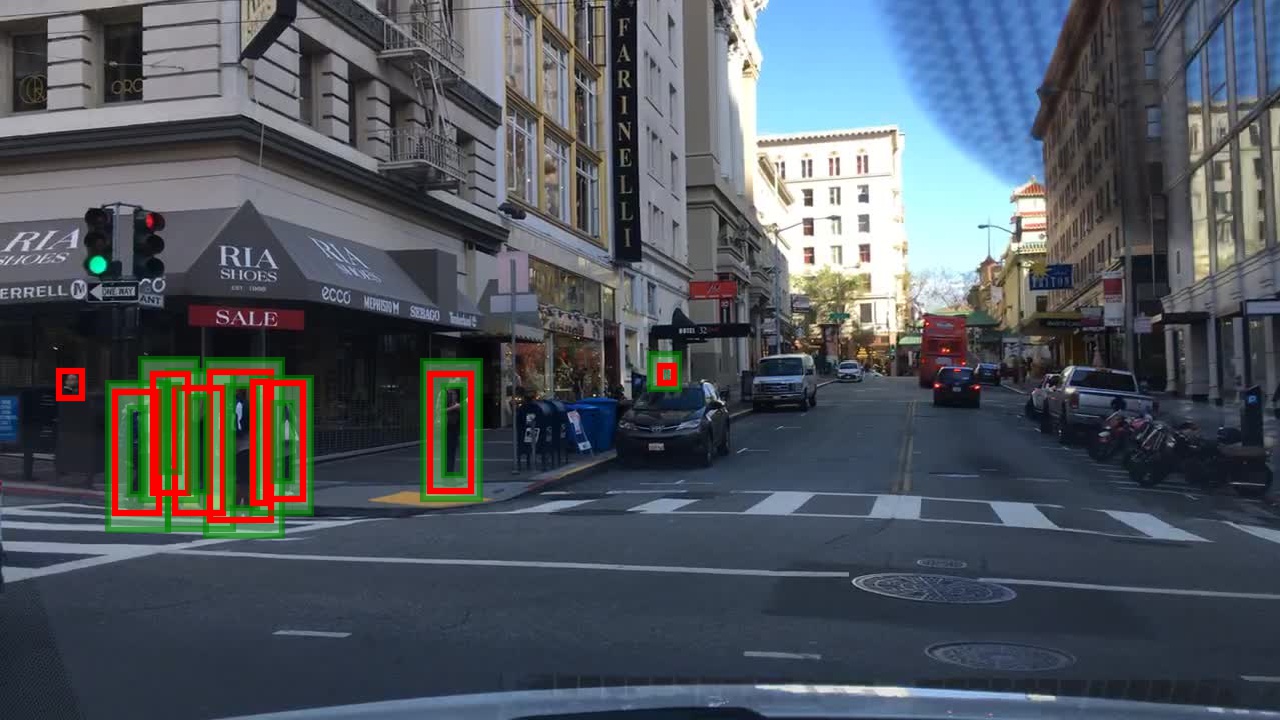}}
    \end{subfigure}
    \vspace{.005\textwidth}%
    \newline  
    \begin{subfigure}{.31\textwidth}
        \centering
        \resizebox{\textwidth}{!}{\includegraphics{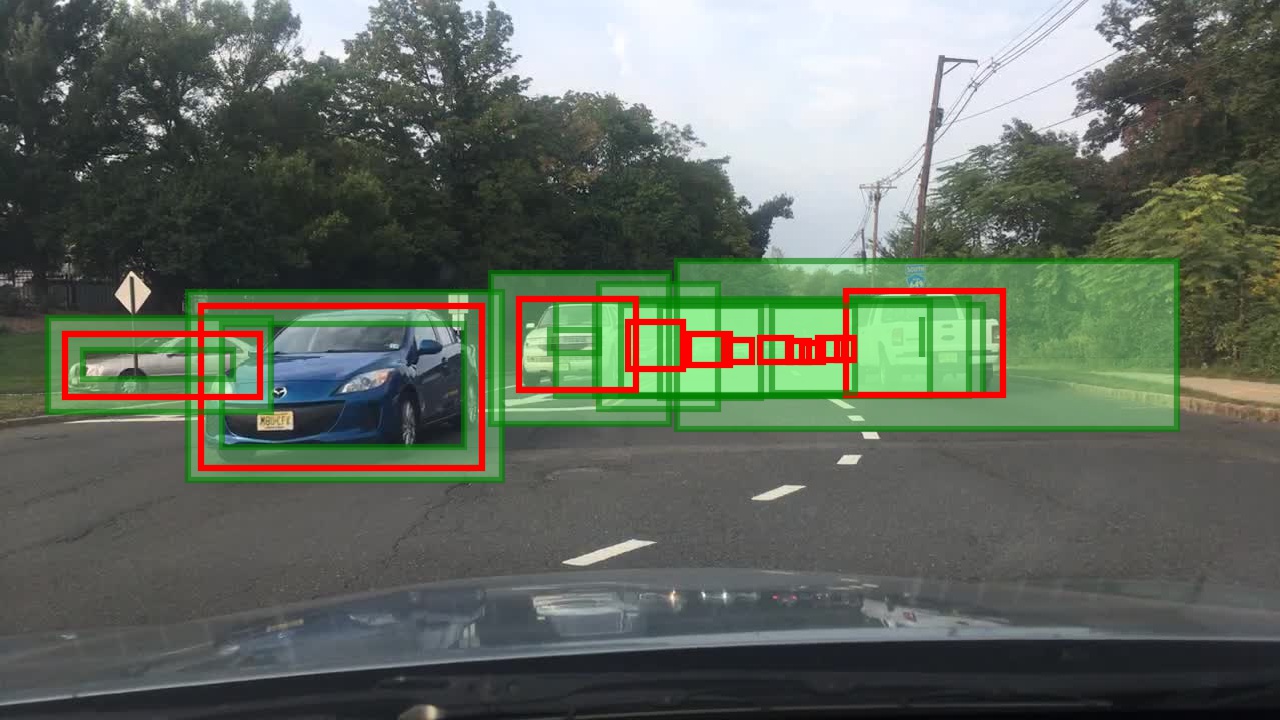}}
    \end{subfigure}%
    \hspace{.005\textwidth}%
    \begin{subfigure}{.31\textwidth}
        \centering
        \resizebox{\textwidth}{!}{\includegraphics{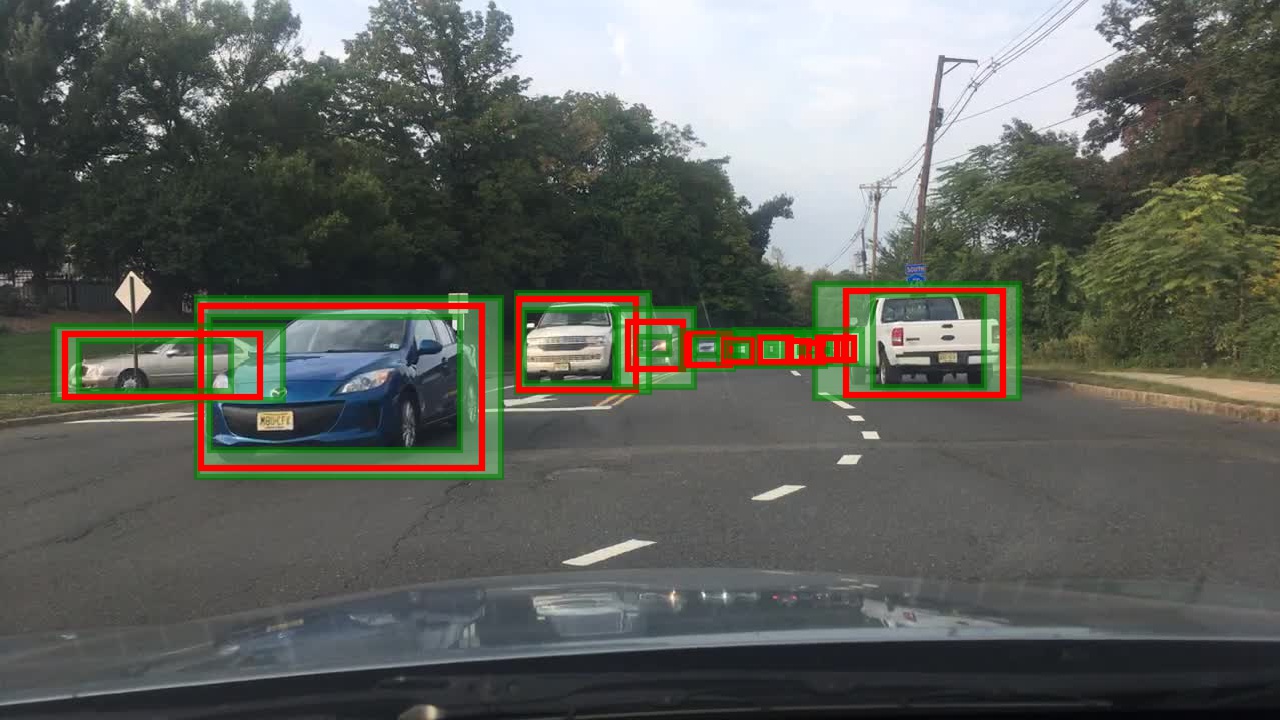}}
    \end{subfigure}%
    \hspace{.005\textwidth}%
    \begin{subfigure}{.31\textwidth}
        \centering
        \resizebox{\textwidth}{!}{\includegraphics{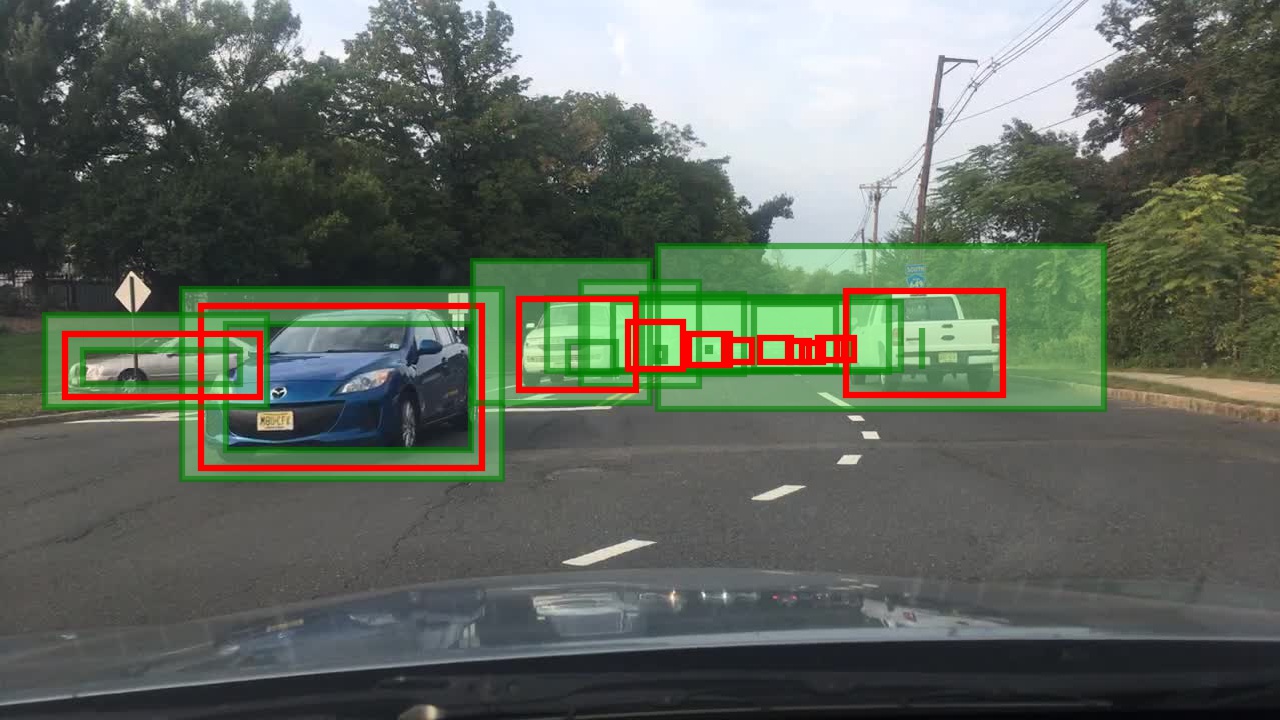}}
    \end{subfigure}
    
    \caption{Examples of conformal bounding box intervals produced by our two-step approach on BDD100k for a mixed set of classes. \emph{Left to right by column}: using ClassThr in combination with Box-Std, Box-Ens or Box-CQR. True bounding boxes are in red, two-sided prediction interval regions are shaded in green.}
    \label{fig:pi-plots-bdd100k}
\end{figure*}